\documentclass[12pt]{l4dc2021}

\newcommand{\arxivversion}[0]{}

\ifjmlrutilsmaths %
    \title[Investigating Compounding Prediction Errors]{Investigating Compounding Prediction Errors in \\ Learned Dynamics Models}
    \usepackage{times}

    \author{%
    \Name{Nathan Lambert}\Email{nol@berkeley.edu}\\
    \Name{Kristofer Pister}\Email{ksjp@berkeley.edu}\\
    \addr University of California, Berkeley
    \AND
      \Name{Roberto Calandra} \Email{rcalandra@fb.com}  \\
      \addr Facebook AI Research 
    }

    \usepackage{wrapfig}

\else
    \usepackage{arxiv}
    \usepackage{subcaption}
    \usepackage{wrapfig}
    \usepackage[english]{babel}
    \usepackage{amsthm}
    \theoremstyle{remark}
    \newtheorem{remark}{Remark}
    \title{%
    Understanding Compounding Prediction Errors in Learned Dynamics Models}
    \author{
  Nathan Lambert \\
  University of California, Berkeley\\
  \texttt{nol@berkeley.edu} \\
  \And
  Roberto Calandra \\
  Facebook AI Research \\
  \texttt{rcalandra@fb.com}  \\
  \And
 Kristofer Pister \\
  University of California, Berkeley\\
  \texttt{ksjp@berkeley.edu} \\
}
\fi

\usepackage{amsmath}
\usepackage{graphicx}
\usepackage{amsfonts}
\usepackage{siunitx}
\usepackage{dsfont}
\usepackage{hyperref}
\usepackage{scalerel}
\newcommand{\sect}[1]{Sec.~\ref{#1}}
\newcommand{\fig}[1]{Fig.~\ref{#1}}
\newcommand{\eq}[1]{Eq.~\eqref{#1}}

\newcommand{\tab}[1]{Tab.~\ref{#1}}

\usepackage{float}
\usepackage{booktabs} %

\usepackage{siunitx}
\usepackage{xcolor}

\newcommand{\mdptime}[0]{t}
\newcommand{\mdpstate}[0]{s}
\newcommand{\mdppredstate}[0]{\hat{s}}
\newcommand{\mdpaction}[0]{a}
\newcommand{\nextstate}[0]{\mdpstate_{\mdptime+1}}
\newcommand{\state}[0]{\mdpstate_\mdptime}
\newcommand{\action}[0]{a_t}
\newcommand{\hor}[0]{h}
\newcommand{\dynmod}[0]{f}

\usepackage{tikz}

\newcommand{\cblock}[3]{
  \hspace{-1.5mm}
  \begin{tikzpicture}
    [
    node/.style={square, minimum size=10mm, thick, line width=0pt},
    ]
    \node[fill={rgb,255:red,#1;green,#2;blue,#3}] () [] {};
  \end{tikzpicture}%
}
\makeatletter
\newcommand{\bigplus}{%
  \DOTSB\mathop{\mathpalette\mattos@bigplus\relax}\slimits@
}
\newcommand\mattos@bigplus[2]{%
  \vcenter{\hbox{%
    \sbox\z@{$#1\sum$}%
    \resizebox{!}{0.9\dimexpr\ht\z@+\dp\z@}{\raisebox{\depth}{$\m@th#1+$}}%
  }}%
  \vphantom{\sum}%
}
\makeatother
\usepackage{amssymb}
\usepackage{fdsymbol}
\usepackage{wasysym}
\newcommand{\code}[0]{\url{https://github.com/natolambert/continuousprediction/tree/compound}}

\renewcommand{\vec}[1]{\boldsymbol{#1}}				%

\newcommand{\D}[0]{\mathcal{D}} 				%

\DeclareMathOperator*{\argmin}{arg\,min}

\newcommand{\pole}[0]{\rho}

\begin{document}
\maketitle

\begin{abstract}
    Accurately predicting the consequences of agents' actions is a key prerequisite for planning in robotic control. 
Model-based reinforcement learning (MBRL) is one paradigm which relies on the iterative learning and prediction of state-action transitions to solve a task.
Deep MBRL has become a popular candidate, using a neural network to learn a dynamics model that predicts with each pass from high-dimensional states to actions. 
These ``one-step'' predictions are known to become inaccurate over longer horizons of composed prediction -- called the compounding error problem.
Given the prevalence of the compounding error problem in MBRL and related fields of data-driven control, we set out to understand the properties of and conditions causing these long-horizon errors.
In this paper, we explore the effects of subcomponents of a control problem on long-term prediction error: including choosing a system, collecting data, and training a model.
\ifx\arxivversion\undefined
With this analysis of compounding error, we delineate a set of key considerations for practitioners to understand the potential errors when modeling a new system.
\else
\fi
These detailed quantitative studies on simulated and real-world data show that the underlying dynamics of a system are the strongest factor determining the shape and magnitude of prediction error. 
Given a clearer understanding of compounding prediction error, researchers can implement new types of models beyond ``one-step'' that are more useful for control. 
\end{abstract}

\ifjmlrutilsmaths
    \begin{keywords}
    Dynamics Modeling, Robotics, Model-based Reinforcement Learning
    \end{keywords}
\else
    \keywords{Dynamics Modeling, Robotics, Long-term Predictions}
\fi

\section{Introduction}
\label{sec:intro}

Learning accurate dynamics models is crucial to many real-world robotics applications.
Central to the origins of dynamics model learning is the field of optimal control, where a system can be controlled when given access to an analytical dynamics model~\citep{kirk2004optimal}.
To apply these successful control and planning techniques to systems without known dynamics, system identification emerged as a set of techniques designed to compute a representative set of system parameters (e.g. mass of a robotic component) key to expressing the motion and use them with optimal control~\citep{ljung1999system}.
With the growth of data-driven systems, learning dynamics models for control has shifted to be increasingly task-centric, online, and free of prior knowledge of a system.
The field of model-based reinforcement learning (MBRL) showcases this process and has been used to solve many robotic tasks by iteratively learning a black-box model~\citep{Deisenroth2011PILCO,chua2018deep,nagabandi2019deep,janner2019trust}.
Central to recent successes in model-based reinforcement learning are one-step dynamics models; where they have been used for online model predictive control (MPC)~\citep{chua2018deep,nagabandi2019deep, lambert2019low} or value-estimation and offline rollouts of imagined policies~\citep{janner2019trust}. 
These models are known to be subject to an issue of \textit{compounding prediction error}~\citep{clavera2018model, wang2019benchmarking}, where long-horizon prediction often diverge over time to the point of being unusable.
In this work, we look to understand and study the causes of these compounding errors in order to improve the performance of future model-predictive agents.

Regardless of the prediction errors, the simple one-step parametrization with deep neural networks has been successful over a variety of real-world and simulated applications.
In order to predict far into the future, the models use composed function passes, where even small errors can grow rapidly to make the predictions difficult to rely on for ranking proposed actions.
Specifically, this paper is centered around the questions of: Why is there compounding error? What are the numerical characteristics of the compounding error? What model design choices most heavily impact compounding error?
In order to address this, the experiments are structured to address three major factors in deploying a learned model for control: the underlying system dynamics, system noise, and system dimensionality; the distribution of collected data; and how the model is trained.
The paper contains a wide analysis on deep one-step predictive models on simulated and real-world systems showing the strongest correlation between unstable underlying system dynamics and high-error predictions. 
\ifx\arxivversion\undefined
\else
With this depth, a set of takeaways are included for the reader to be better aware of why and how compounding error could impact their systems.
\fi
With this information, we believe that practices in MBRL and other data-driven predictive agents can be improved to better leverage dynamics models for control.

\section{Related Works}
\label{sec:related}
\paragraph{Model-based Reinforcement Learning} 
Much recent work in model-based reinforcement learning (MBRL) uses one-step models, without explicitly addressing compounding error, to iteratively learn the agents environment and then leverage it for control~\citep{Deisenroth2011PILCO,williams2017information,chua2018deep,lambert2019low,nagabandi2019deep}.
Given the recent successes, the training of the forward one-step models is not well studied and a variety of numerical behaviors have been recorded.
For example, the Model-based Policy Optimization (MBPO) algorithm uses short horizons to avoid the compounding error problem~\citep{janner2019trust}, but there-in loses out on a lot of the potential for having a learned model and anticipating the future.
With another MBRL algorithm, probabilistic ensembles with trajectory sampling (PETS)~\citep{chua2018deep}, dynamically tuning the predictive horizon and model training dramatically improves the downstream performance~\citep{zhang2021importance}.
The centrality of predictive horizon to performance indicates that the compounding error over time is important to future of MBRL, but its causes are not well understood.
Another horizontal study of model learning in MBRL proposes and details the metrics of many different model variants used in MBRL~\citep{kegl2021model}, but does not provide insight in how to limit the effects of compounding error.

\paragraph{Learning One-step Dynamics Models}
Single-step learned dynamics models are effective across many domains despite their well known compounding errors.
Early examples include using one-step models for studying robot dynamics~\citep{punjani2015deep} or learning a model for linear controllers~\citep{bansal2016learning}.
Additionally, other model types such as single-step Gaussian Processes (GPs)~\citep{Deisenroth2011PILCO} and linear models~\citep{fu2016one, bansal2017goal} have been applied to multiple lower-dimensional robotic learning tasks.
GPs are exciting due to their more structured handling of uncertainty, which can allow for safety in long-horizon predictions~\citep{koller2018learning}, but they have not scaled as well as deep learning in high-dimension and large-data tasks.
There has been substantial development in deep learning and other data-driven methods to facilitate more useful methods for learning models for control.
Structured mechanical models~\citep{gupta2020structured} and Lagrangian neural networks~\citep{cranmer2020lagrangian} both use a constrained learning setup to restrict their one-step dynamics models to those satisfying smooth constraints and de-valuing transitions likely to be governed by random noise.
These methods have been shown to be data-efficient for smooth systems, but their effect on compounded prediction accuracy is not yet studied.

\paragraph{Mitigating Compounding Error}
Compounding error is referenced in many recent MBRL papers, including state-of-the-art algorithms~\citep{clavera2018model, chua2018deep} and benchmarking efforts~\citep{wang2019benchmarking}.
Recently, more methods have proposed potential solutions.
\citep{heess2015learning} avoids compounding error by only predicting with the model with real observations as inputs to avoid distribution drift via long predictions.
Most methods avoid compounding error by tweaking the dynamics model optimization, including imitation-learning inspired predictive models~\citep{venkatraman2015improving}, multi-step value estimators~\citep{asadi2019combating}, and flexible prediction horizons~\citep{xiao2019learning}.
Most work using model predictive control uses predictive horizons of 20-30 steps with limited cross validation of model accuracy versus prediction horizon or agent performance, as suggested in~\citep{lambert2020objective}. 
The trajectory-based model proposes a new training paradigm to avoid the compounding error by embedding time dependence in prediction, but it is currently limited by its  requirement of closed form controllers~\citep{lambert2020learning}.

\section{Background}
\paragraph{Markov Decision Process} 
We now describe the formulation we use to evaluate compounding error.
At time $t$, the environment is represented by the state $\state \in \mathbb{R}^{d_s}$, the action  $\action \in \mathbb{R}^{d_a}$, and the reward function $r(\state,\action): \mathbb{R}^{d_s \times d_a} \mapsto \mathbb{R}$ following the Markov Decision Process (MDP)~\citep{bellman1957markovian}.
With this data, the task is learn a dynamics model $f_\theta(\state, \action)$ to represent the transition function $f : \mathbb{R}^{d_s \times d_a} \mapsto \mathbb{R}^{d_s}$.
The data $\D=\big\{(s_i, a_i, s_{i+1}) \big\}_{i=1}^N$ used to train the model is often subject to the behavior of an agents' control law $\pi(\cdot)$, called the policy.

\paragraph{One-step Dynamics Models}
Given the state~$\state$, and the action~$\action$, one-step dynamics models predicts the next state~$\nextstate$ of an MDP. 
A given model object can use different prediction formulations.
We use the delta-state formulation that is popular for regularizing the prediction distribution, $\nextstate =\state+\dynmod_\theta(\state,\action)$, and compare to the true-state variant, $\nextstate =\dynmod_\theta(\state,\action)$.
The true-state models are denoted with a $\text{-S}$, such as the true-state probabilistic ensemble \textit{PE-S}.
These formulations can be used with multiple loss functions, including Mean Squared Error (MSE) for deterministic models and Negative Log Likelihood (NLL) for probabilistic models.
All model types can be used with an ensemble that weights predictions across multiple trained models, denoted with \textit{E}.

In this work we primarily compare the delta-state and true-state prediction formulations with simple deterministic models (\textit{D}) and with rich probabilistic ensembles (\textit{PE}).
For the probabilistic models, we propagate the trajectories with expectation based propagation, more options for the \textit{PE} models are detailed in \cite{chua2018deep}.
We compare one-step models to \textit{linear models} (LIN) based on least-squares learning of a linear predictor and \textit{zero models} (ZERO) that return a predicted state of the zero vector at each step.
Additional model training details are included in \sect{sec:training}.
\ifx\arxivversion\undefined

\else

\fi

\section{Methodology}
\subsection{Problem Formulation: Multi-step Compounding Error}
For the prediction of the long-term future, the one-step dynamic model is recursively applied as
\begin{equation}
\mdppredstate_{\mdptime+\hor}=\dynmod_\theta(\ldots \dynmod_\theta(\dynmod_\theta(\mdpstate_\mdptime,\mdpaction_\mdptime),\mdpaction_{\mdptime+1}) \ldots , \mdpaction_{\mdptime+\hor})\,.
\label{eq:onestepprop}
\end{equation}
Any parametrization of the dynamics model $\dynmod$ carries a prediction error~$\epsilon_\mdptime = \hat{s}_\mdptime - s_\mdptime$. 
It is often observed that this error grows multiplicatively by the next prediction's input being subject to all past errors in the prediction, as 
\begin{equation}
\mdppredstate_{\mdptime+\hor}=\dynmod_\theta(\ldots \dynmod_\theta(\dynmod_\theta(\mdpstate_\mdptime,\mdpaction_\mdptime)+\epsilon_t,\mdpaction_{\mdptime+1})+\epsilon_{\mdptime+1} \ldots , \mdpaction_{\mdptime+\hor})+\epsilon_{\mdptime+\hor}\,.
\label{eq:onesteperrorprop}
\end{equation}

The central metric we will use to quantify and visualize compounding error is the mean-squared prediction error (MSE) at each step. %
The action sequence used with a given trajectory, $\{\mdpaction_0, \mdpaction_1, \ldots, \mdpaction_h\}$ is provided when planning the trajectory and measuring its performance.
To that end, we train a dynamics model~$f_\theta$, and use it to predict to a horizon~$\hor$, steps into the future, generating a predicted trajectory $\hat{s}_i \forall i \in [1,\hor]$.
Then, the predicted error is computed by summing across all of the state dimensions at the given time-step as
$    \text{MSE}_t = \sum_{d=0}^{d_s} \| \mdppredstate_{t,d} - \mdpstate_{t,d}\|^2_2\,.
    \label{eq:MSE} $
    
For each experiment and environment, the MSE is normalized per-state to $[0,1]$ to make the calculated MSE represent average predictive accuracy proportional to the relative state error rather than the numerical error (e.g. to normalize across states of different state types like positions and velocities). 
We hope this makes the errors shown more intuitive -- an mean squared error of 1 represents an error across each state averaging to 100\%.

\begin{figure}[t]
    \centering
    \ifjmlrutilsmaths
        \subfigure[Poles at $0.05$.]{
        \includegraphics[width=0.23\linewidth]{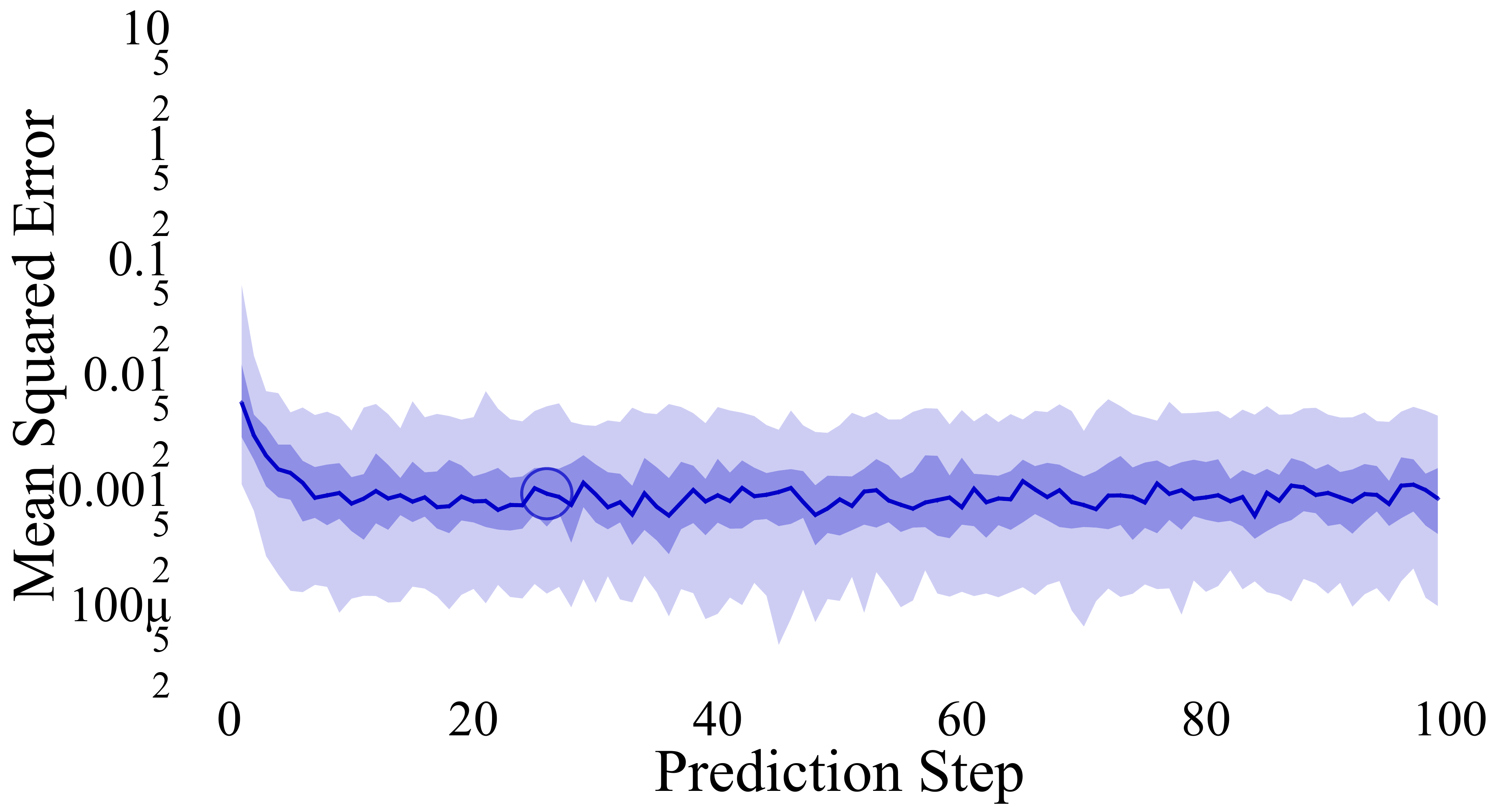}
        \label{fig:p005}
        }
        \subfigure[Poles at $0.10$.]{
        \includegraphics[width=0.23\linewidth]{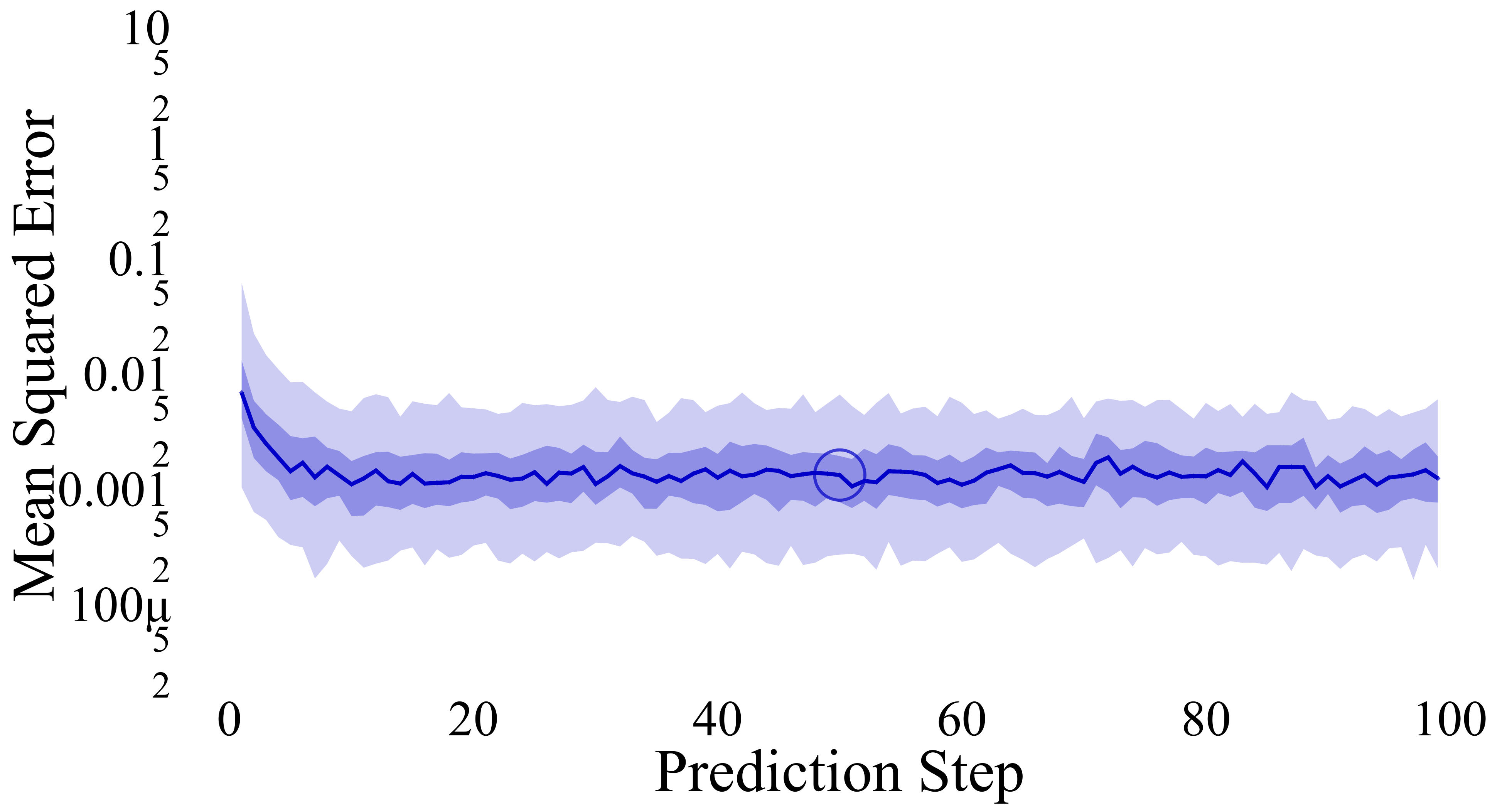}
        \label{fig:p010}
        }
        \subfigure[Poles at $0.25$.]{
        \includegraphics[width=0.23\linewidth]{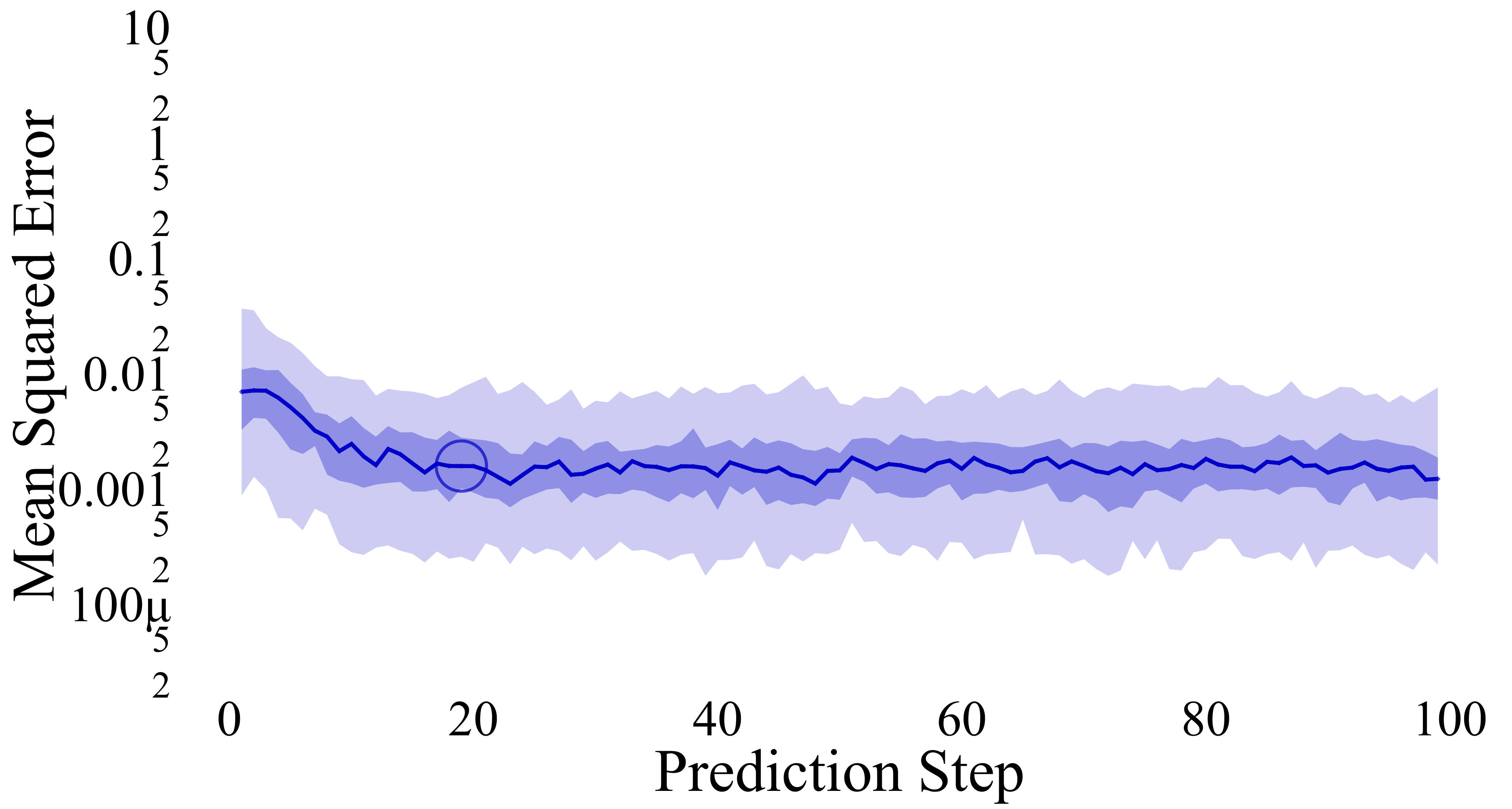}
        \label{fig:p025}
        }
        \subfigure[Poles at $0.50$.]{
        \includegraphics[width=0.23\linewidth]{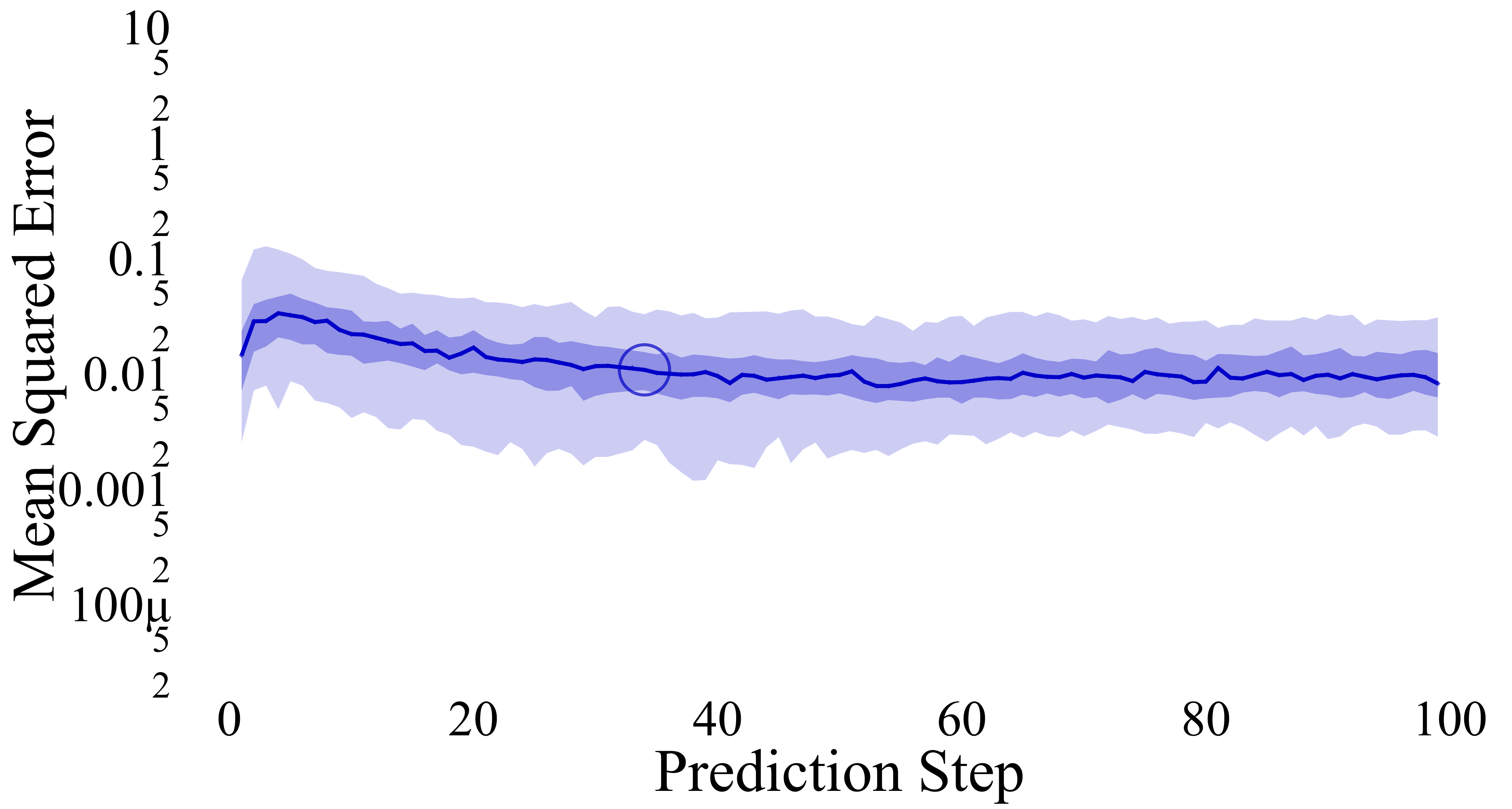}
        \label{fig:p050}
        }
        \\
        \subfigure[Poles at $0.75$.]{
        \includegraphics[width=0.23\linewidth]{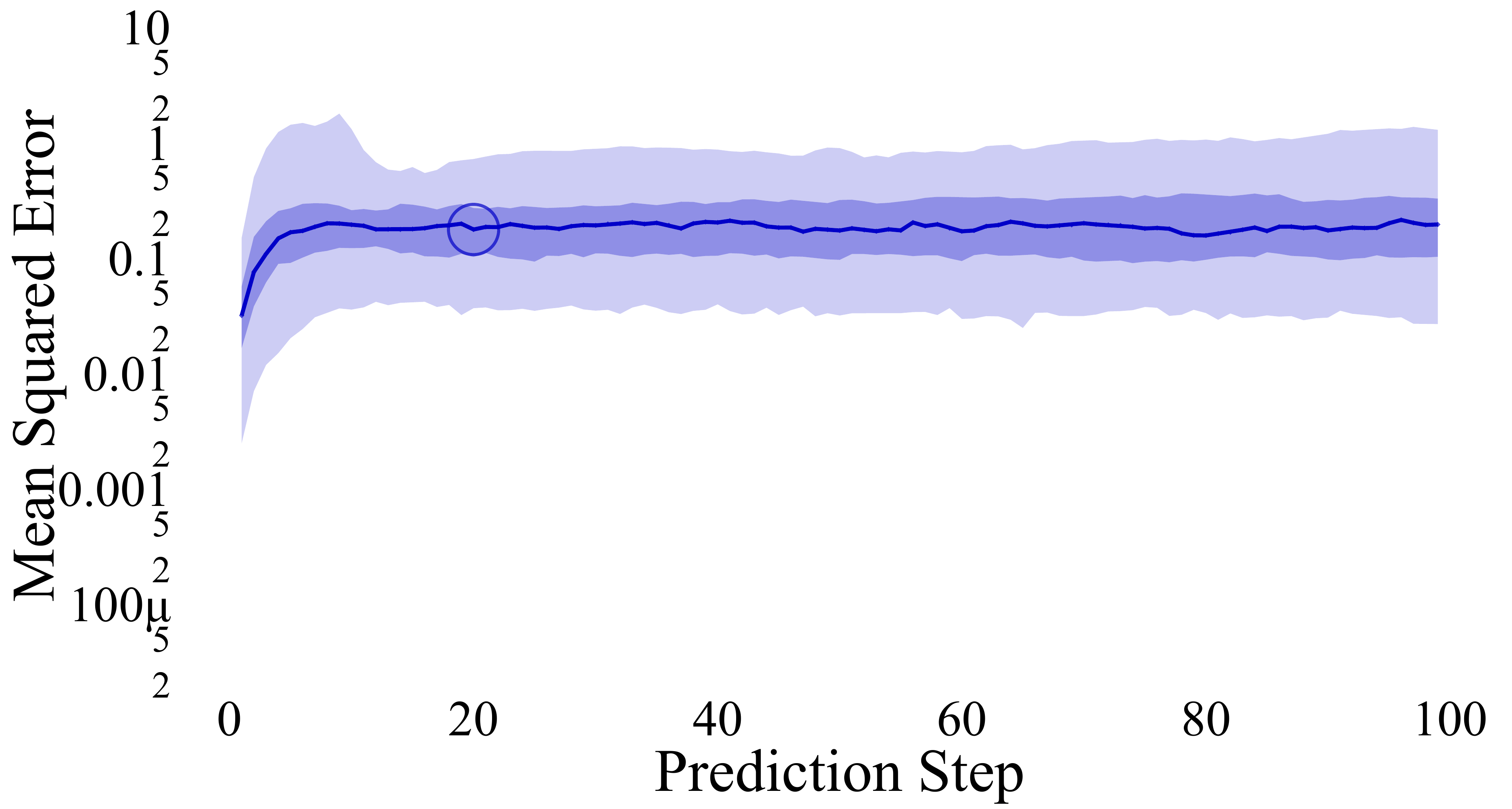}
        \label{fig:p075}
        }
        \subfigure[Poles at $0.90$.]{
        \includegraphics[width=0.23\linewidth]{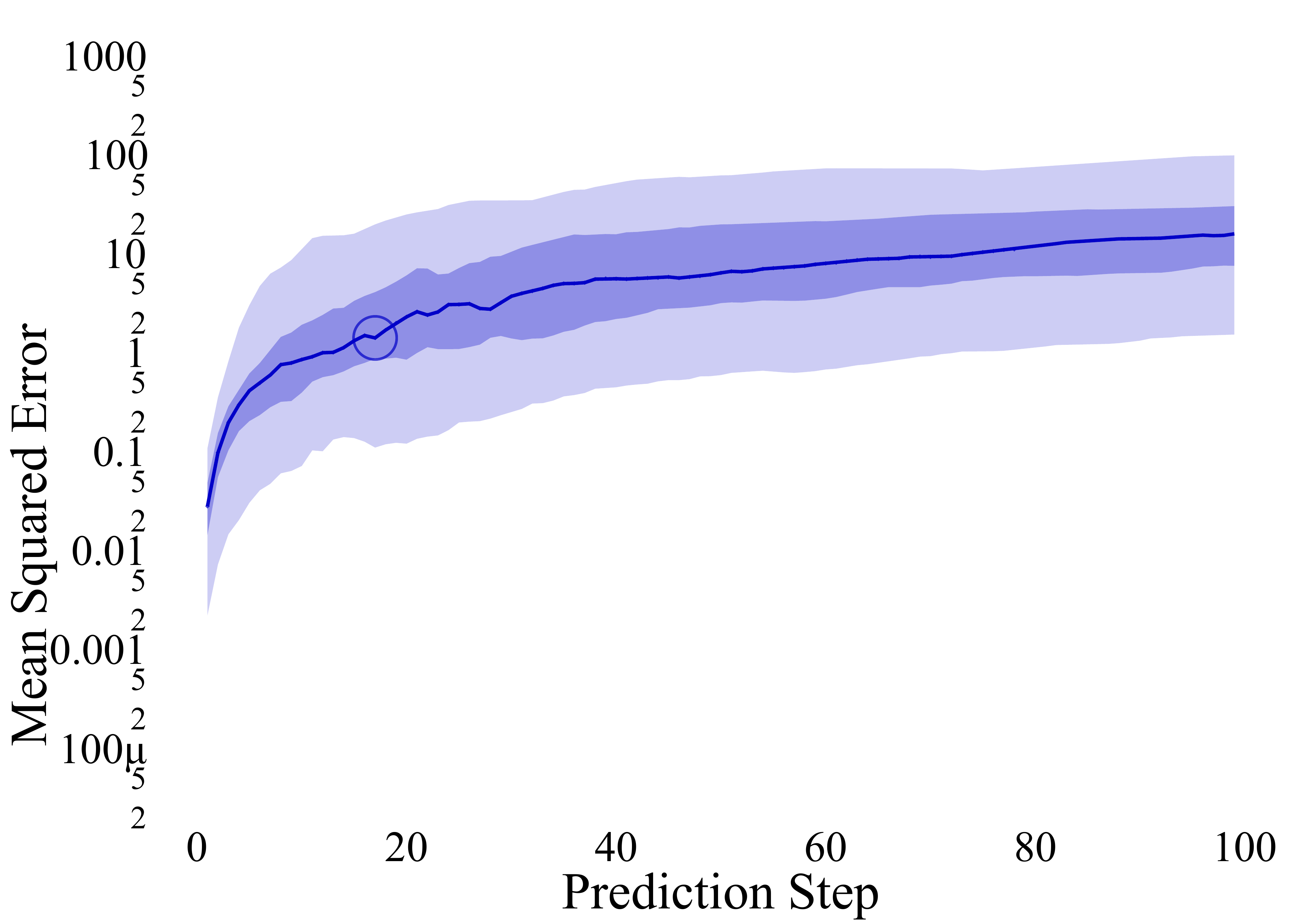}
        \label{fig:p090}
        }
        \subfigure[Poles at $0.95$.]{
        \includegraphics[width=0.23\linewidth]{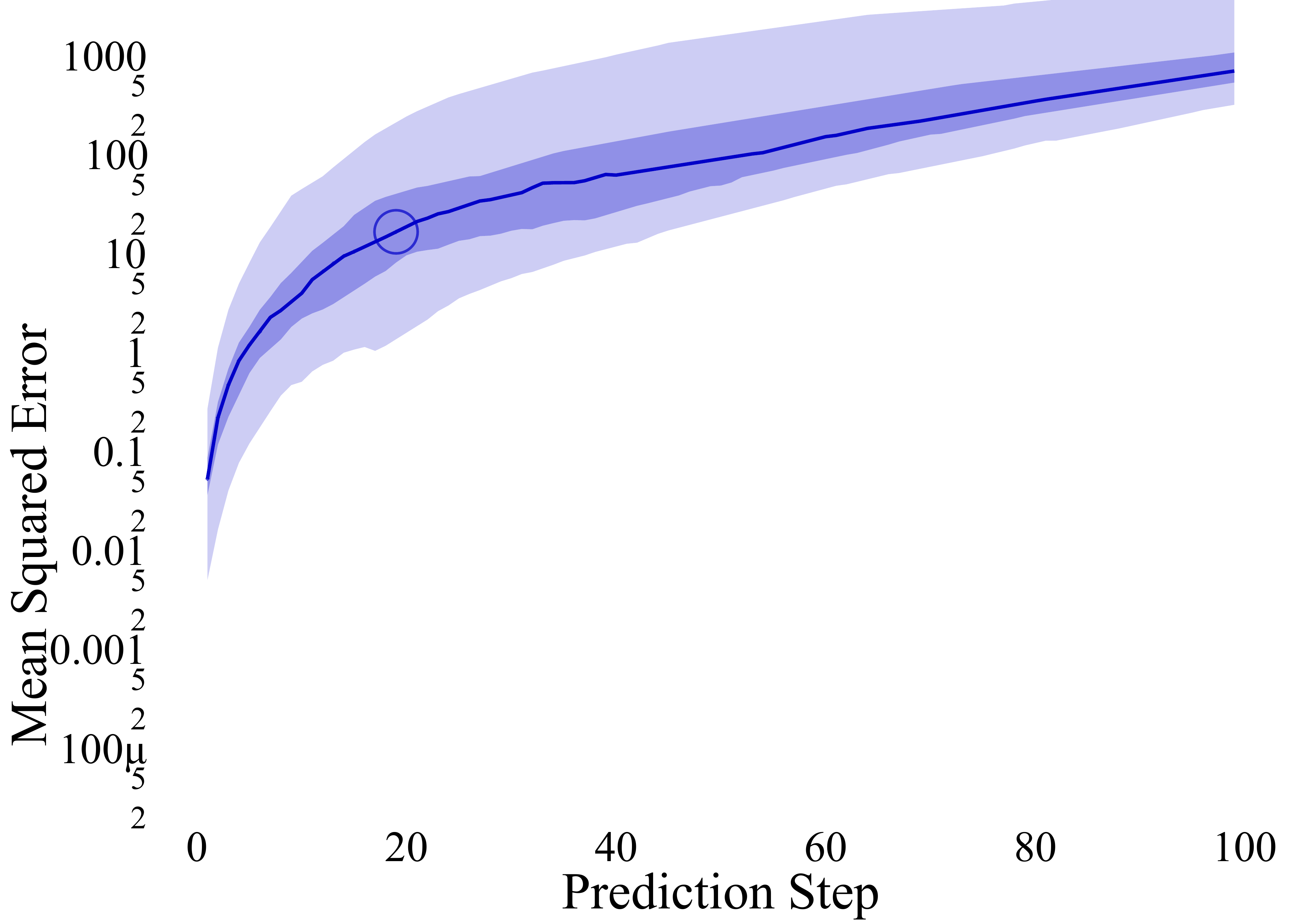}
        \label{fig:p095}
        }
        \subfigure[Poles at $1.00$.]{
        \includegraphics[width=0.23\linewidth]{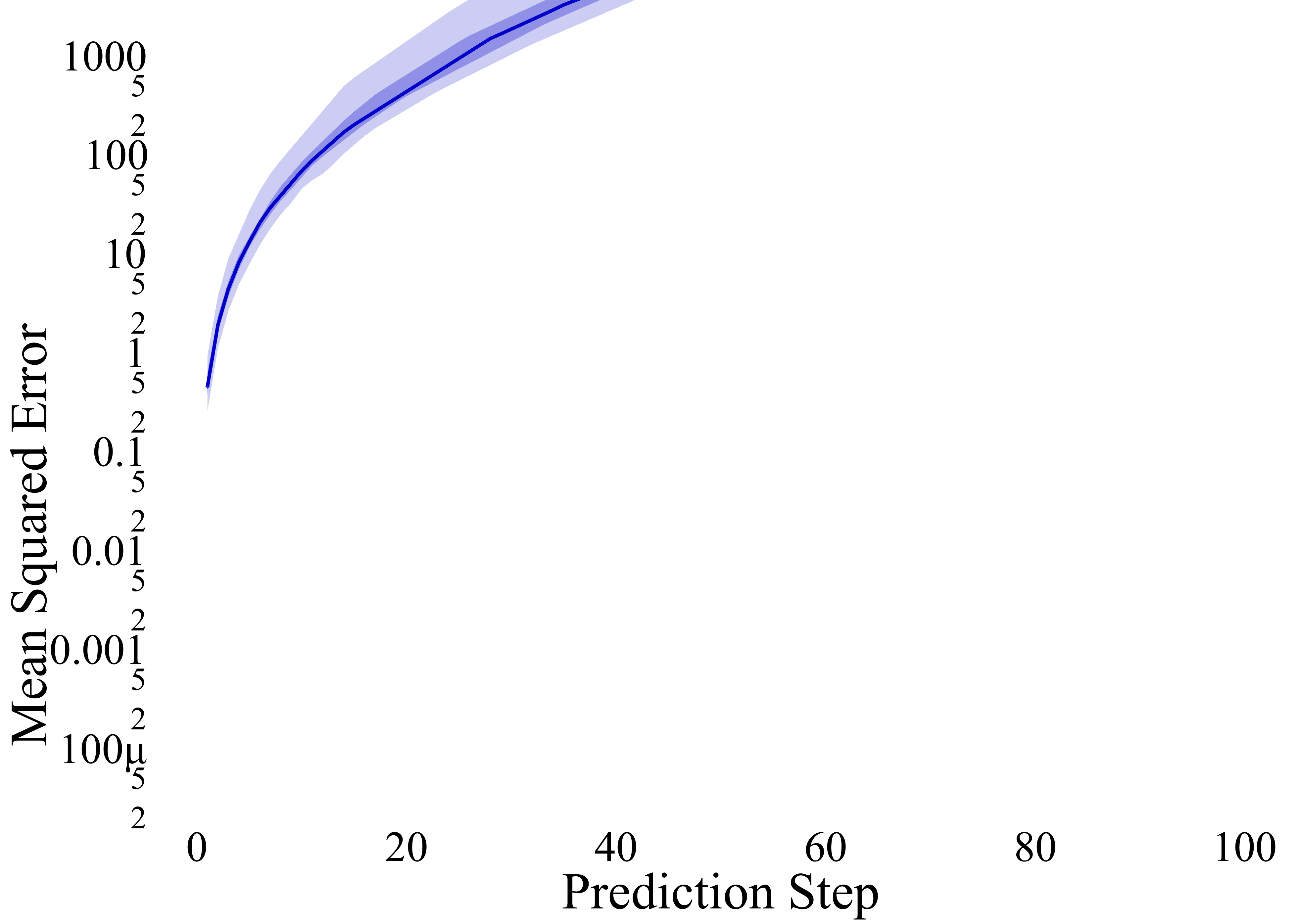}
        \label{fig:p100}
        }
        
    \else
        \begin{subfigure}[t]{0.3\linewidth}
            \centering
            \includegraphics[width=\linewidth]{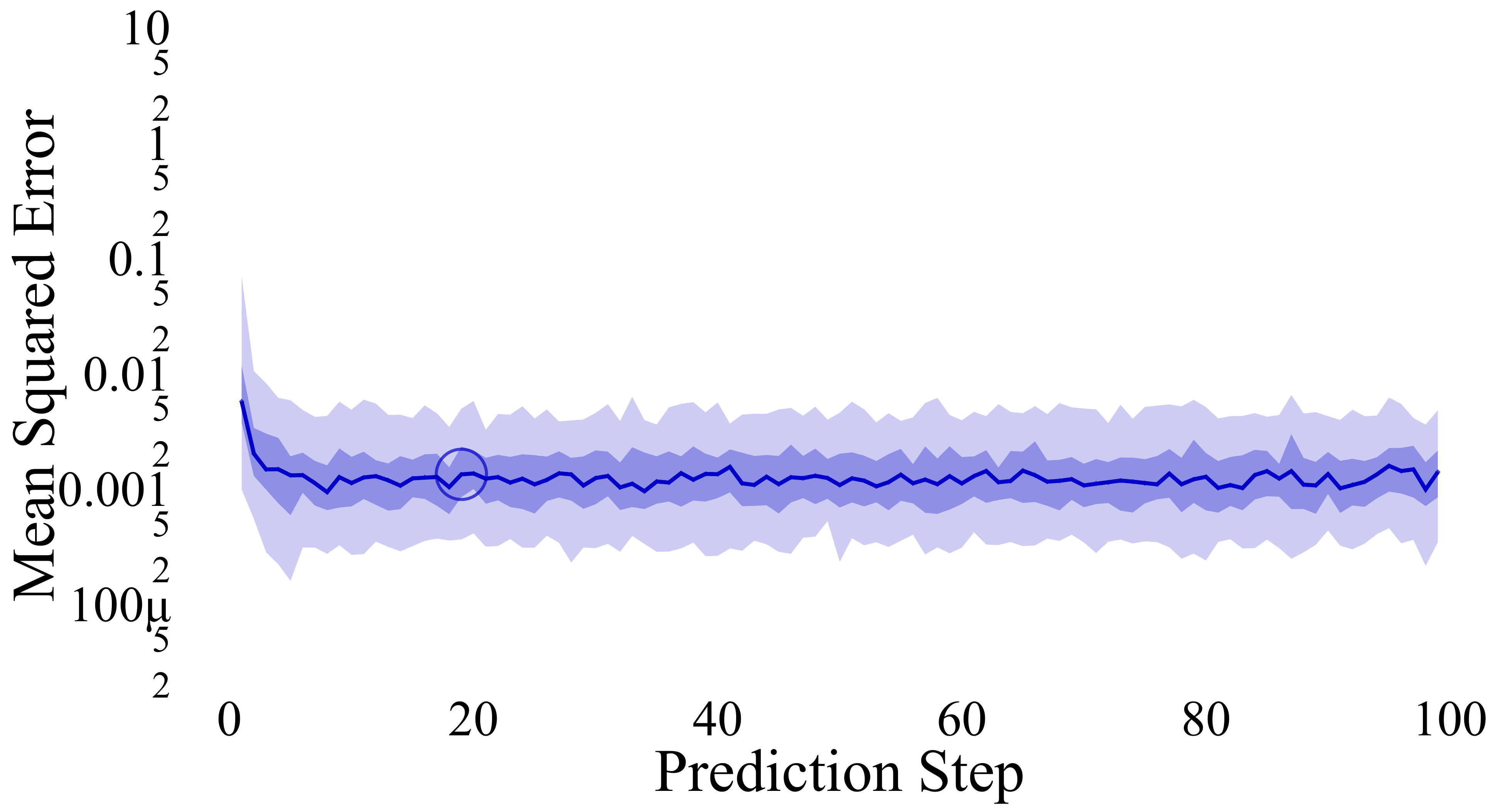}
            \caption{Poles, $\rho$,  $0.01$.}    
            \label{fig:256hidp001}
        \end{subfigure}
        \hfill
        \begin{subfigure}[t]{0.3\linewidth}  
            \centering 
            \includegraphics[width=\linewidth]{figures/p005/256hid.pdf}
            \caption{Poles, $\rho$,  $0.05$.}    
            \label{fig:256hidp005}
        \end{subfigure}
        \hfill
        \begin{subfigure}[t]{0.3\linewidth}
            \centering
            \includegraphics[width=\linewidth]{figures/p010/256hid.pdf}
            \caption{Poles, $\rho$,  $0.1$.}    
            \label{fig:256hidp010}
        \end{subfigure}
        \\
        \begin{subfigure}[t]{0.3\linewidth}  
            \centering 
            \includegraphics[width=\linewidth]{figures/p025/256hid.pdf}
            \caption{Poles, $\rho$,  $0.25$.}    
            \label{fig:256hidp025}
        \end{subfigure}
        \hfill
        \begin{subfigure}[t]{0.3\linewidth}  
            \centering 
            \includegraphics[width=\linewidth]{figures/p050/256hid.pdf}
            \caption{Poles, $\rho$,  $0.5$.}    
            \label{fig:256hidp050}
        \end{subfigure}
        \hfill
        \begin{subfigure}[t]{0.3\linewidth}
            \centering
            \includegraphics[width=\linewidth]{figures/p075/256hid.pdf}
            \caption{Poles, $\rho$,  $0.75$.}    
            \label{fig:256hidp075}
        \end{subfigure}
        \\
        \begin{subfigure}[t]{0.3\linewidth}  
            \centering 
            \includegraphics[width=\linewidth]{figures/p090/256hid_tall.pdf}
            \caption{Poles, $\rho$,  $0.90$.}    
            \label{fig:256hidp090}
        \end{subfigure}
        \hfill
        \begin{subfigure}[t]{0.3\linewidth}  
            \centering 
            \includegraphics[width=\linewidth]{figures/p095/256hid_tall.pdf}
            \caption{Poles, $\rho$,  $0.95$.}    
            \label{fig:256hidp950}
        \end{subfigure}
        \hfill
        \begin{subfigure}[t]{0.3\linewidth}
            \centering
            \includegraphics[width=\linewidth]{figures/p100/256hid_tall.pdf}
            \caption{Poles, $\rho$,  $1.0$.}    
            \label{fig:256hidp100}
        \end{subfigure}
    \fi
    \caption{
    Showing the compounding errors, formally the per-step MSE (median, $65^\text{th}$, and $95^\text{th}$ percentiles), of state-space systems shown in \eq{eq:statespace_mat} with different poles, $\pole$.
    Compounding errors vary substantially with the underlying poles of the environment and diverge when the poles approach instability. 
    All models are trained and evaluated on separate datasets of 100 trajectories.
    }
    \label{fig:compound}
    \vspace{-10pt}
\end{figure}

\subsection{Experimental Setting}
Here we provide a concise summary of the studied systems, with more details available in~\sect{sec:env_extra}.

\paragraph{State-space System}
To test the possible causes of compounding error, we test the ability of deep one-step models to predict variations of a clearly defined system.
Consider a state-space system defined with a state $s\in \mathbb{R}^{3}$ and an action $a\in\mathbb{R}$ as
$
    \mdpstate_{t+1} = \vec{A} \mdpstate_t + \vec{B} \mdpaction_t + \omega_t.
    \label{eq:statespace}
$
Therein, we define $\vec{A}$ and $\vec{B}$ as follows to control the poles of the system:
\begin{align}
    \vec{A} &= \begin{bmatrix}
    \rho & a_1 & a_2 \\
    0 & \rho & a_3 \\
    0 & 0 & \rho
    \end{bmatrix}, \quad
    \vec{B} = \begin{bmatrix}
    b_1 \\
    b_2 \\
    b_3 
    \end{bmatrix} \,.
    \label{eq:statespace_mat}
\end{align}
We set the desired eigenvalues, or poles, of the system to be $\rho$.
The other parameters of the system are sampled randomly for each trial as $a_i,b_i \sim \mathcal{U}(-1,1)$, which act as a source of uncertainty.
The default process-noise in the environment $\omega_i \sim \mathcal{U}(-0.01,0.01)$.
All actions are chosen randomly from $\mathcal{U}(-1,1)$ and act via the randomly generated $\vec{B}$ matrices.

\paragraph{Other Environments} 
We present simulated results from three other simulated environments: the Cartpole, Reacher, and Quadrotor tasks (for more details, see \citep{lambert2020learning}).
In the Cartpole ($d_s=4$, $d_a=1$) and Quadrotor ($d_s=9$, $d_a=4$) tasks the agent attempts to balance the agent around an unstable fixed point.
In the Reacher manipulation task ($d_s=15$, $d_a=5$), the agent must control the end-effector to a randomly generated point in the state-space.

\section{Experimentally Studying Compounding Error}
\label{sec:experiments}
In this section we progress through different components of learning a model for control and detail their effect on compounding prediction error.
We start with the central consideration -- the underlying dynamics of the system, then we investigate two key important characteristics of a system: noise levels and state-space dimensionality.
Next, we explore an important question for MBRL: the effect of dataset density on prediction accuracy.
Finally, we conclude by showing that many model training effects such as the model parametrization and normalization play a lesser role in compounding errors.
Across these experiments we enact large sweeps across key properties to illustrate macro trends, which comes at the loss of specificity on a per-environment basis -- specific applications can benefit from finer analysis of these variables. 
Code for reproducing the experiments is available~\footnote{Code: \code} and additional experiments are included in~\sect{sec:additional}.

\begin{figure}
    \centering
    \ifjmlrutilsmaths
        \subfigure[Cartpole. $d_s + d_a = 5$.]{
        \includegraphics[width=0.31\linewidth]{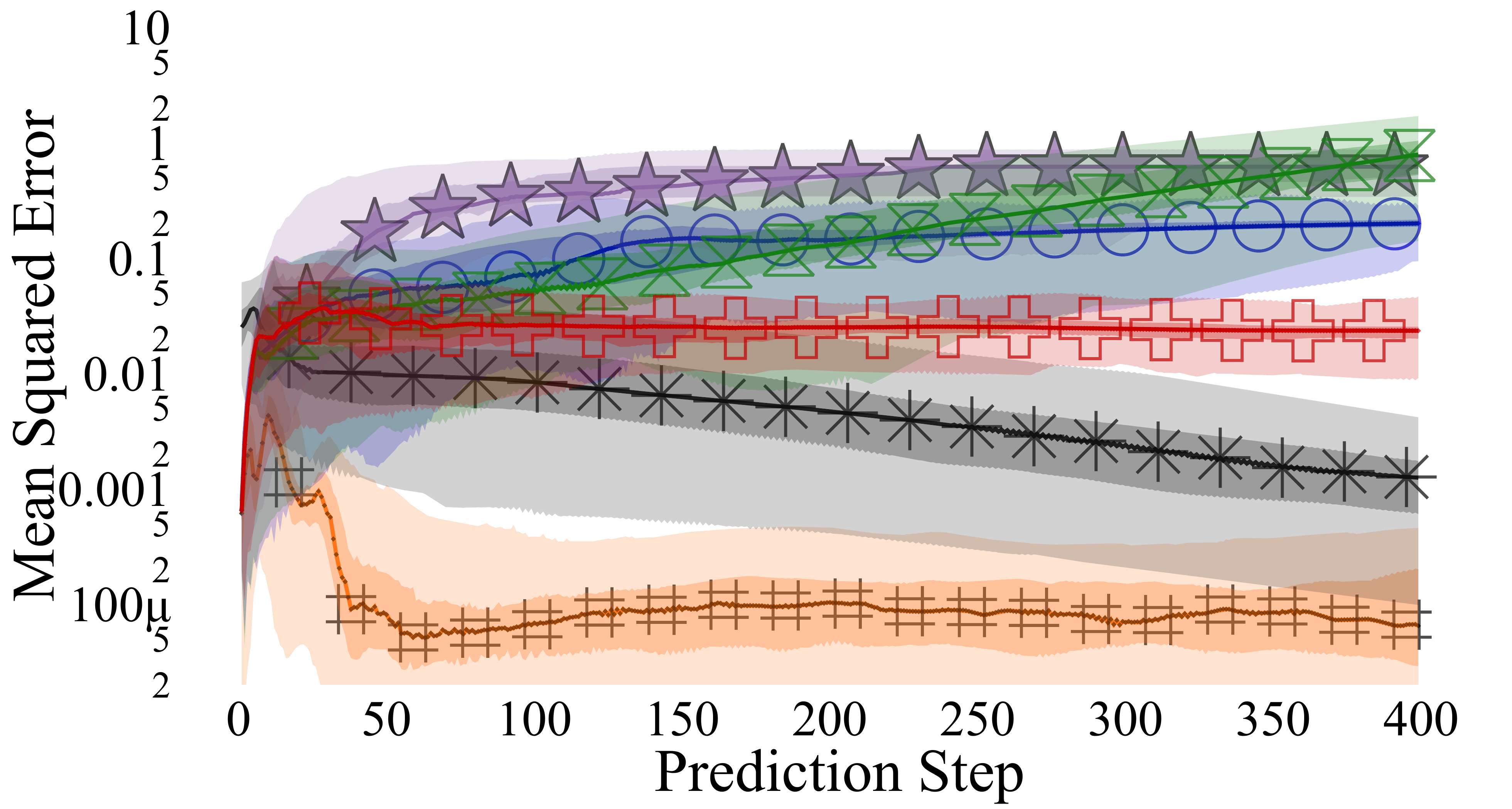}
        \label{fig:cartpole-compare}}
        \hfill
        \subfigure[Quadrotor. $d_s + d_a = 13$.]{
        \includegraphics[width=0.31\linewidth]{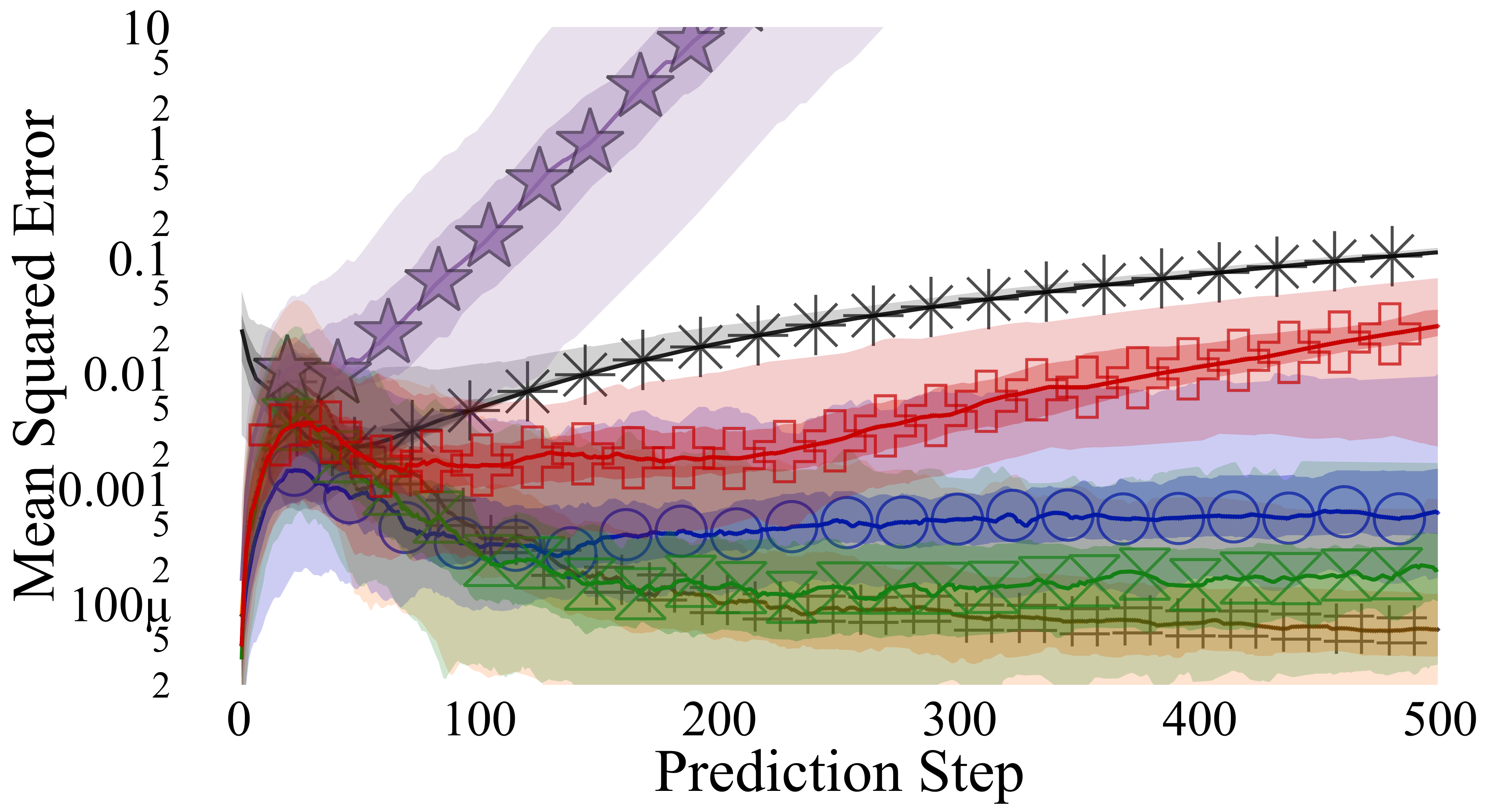}
        \label{fig:crazyflie-compare}}
        \hfill
        \subfigure[Reacher. $d_s + d_a = 20$.]{
        \includegraphics[width=0.31\linewidth]{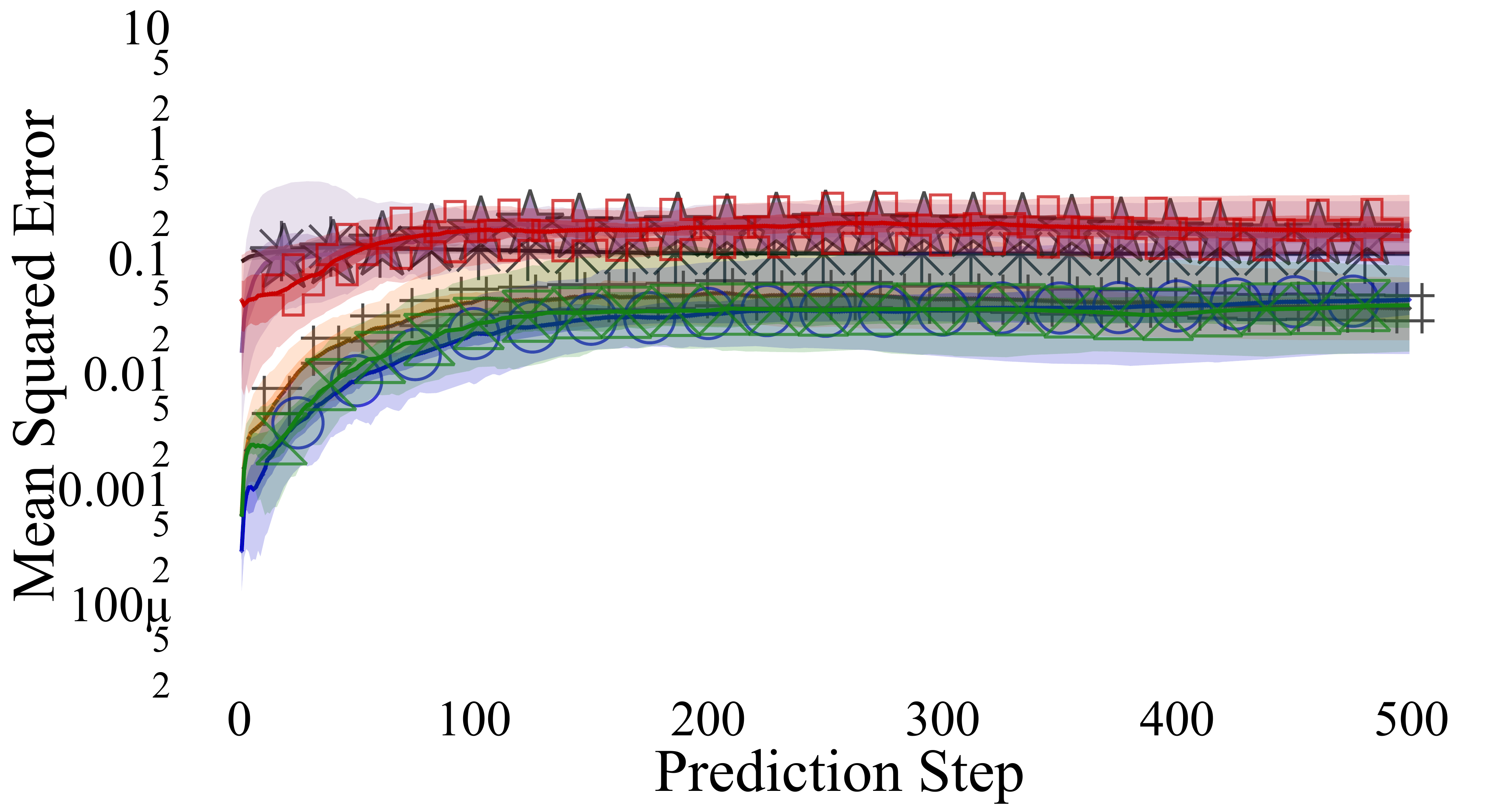}
        \label{fig:reacher-compare}
}
    \else
         \begin{subfigure}[t]{0.32\linewidth}  
            \centering 
            \includegraphics[width=\linewidth]{figures/deltaVtrue/cartpole_full.pdf}
            \caption{Cartpole. $d_s + d_a = 5$. }  
            \label{fig:cp-mod-fb}
        \end{subfigure}
        ~
        \begin{subfigure}[t]{0.32\linewidth}  
            \centering 
            \includegraphics[width=\linewidth]{figures/deltaVtrue/crazyflie_full.pdf}
            \caption{Quadrotor. $d_s + d_a = 13$. }    
            \label{fig:crazyflie-compare}
        \end{subfigure}
        ~
        \begin{subfigure}[t]{0.32\linewidth}  
            \centering 
            \includegraphics[width=\linewidth]{figures/deltaVtrue/reacher_full.pdf}
            \caption{Reacher. $d_s + d_a = 20$. }  
            \label{fig:reacher-compare}
        \end{subfigure}
    \fi
    \input{figure_latex/legend_full}
    \caption{
    Comparing the MSE of prediction error per-step 
    (median, $65^\text{th}$, and $95^\text{th}$ percentiles) on common model types and parametrizations on simulated robotic tasks of different dynamics, simulators, and dimension.
    There is a trend of error of predictions increasing with the task difficulty, but there is high variability in the performance of any one model type when comparing across platforms.
    All model types are trained and evaluated on the same datasets, maintaining separate datasets of 100 trajectories for test-train split.
    }
    \label{fig:examples-compare}
\end{figure}

\subsection{System: Underlying Dynamics}
\label{sec:dynamics}
The underlying dynamics of a system to be modelled controls the relative complexity of the prediction landscape. 
One way to characterize the behavior of a system's behavior is through the poles, often computed as the eigenvalues for linear systems. 
To continue the state-space example, a discrete-time system is  \textit{stable} when the eigenvalues are within the unit-circle, $\|\rho\|<1$.
We vary the eigenvalues of our state-space system shown in \eq{eq:statespace_mat} across a range of mostly stable and psuedo-stable values to compare the compounding error.

The results in \fig{fig:compound} indicate that as eigenvalues approach instability, prediction accuracy quickly degrades. 
For stable systems, the error growth over time does not compound, but decreases over a short horizon before reaching steady state.
Those systems with truly unstable poles, $\|\rho\|>1$, diverge so rapidly that plotting and computing the magnitude of error is computationally intractable. %
It is also interesting that when the poles are notably stable, the relative \textit{stable-ness} of the system does not have a large baring on prediction accuracy, as shown by the similarity in error between poles of $\rho < 0.25$.
Examples of the compounding error on different simulated robotic tasks is shown in \fig{fig:examples-compare}, which show substantial variation in compounding errors.
With complex robotic systems it is often difficult to directly identify the eigenvalues, further increasing the difficulty of cross-platform comparisons. 
The Cartpole and Quadrotor environments represent stabilization tasks, which have similar error profiles when compared to the Reacher maniuplation task.
The variation between platforms in both the magnitude and shape of compounding error motivates a deeper study of the causes, which could be revealed in other properties of the system such as the dimension or noise.

\ifx\arxivversion\undefined

\else
\begin{remark}
The underlying dynamics of a system have a heavy impact on when the prediction accuracy will be poor. 
Unstable systems should deploy one-step models with extra care.
\end{remark}

\begin{remark}
The shape of long-term prediction error for simple, stable-systems is not one of multiplicative growth. As the horizon increases the error rapidly reaches a steady-state value.
\end{remark}
\fi

\begin{figure}[t]
    \centering

    \ifjmlrutilsmaths
        \subfigure[$0\times$ noise.]{
        \includegraphics[width=0.315\linewidth]{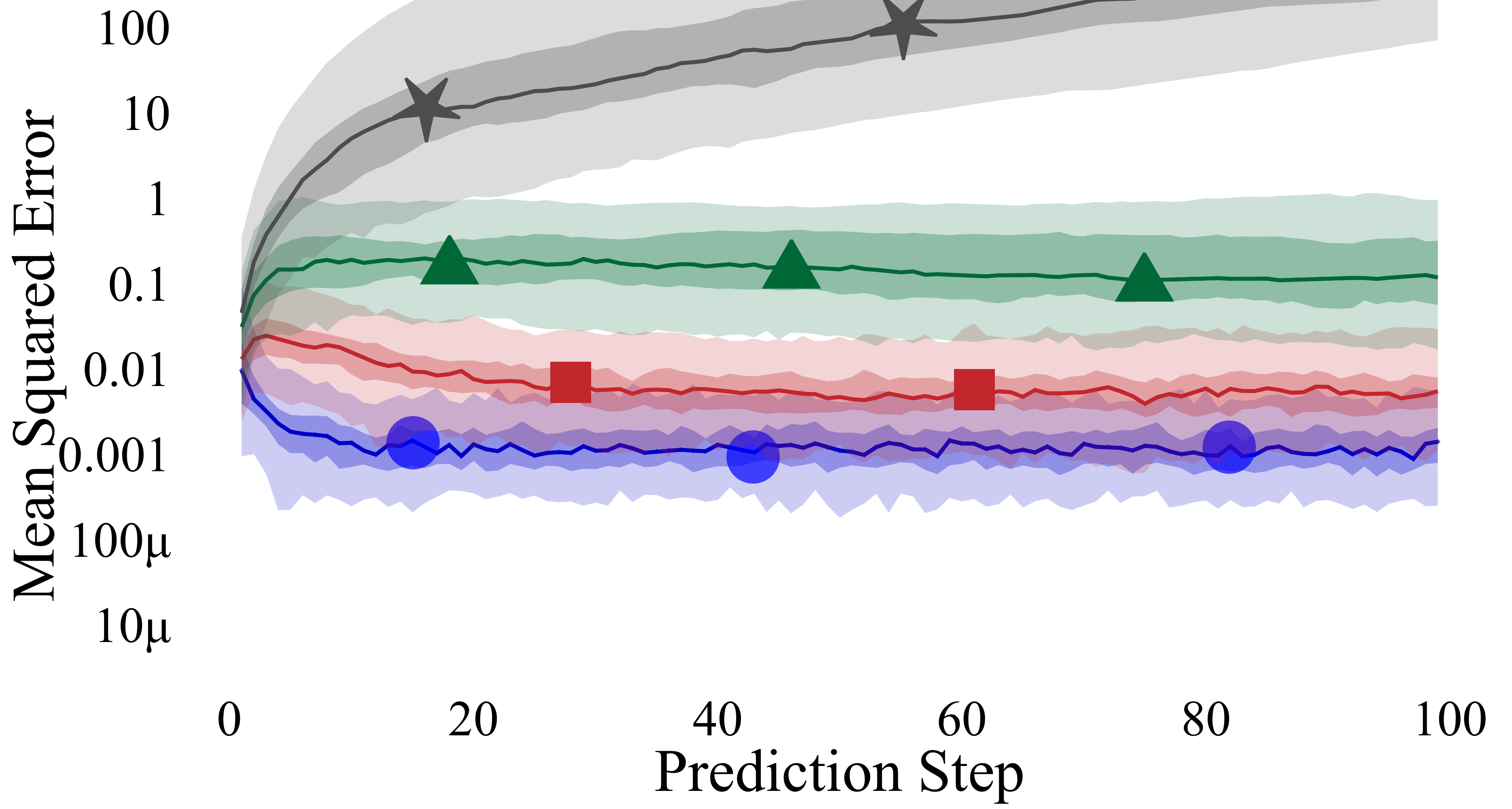}
        }
        \hfill
        \subfigure[$10\times$ noise.]{
        \includegraphics[width=0.315\linewidth]{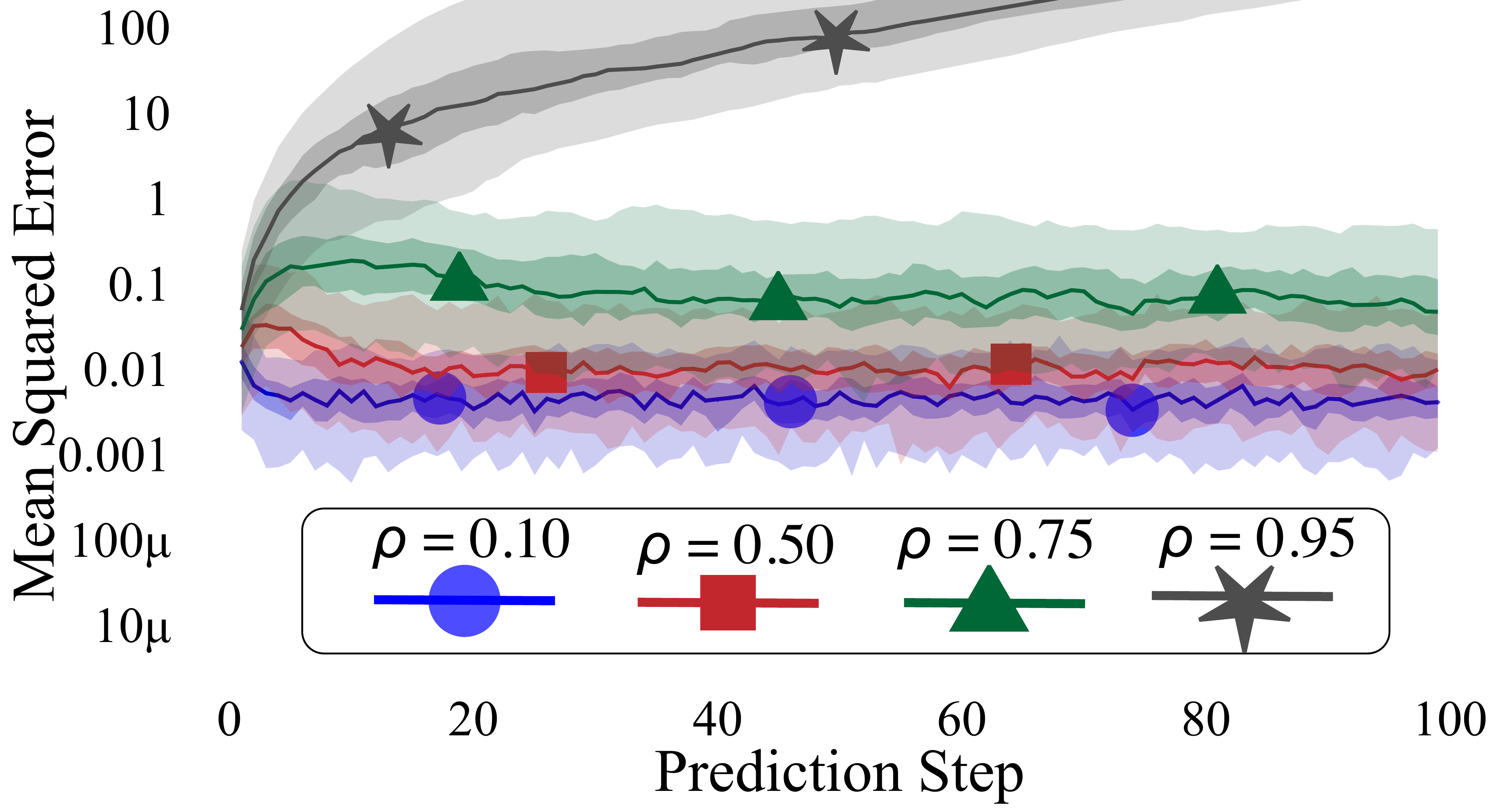}
        }
        \hfill
        \subfigure[$100\times$ noise.]{
        \includegraphics[width=0.315\linewidth]{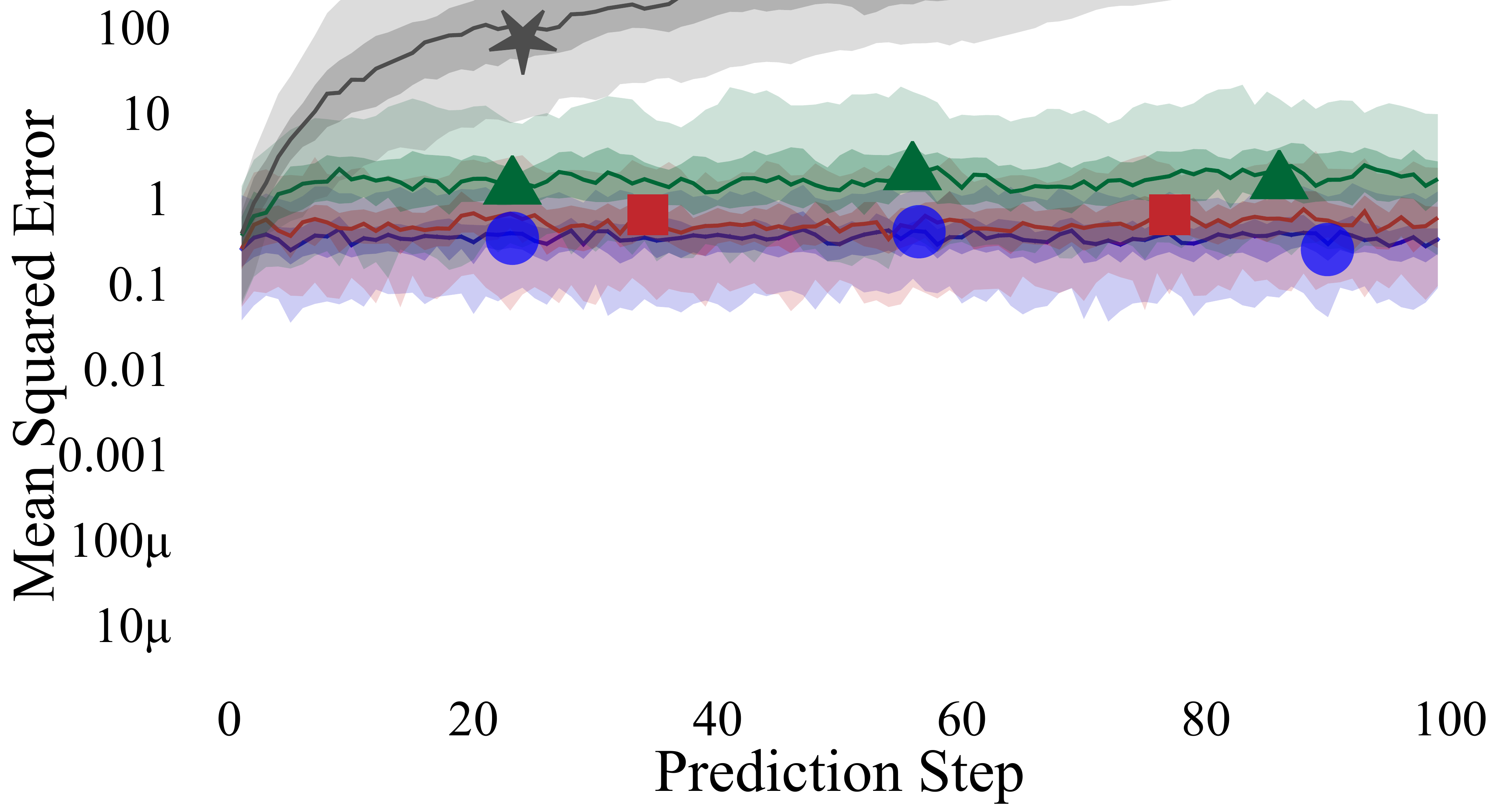}
}
    \else
        \begin{subfigure}[t]{0.32\linewidth}
            \centering
            \includegraphics[width=\linewidth]{figures/noise_00_combine.pdf}
            \caption{\centering $0\times$ noise.
            }    
            \label{fig:comparep0.1}
        \end{subfigure}
        \hfill
        \begin{subfigure}[t]{0.32\linewidth}
            \centering
            \includegraphics[width=\linewidth]{figures/noise_01_combine.pdf}
            \caption{\centering $10\times$ noise.
            }    
            \label{fig:comparep0.75}
        \end{subfigure}
        \hfill
        \begin{subfigure}[t]{0.32\linewidth}  
            \centering 
            \includegraphics[width=\linewidth]{figures/noise_10_combine.pdf}
            \caption{\centering $100\times$ noise.
            }    
            \label{fig:comparep1.0}
        \end{subfigure}
    \fi
    \caption{
    Comparing prediction accuracy when increasing the levels of process noise in the system above and below the default of $\omega_t \sim \mathcal{U}(-0.01,0.01)$ on all dimensions (median, $65^\text{th}$, and $95^\text{th}$ percentiles).
    The error between a system with default (shown in \fig{fig:compound}) and zero process noise shows that the default noise from the random action matrices determines the resulting prediction accuracy.
    An interesting feature is that when increasing the process noise from $10\times$ to $15\times$, the modelling accuracy degrades by a factor of 15. 
    }
    \label{fig:compare_noise}
    \vspace{-10pt}
\end{figure}
\subsection{System: Process Noise}
\label{sec:noise}
The underlying noise %
within the dynamics has a substantial impact on measurement and evolution of any dynamical system.
The default state-space system has only process noise, sampled uniformly $\omega_i \sim \mathcal{U}(-0.01,0.01)$.
To measure the effects of this noise, we measure the prediction accuracy with the following multiples of the original noise: $0\times, 10\times, 100\times$.
The results, shown in \fig{fig:compare_noise} indicate that noise can control the maximum accuracy.
This floor is a primary contributor to model inaccuracy for stable poles $\rho= \{0.1,0.5,0.75\}$, but when the poles approach instability $\rho=0.95$, the compounding error is similar across all noise levels.

\ifx\arxivversion\undefined

\else
An interesting observation of the learning process is the relation between the random actions, which the networked is informed of, and that of the process noise. 
By default, the input matrices $\vec{B}$ are randomized along with the control policy in each trajectory, so they computationally impossible for the general dynamics model to learn.
In \fig{fig:compare_noise_noB}, the actions are zeroed so they no longer contribute as a disturbance on system dynamics, and the prediction accuracy shape is similar, but improves performance by about 100$\times$ when compared to \fig{fig:compound}.

\begin{wrapfigure}{r}{0.55\textwidth}
    \centering
    \includegraphics[width=\linewidth]{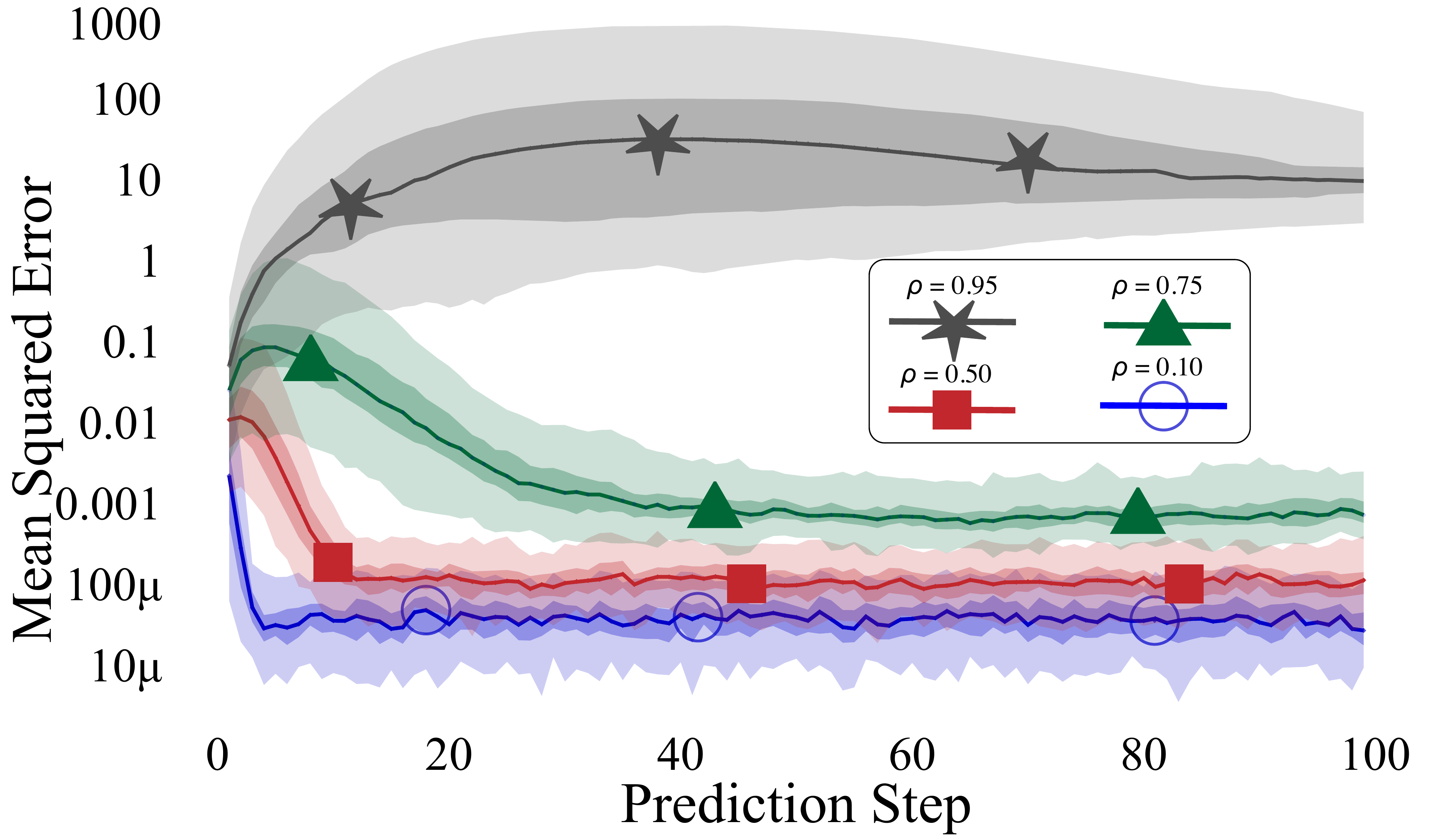}
    \vspace{-20pt}
    \caption{ 
    Showing how the randomly sampled input matrices, $\vec{B}$, and actions affect the per-step MSE with different eigenvalues (median, $65^\text{th}$, and $95^\text{th}$ percentiles) by collecting new data and evaluate newly trained models with $\vec{B}=0$.
    The random actions are the second leading cause of prediction error behind the unstable eigenvalues.}
    \label{fig:compare_noise_noB}
    \vspace{-15pt}
\end{wrapfigure}
\fi

For the robotic tasks shown in \fig{fig:examples-compare}, the default noise varies dramatically.
The Reacher has an unreported and low noise level (hidden within the Mujoco simulator), the Cartpole has uniform noise $\mathcal{U}(-0.1,0.1)$ on all states, and the  Quadrotor has a noise sampled from $\mathcal{N}(0,0.0001)$.
None of these simulators vary the level of noise relative to the type of the state variables (for example, a position in meters can take much lower magnitudes than an angle in degrees).
The learned models on the quadrotor system converged to substantially lower levels, potentially indicating that deep models continue to improve with substantial noise reduction.

\ifx\arxivversion\undefined

\else
\begin{remark}
Process noise impacts the peak prediction accuracy of a one-step dynamics model, but does not cause composed predictions to diverge more rapidly. 
\end{remark}
\fi
\begin{figure}[t]
    \centering
    \ifjmlrutilsmaths
        \subfigure[$d_s=9$.]{
        \includegraphics[width=0.3\linewidth]{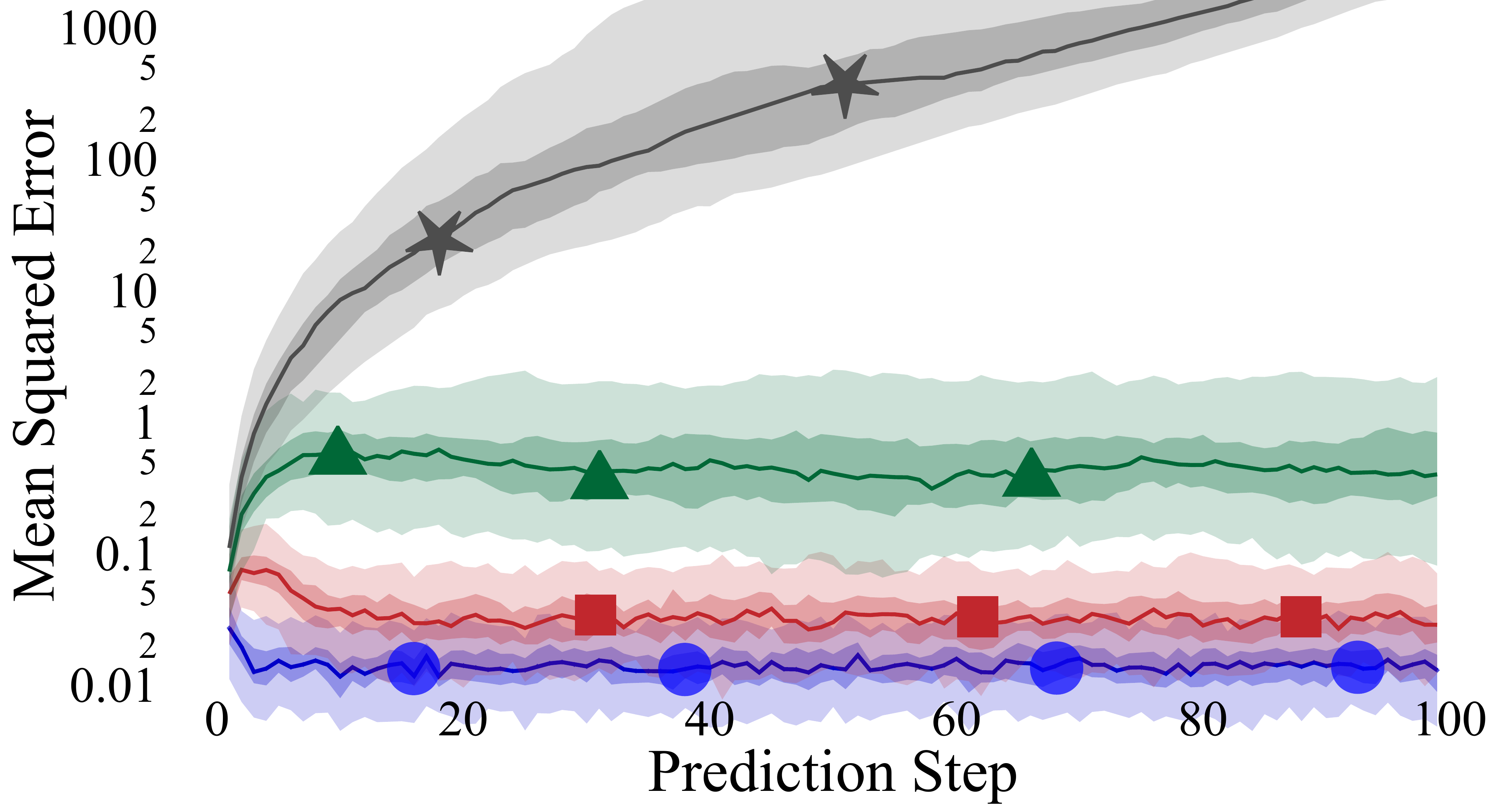}
        }
        \hfill
        \subfigure[$d_s=27$.]{
        \includegraphics[width=0.3\linewidth]{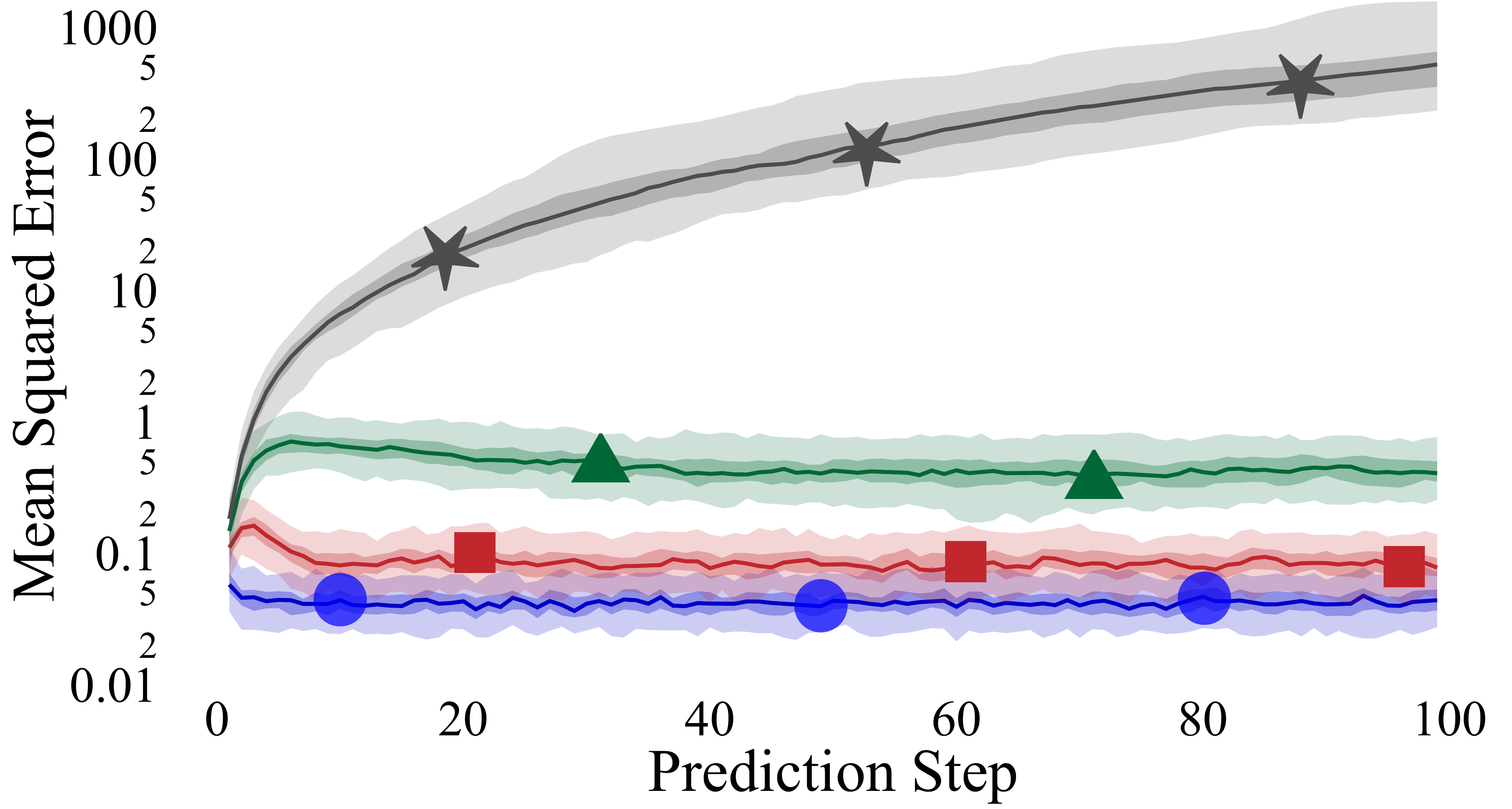}
        }
        \hfill
        \subfigure[$d_s=81$.]{
        \includegraphics[width=0.3\linewidth]{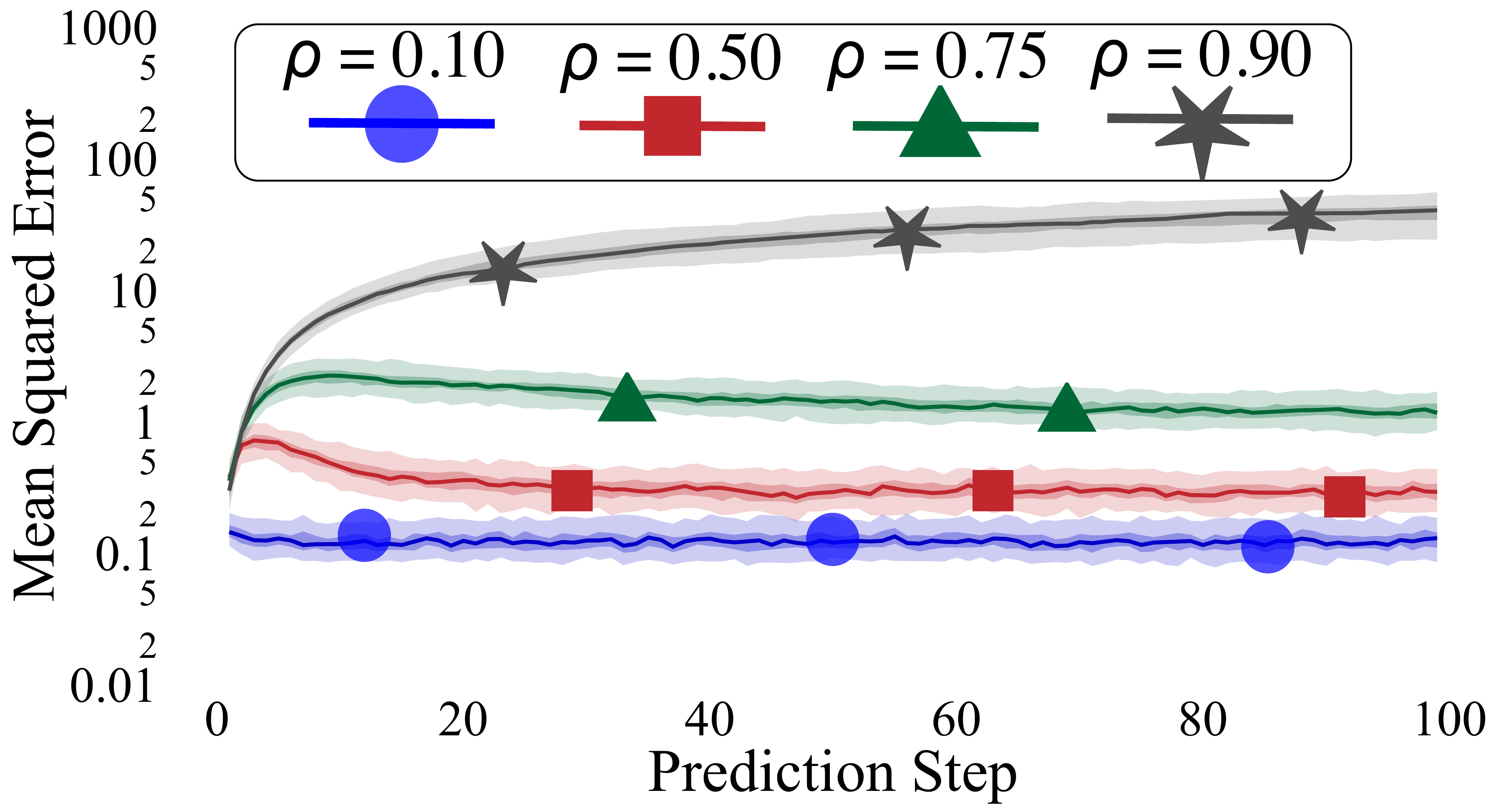}
        }
    \else
        \begin{subfigure}[t]{0.24\linewidth}  
            \centering 
            \includegraphics[width=\linewidth]{figures/p010/9dim_c.pdf}
            \caption{\centering Poles at $0.1$, $d_s=9$.}    
            \label{fig:comparep1.0}
        \end{subfigure}
        \hfill
        \begin{subfigure}[t]{0.24\linewidth}  
            \centering 
            \includegraphics[width=\linewidth]{figures/p050/9dim_c.pdf}
            \caption{\centering Poles at $0.5$, $d_s=9$.}    
            \label{fig:comparep1.0}
        \end{subfigure}
        \hfill
        \begin{subfigure}[t]{0.24\linewidth}  
            \centering 
            \includegraphics[width=\linewidth]{figures/p075/9dim_c.pdf}
            \caption{\centering Poles at $0.75$, $d_s=9$.}    
            \label{fig:comparep1.0}
        \end{subfigure}
        \hfill
        \begin{subfigure}[t]{0.24\linewidth}  
            \centering 
            \includegraphics[width=\linewidth]{figures/p090/9dim_c.pdf}
            \caption{\centering Poles at $0.90$, $d_s=9$.}    
            \label{fig:comparep1.0}
        \end{subfigure}
        \\
        \begin{subfigure}[t]{0.24\linewidth}  
            \centering 
            \includegraphics[width=\linewidth]{figures/p010/27dim.pdf}
            \caption{\centering Poles at $0.10$, $d_s=27$.}    
            \label{fig:comparep1.0}
        \end{subfigure}
        \hfill
        \begin{subfigure}[t]{0.24\linewidth}  
            \centering 
            \includegraphics[width=\linewidth]{figures/p050/27dim.pdf}
            \caption{\centering Poles at $0.50$, $d_s=27$.}    
            \label{fig:comparep1.0}
        \end{subfigure}
        \hfill
        \begin{subfigure}[t]{0.24\linewidth}  
            \centering 
            \includegraphics[width=\linewidth]{figures/p075/27dim.pdf}
            \caption{\centering Poles at $0.75$, $d_s=27$.}    
            \label{fig:comparep1.0}
        \end{subfigure}
        \hfill
        \begin{subfigure}[t]{0.24\linewidth}  
            \centering 
            \includegraphics[width=\linewidth]{figures/p090/27dim.pdf}
            \caption{\centering Poles at $0.90$, $d_s=27$.}    
            \label{fig:comparep1.0}
        \end{subfigure}
        \\
        \begin{subfigure}[t]{0.24\linewidth}  
            \centering 
            \includegraphics[width=\linewidth]{figures/p010/81dim.pdf}
            \caption{\centering Poles at $0.10$, $d_s=81$.}    
            \label{fig:comparep1.0}
        \end{subfigure}
        \hfill
        \begin{subfigure}[t]{0.24\linewidth}  
            \centering 
            \includegraphics[width=\linewidth]{figures/p050/81dim.pdf}
            \caption{\centering Poles at $0.50$, $d_s=81$.}    
            \label{fig:comparep1.0}
        \end{subfigure}
        \hfill
        \begin{subfigure}[t]{0.24\linewidth}  
            \centering 
            \includegraphics[width=\linewidth]{figures/p075/81dim.pdf}
            \caption{\centering Poles at $0.75$, $d_s=81$.}    
            \label{fig:comparep1.0}
        \end{subfigure}
        \hfill
        \begin{subfigure}[t]{0.24\linewidth}  
            \centering 
            \includegraphics[width=\linewidth]{figures/p090/81dim.pdf}
            \caption{\centering Poles at $0.90$, $d_s=81$.}    
            \label{fig:comparep1.0}
        \end{subfigure}
    \fi
\caption{Comparing compounding error with different state dimensions to see if input-output prediction size challenges the models when the underlying dynamics are regularized (shown is the MSE with median, $65^\text{th}$, and $95^\text{th}$ percentiles).
State size does not have a substantial effect on the modeling error (the decreased variance of the error at each step in the state-sizes could be due to averaging over more state dimensions).
    }

    \label{fig:state_dim}
\end{figure}
\begin{wrapfigure}{r}{0.5\textwidth}
    \centering
    \includegraphics[width=\linewidth]{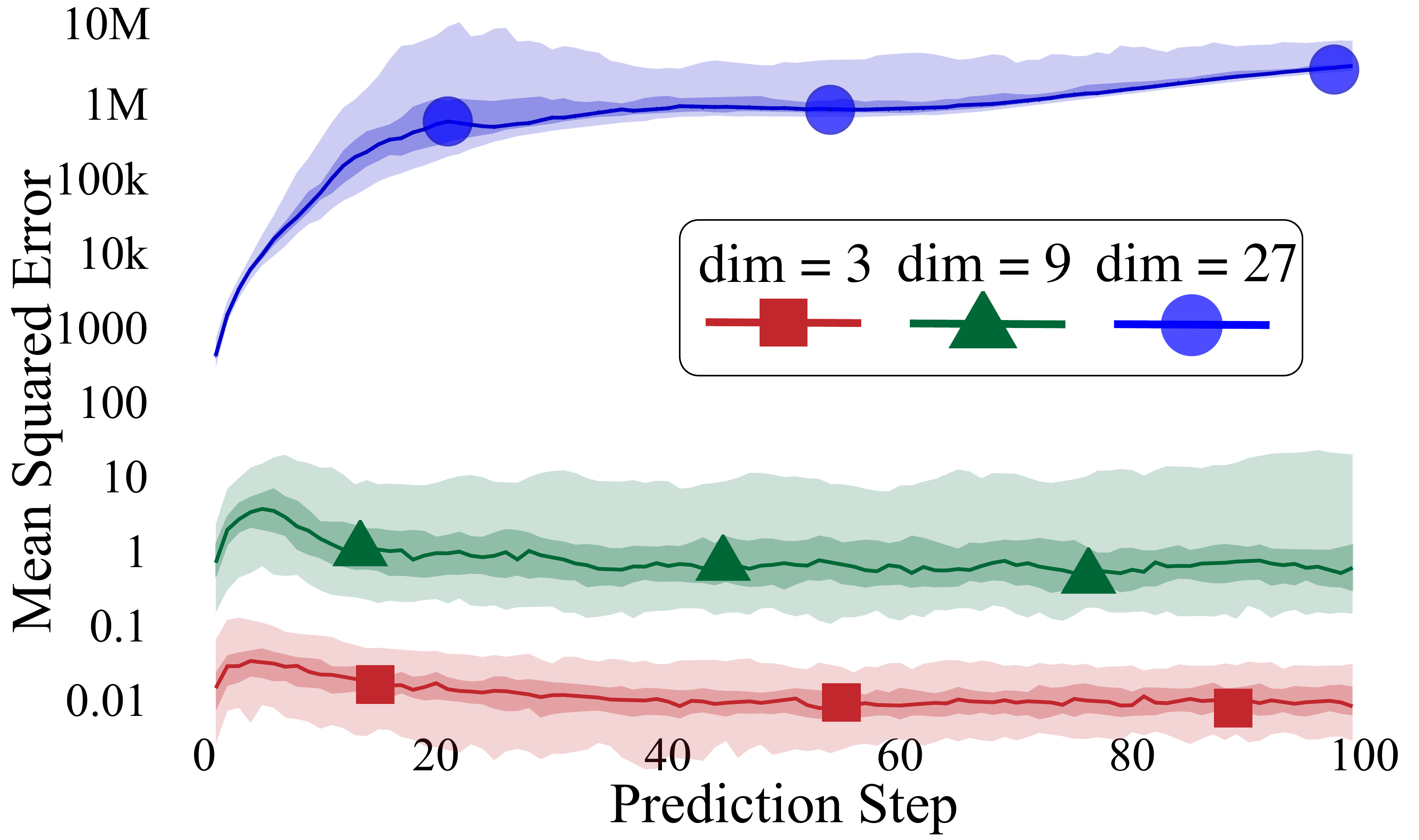}
    \vspace{-25pt}
    \caption{The prediction accuracy versus \textit{unregularized} state-dimension growth given a set pole at $\rho=0.5$ for all systems predictions.
    Specifically, in this figure the matrix norm, $\|\vec{A}\|_\infty$, of the state-dynamics grows with dimension.
    }
    \label{fig:dim_diverge}
    \vspace{-20pt}
\end{wrapfigure}
\subsection{System: State-space Dimension}
\label{sec:dim}
A large motivation to using deep models for predicting dynamics is the ability to extend to higher dimensional tasks.
While early work has shown that deep networks are useful for high dimensional tasks (such as \citep{nagabandi2019deep,lambert2019low} with state-action spaces over 40 dimensional), given the difficulty of comparing across different systems more controlled studies of prediction dimensionality are needed.
Learning one-step models scale the input and output dimensions with respect to the environment state.
To evaluate this, we scale the dimensionality of the state-space system from 3 to 9, 27, and 81, which is shown in \fig{fig:dim_diverge}.
The effect of increasing the state without normalizing the underlying dynamics, $\vec{A}$, is a rapid increase in the compounding error because the matrix norm grows with state-dimension. 

The underlying trajectories act more unstable when increasing the dimension of the state-space system because each state is a weighted sum of the current states.
Without normalization the summation representing a linear transition continues to grow with state dimension.
As a second experiment, the maximum row norm of the state-dynamics matrix, $\vec{A}_\infty$, is bounded to isolate the effects of prediction-dimension from the strong effects of system stability.
As the dimension increases in this subsection, the dynamics are regularized so that $\|\vec{A}\|_\infty \leq 3$, as in the default system.
With such normalization, the standard model types suffer from an increase in baseline prediction error, but the rate of error compounding does not grow, shown in \fig{fig:state_dim}.
Increasing the dimension of the state also reduces the variance of the compounding MSE, which is likely due to averaging over more states rather than a change in prediction dynamics.

\ifx\arxivversion\undefined

\else
\begin{remark}
Increasing the dimension of the system state while maintaining a similar underlying dynamics does not contribute a notable increase to compounding prediction error.
\end{remark}

\begin{wrapfigure}{r}{0.5\textwidth}
    \vspace{-20pt}
    \centering
    \includegraphics[width=\linewidth]{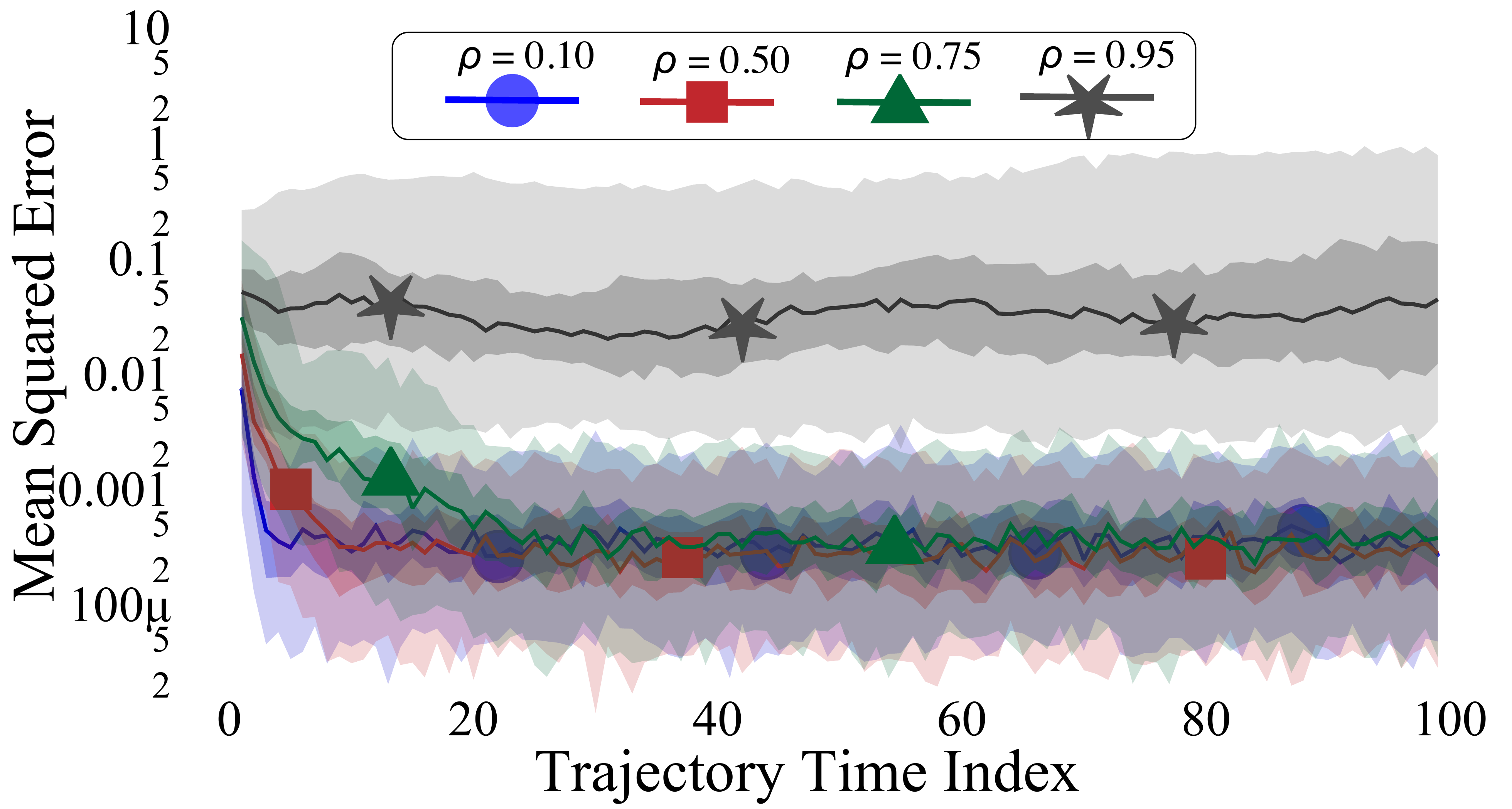}
    \vspace{-10pt}
    \caption{The per time index error for varying state-space systems.
    Again, unexpectedly, the error does not worsen further into the trajectories where state-space coverage is less dense.
    }
    \label{fig:tps-ss}
    \vspace{-25pt}
\end{wrapfigure}
\fi
\subsection{System: Data Distribution \& Density}
\label{sec:tps}
Understanding how model accuracy relates to an underlying training distribution is crucial to advancements in deep learning. 
Scaling the state-dimension of a system effectively reduces the density of data. 
Another axis for comparing the density of training data for a learned model is to compare model accuracy along trajectories understanding the relative density with respect to time.
For stable systems, the data will likely be more dense at higher time indices (as is the case for the Cartpole and the Crazyflie), but for other systems the data-distribution over time in trajectories can take complex forms.
As a proxy to density, we observe the per-step prediction error when predicting from the true state and action to the next state along each trajectory from the initial state $\mdpstate_0$ and at each intermediate state $\mdpstate_t$ (rather than the composed predictions as in most of this work). 
\begin{figure}
    \centering
    \ifjmlrutilsmaths
        \subfigure[Cartpole.]{
        \includegraphics[width=0.3\linewidth]{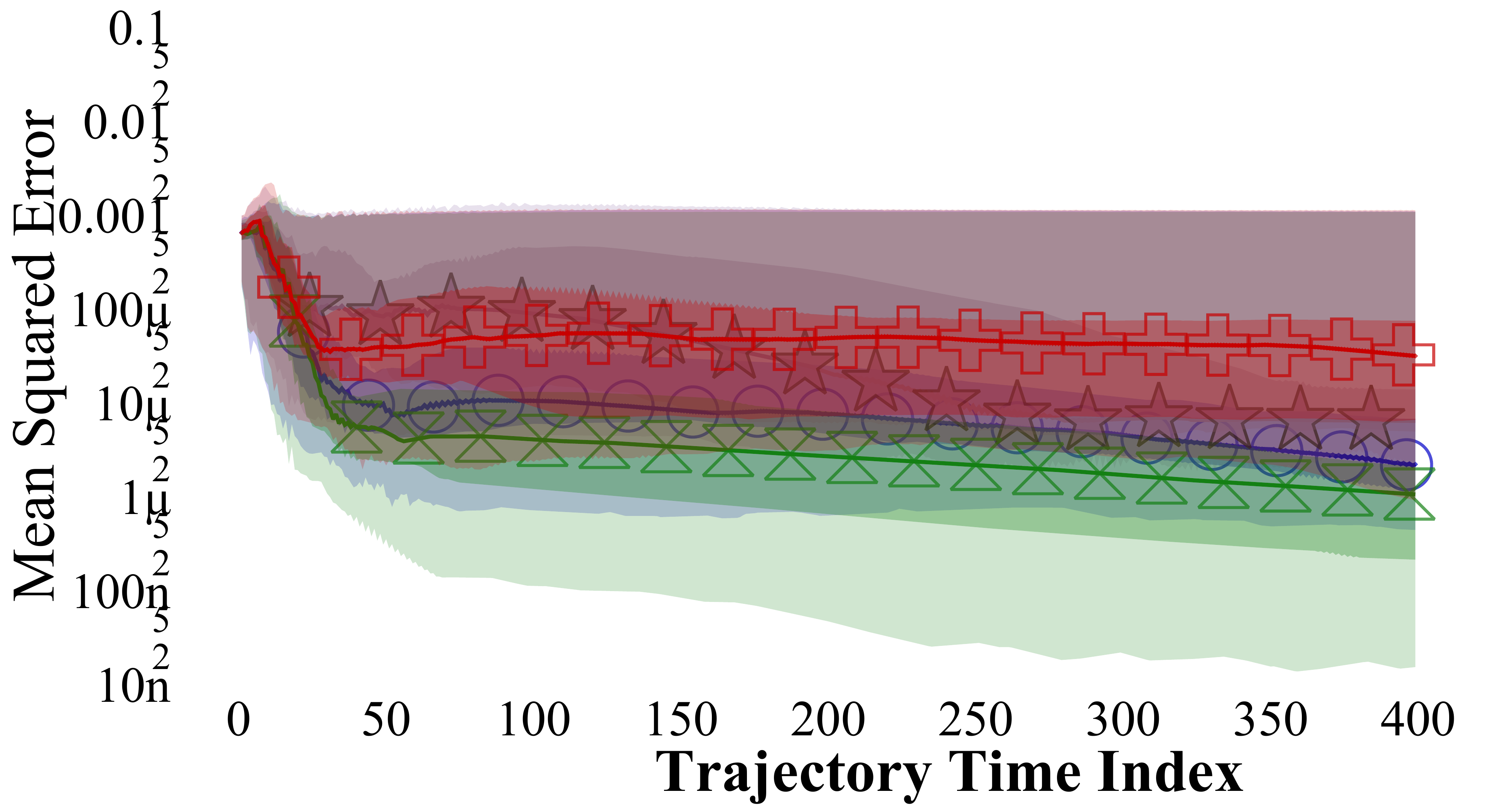}
        }
        \hfill
        \subfigure[Crazyflie.]{
        \includegraphics[width=0.3\linewidth]{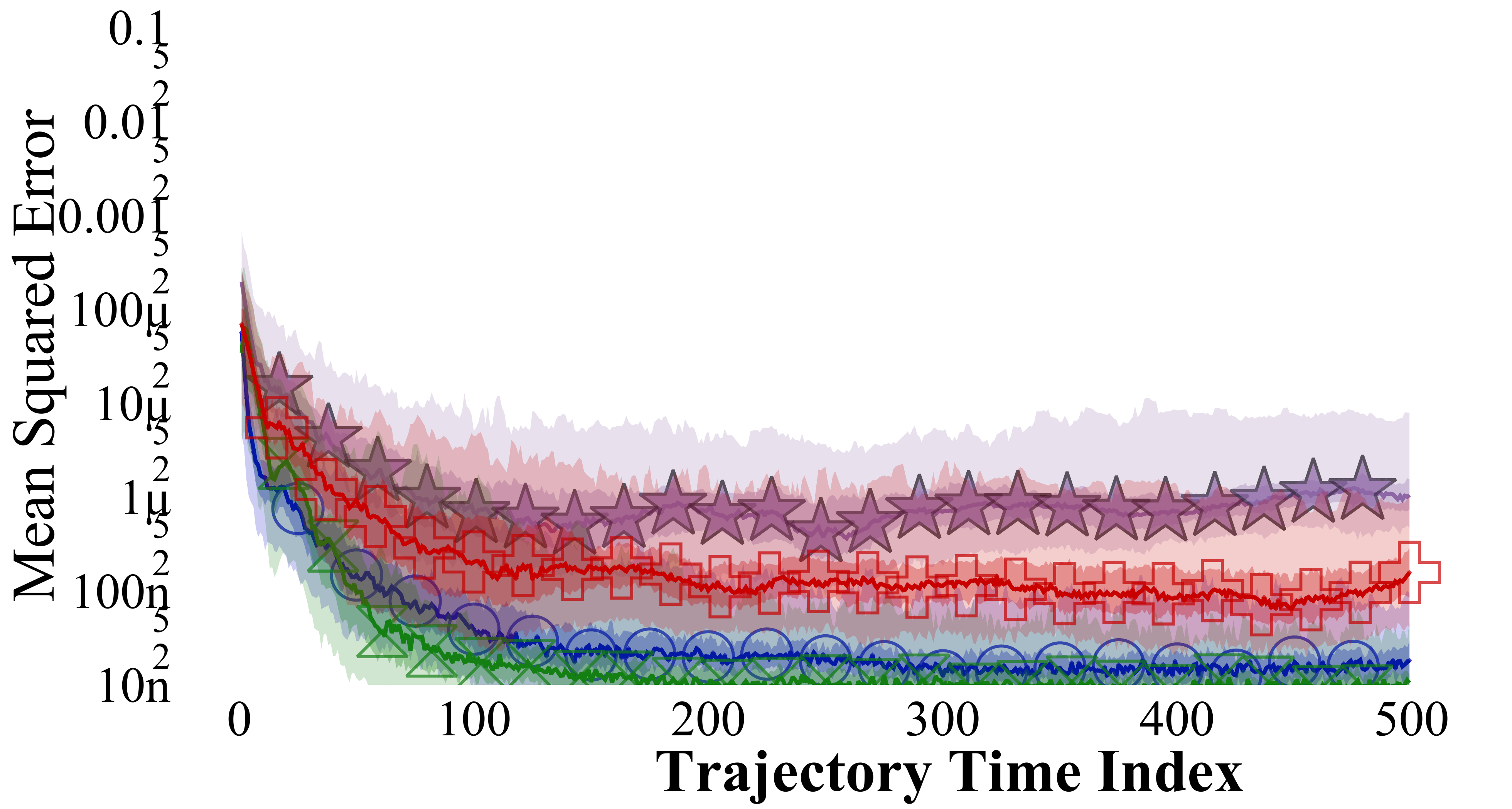}
        }
        \hfill
        \subfigure[Reacher.]{
        \includegraphics[width=0.3\linewidth]{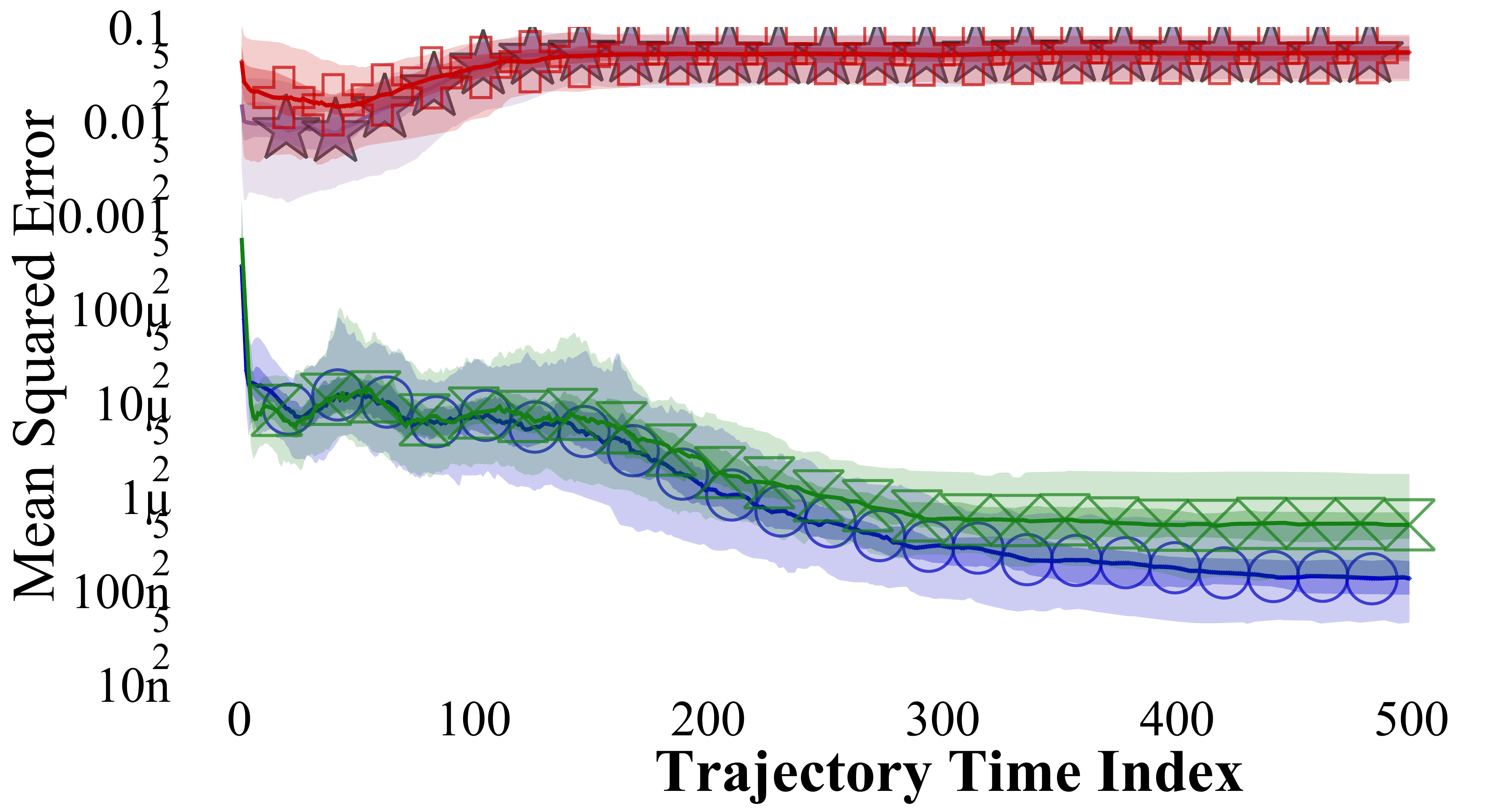}
        }
    \else
     \begin{subfigure}[t]{0.32\linewidth}  
        \centering 
        \includegraphics[width=\linewidth]{figures/traj-per-step/traj-per-step-cp.pdf}
        \caption{Cartpole.}  
        \label{fig:tps-cp}
    \end{subfigure}
    ~
    \begin{subfigure}[t]{0.32\linewidth}  
        \centering 
        \includegraphics[width=\linewidth]{figures/traj-per-step/traj-per-step-cf.pdf}
        \caption{Crazyflie.}    
        \label{fig:tps-cf}
    \end{subfigure}
    ~
    \begin{subfigure}[t]{0.32\linewidth}  
        \centering 
        \includegraphics[width=\linewidth]{figures/traj-per-step/traj-per-step-rch.pdf}
        \caption{Reacher.}  
        \label{fig:tbs-rch}
    \end{subfigure}
    \fi
    \input{figure_latex/legend}
    \caption{
The per-step model prediction error, rather than accumulated prediction error, (median, $65^\text{th}$, and $95^\text{th}$ percentiles) across trajectories in simulated robotic experiments.
This highlights the relative error of a true state-action pair at a given time index in a trajectory.
In these examples, the per-step error decreases as the controllers stabilize the robots towards the stationary target points.
}
    \label{fig:tps}
\end{figure}
\ifx\arxivversion\undefined
The results of the across-time one-step errors are shown in \fig{fig:tps} for the simulated robots, where the error-per step mirrors the relative density of the trajectories.
\else
The results of the across-time one-step errors are shown in \fig{fig:tps} for the simulated robots and \fig{fig:tps-ss} for the state-space system.
\fi

\ifx\arxivversion\undefined

\else
\begin{remark}
Dynamics model accuracy mirrors the distribution of training data, which is not necessarily correlated with the task of interest. 
\end{remark}
\fi

\begin{figure}
    \centering
    \ifjmlrutilsmaths
        \subfigure[Zero prediction model.]{
        \includegraphics[width=0.3\linewidth]{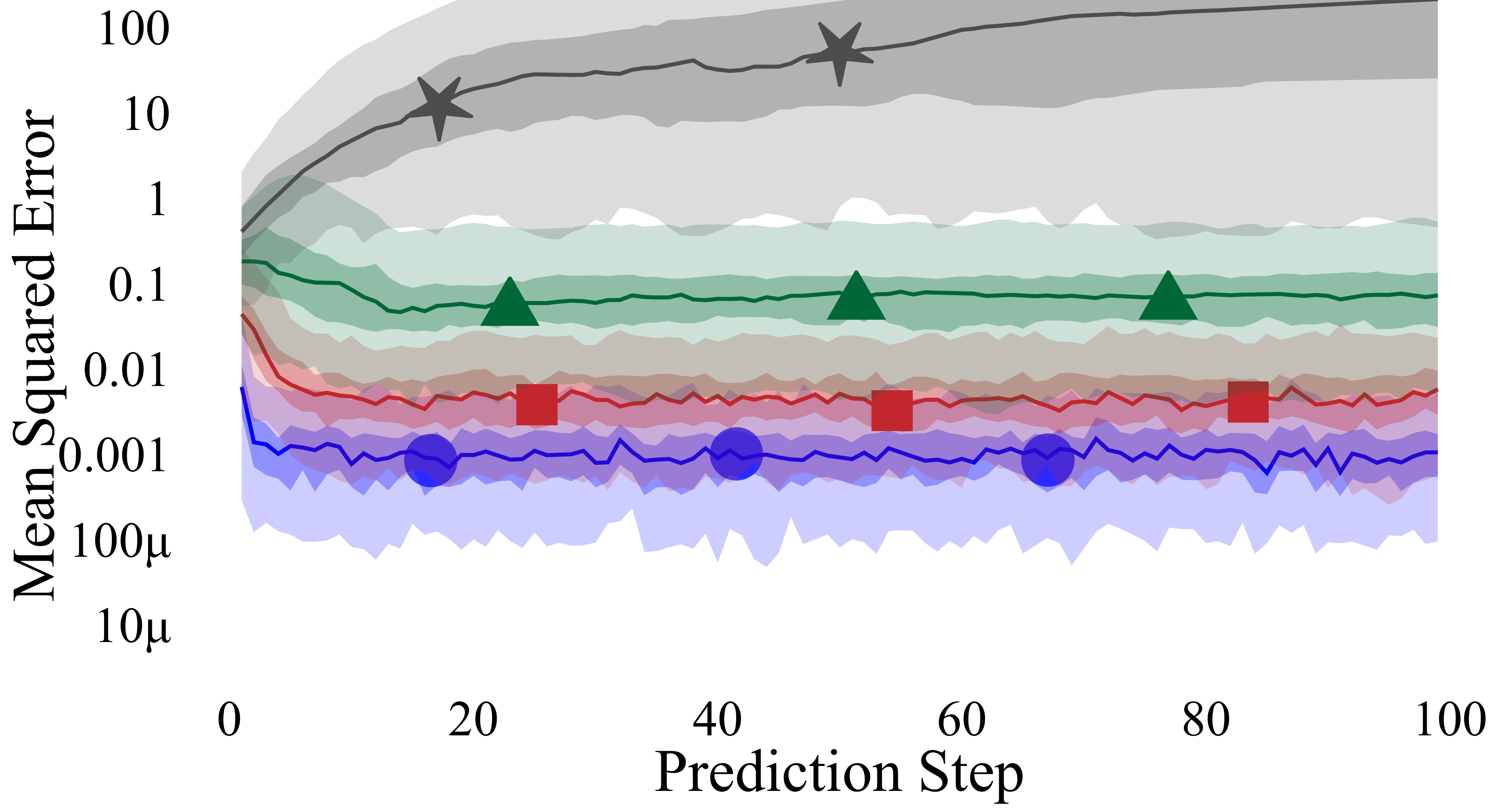}
        }
        \hfill
        \subfigure[Linear model.]{
        \includegraphics[width=0.3\linewidth]{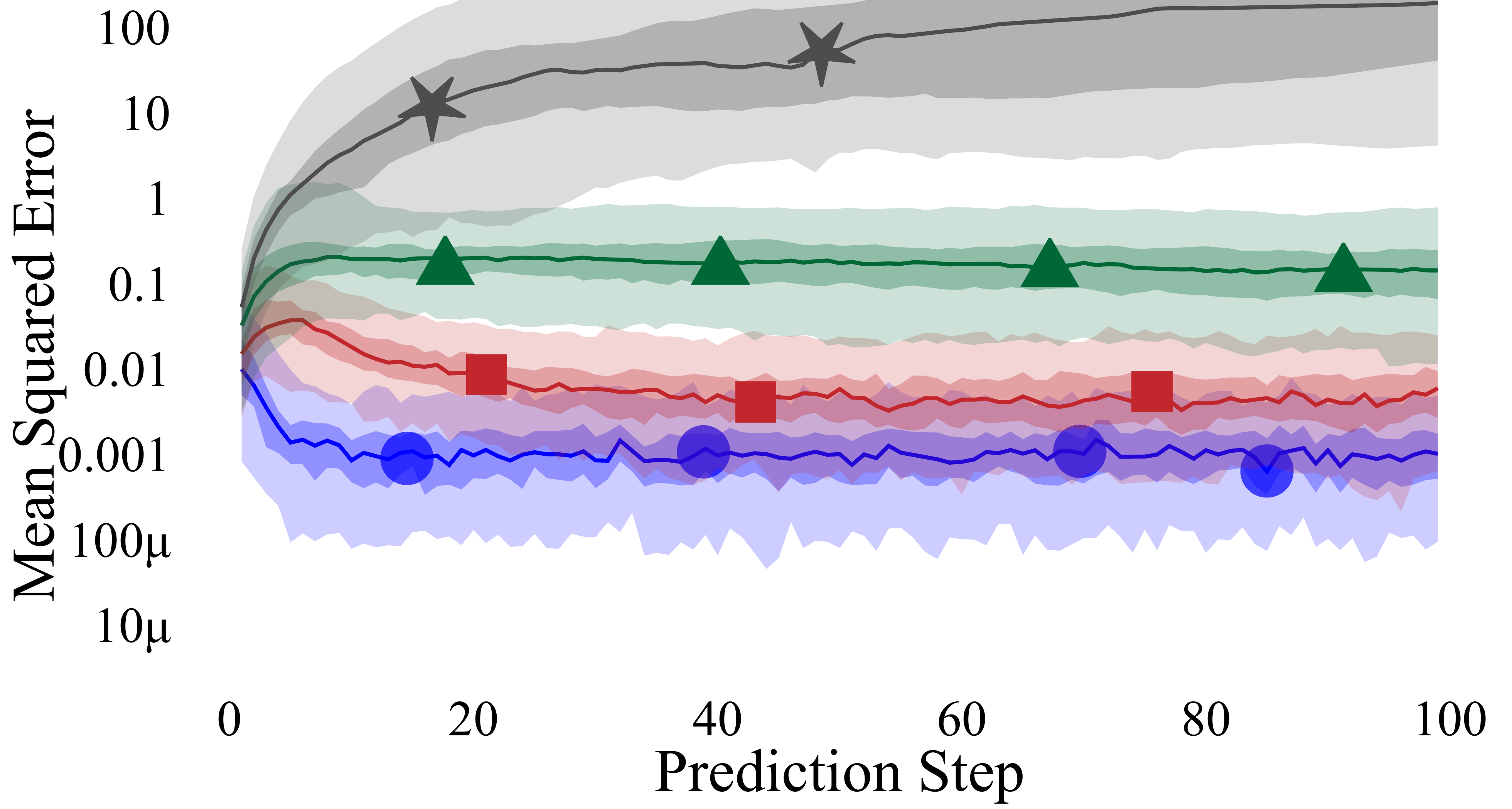}
        }
        \hfill
        \subfigure[Default Neural Network.]{
        \includegraphics[width=0.3\linewidth]{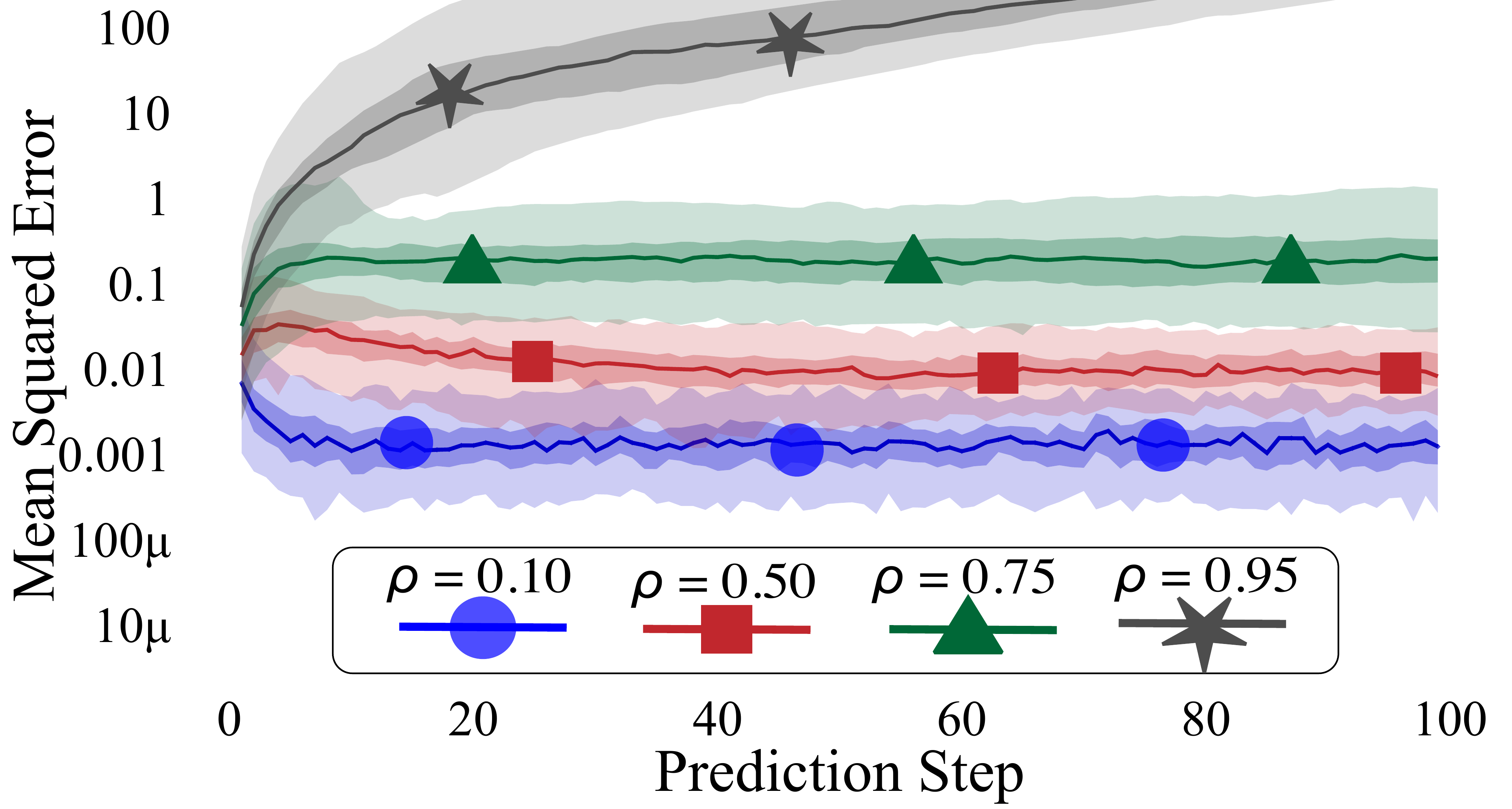}
    }
    \else
     \begin{subfigure}[t]{0.32\linewidth}  
        \centering 
        \includegraphics[width=\linewidth]{figures/simple_models/combine_zero.pdf}
        \caption{All predictions 0.}  
        \label{fig:simple-models-a}
    \end{subfigure}
    ~
    \begin{subfigure}[t]{0.32\linewidth}  
        \centering 
        \includegraphics[width=\linewidth]{figures/simple_models/combine_linear.pdf}
        \caption{Linear model. }    
        \label{fig:simple-models-b}
    \end{subfigure}
    ~
    \begin{subfigure}[t]{0.32\linewidth}  
        \centering 
        \includegraphics[width=\linewidth]{figures/noise_combine.pdf}
        \caption{Default Neural Network.}  
        \label{fig:simple-models-c}
    \end{subfigure}
    \fi
    \caption{
    Comparing the compounding error of the state-space system with simple linear and zero prediction models (shown is the MSE, $65^\text{th}$, and $95^\text{th}$ percentiles per pole). 
    On the simple state-space system, the simple models perform comparably to the neural network, but this does not indicate they would be as useful for control. }
    \label{fig:simple-models}
    \vspace{-10pt}
\end{figure}

\subsection{Model: Prediction Formulation \& Training}
\label{sec:formulation}
Different model parametrizations, particularly ensembles and probabilistic loss functions, have been shown to improve the peak performance of MBRL algorithms~\citep{chua2018deep,janner2019trust}.
It is important to identify if these models are uniformly more accurate in predictions, or if their integration with controllers is important to the performance gains.
As additional model comparisons, we include two \textit{simple model} baselines often omitted in recent MBRL work: a linear model (LIN) and a model predicting 0 (ZERO) to provide context for the prediction errors presented. 
These simple models performances are highlighted in \fig{fig:simple-models} for the state-space system and in \fig{fig:examples-compare} for the other simulated environments.
These simple models are extremely strong baselines in terms of compounding error, but our work does not study their usefulness for control (e.g. the zero prediction model would be useless for control).

\begin{figure}
    \centering
    \ifjmlrutilsmaths
        \subfigure[\centering \textbf{Without} $\cos{\theta},\sin{\theta}$ transform. ]{
        \includegraphics[width=0.34\linewidth]{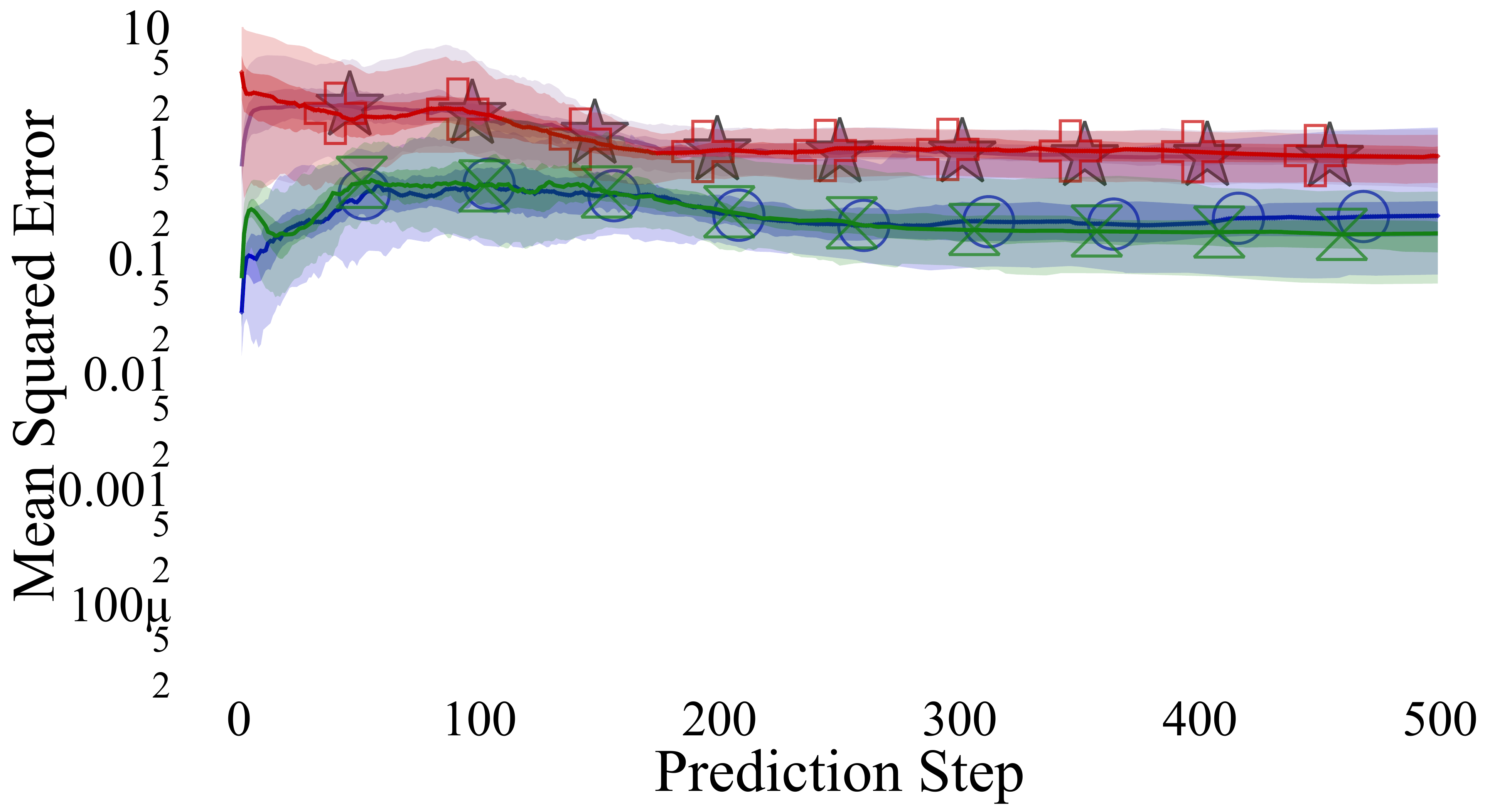}
        }
        \quad
        \subfigure[\centering \textbf{With} $\cos{\theta},\sin{\theta}$ transform. ]{
        \includegraphics[width=0.34\linewidth]{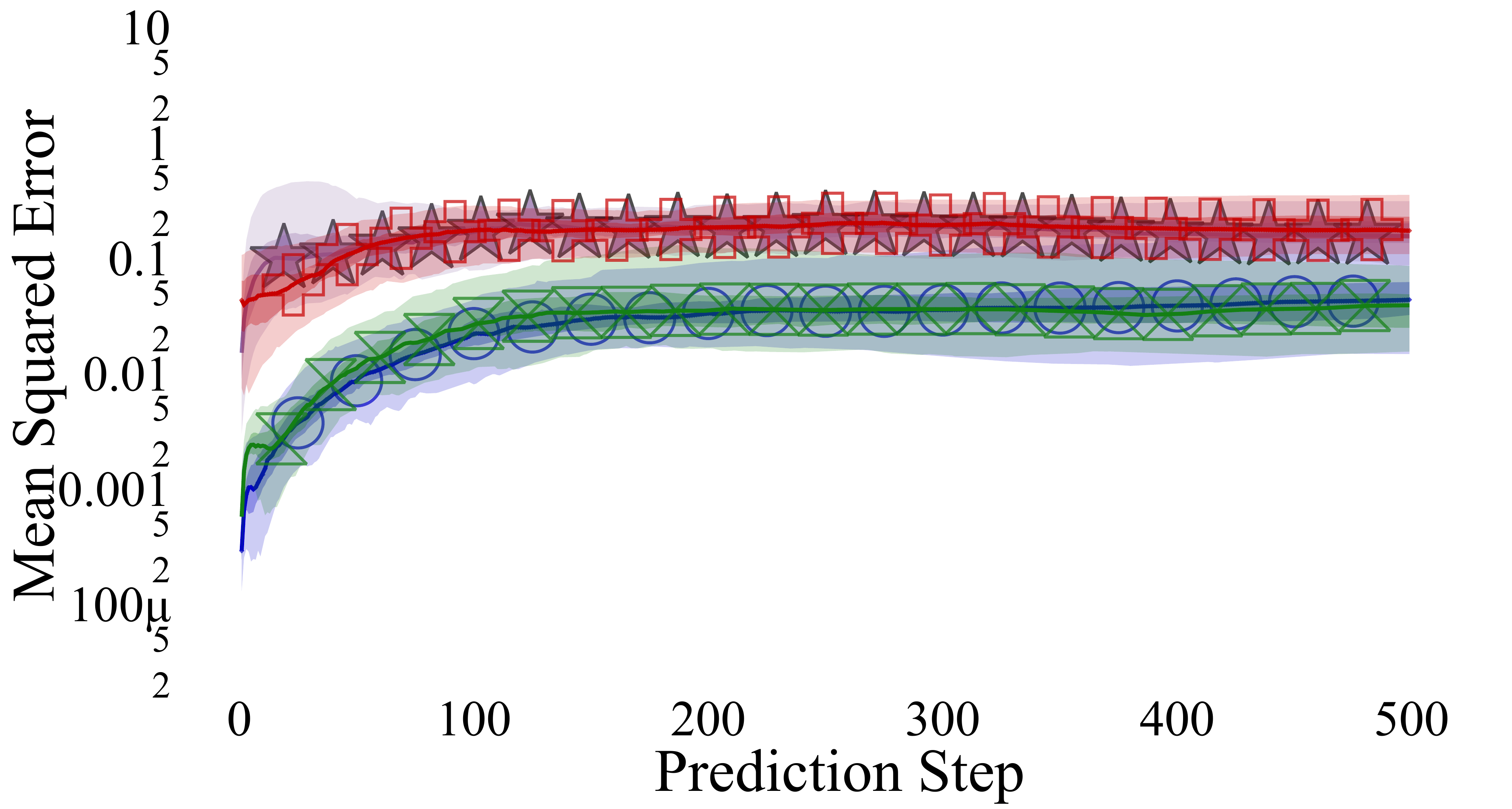}
        }
    \else
   \begin{subfigure}[t]{0.4\linewidth}
        \centering
        \includegraphics[width=\linewidth]{figures/deltaVtrue/reacher_nosincos.pdf}
        \caption{\centering Without sine \& cosine state processing. 
        }    
        \label{fig:reacherwith}
    \end{subfigure}
    \quad
    \begin{subfigure}[t]{0.4\linewidth}  
        \centering 
        \includegraphics[width=\linewidth]{figures/deltaVtrue/reacher.pdf}
        \caption{\centering With preprocessing. 
        }    
        \label{fig:reacherwithout}
    \end{subfigure}
    \fi
    \input{figure_latex/legend}
    \caption{
    Comparing the prediction accuracy (MSE median, $65^\text{th}$, and $95^\text{th}$ percentiles) on the Reacher environment with (\textit{right}) and without (\textit{left}) transforming the joint angles from radians $\theta_i$ to an expanded state for each joint $(\cos{\theta_i},\sin{\theta_i})$ to account for angle wrap-around the interval $[0,2\pi]$.
    The joint angle transformation, while increasing the state dimension from 10 to 15 improves the prediction accuracy substantially on short horizons and at convergence. %
      }
    \label{fig:sincos}
\end{figure}
Another popular modelling tool is shift from a true one-step prediction to that of a delta-state formulation.
For the simulated environments in \fig{fig:examples-compare}, there \textit{can} be an improvement by using the delta-state parametrization or an probabilistic ensemble, but it is not constant across all systems.
Importantly, especially when deploying on real systems, is that the ranking of prediction accuracy per-model is not consistent across environment.
Another implementation trick used in MBRL and other applications of model-learning for control is to map angles and other state-variables that may have discontinuities to smooth representations.
For example, with angles, the state-space can be expanded to be the sine and cosine of each angle, such as done by default in the Reacher environment. 
The increase in state dimension for smooth state-space improves the prediction accuracy notably at short horizons ($\hor <50$) and at convergence in \fig{fig:sincos} -- confirming results presented in \sect{sec:dim} that it is not a crucial factor for prediction accuracy when the underlying dynamics are constant.

We tested numerous other model types and neural network parametrizations (e.g. depth, layer size, parameter tuning, normalization, training set size, etc.) on the state-space system, but they had minimal effect on the prediction accuracy. 
The additional results can be found in the Appendices.

\ifx\arxivversion\undefined

\else
\begin{remark}
Different model parametrizations can have substantial impact on prediction accuracy, but more importantly the ranking of models is not maintained across environments, so multiple models should be validated on every new application. %
\end{remark}
\fi

\begin{figure}
    \centering
    \subfigure[ Crazyflie Quadrotor. %
        ]{
        \includegraphics[width=0.39\linewidth]{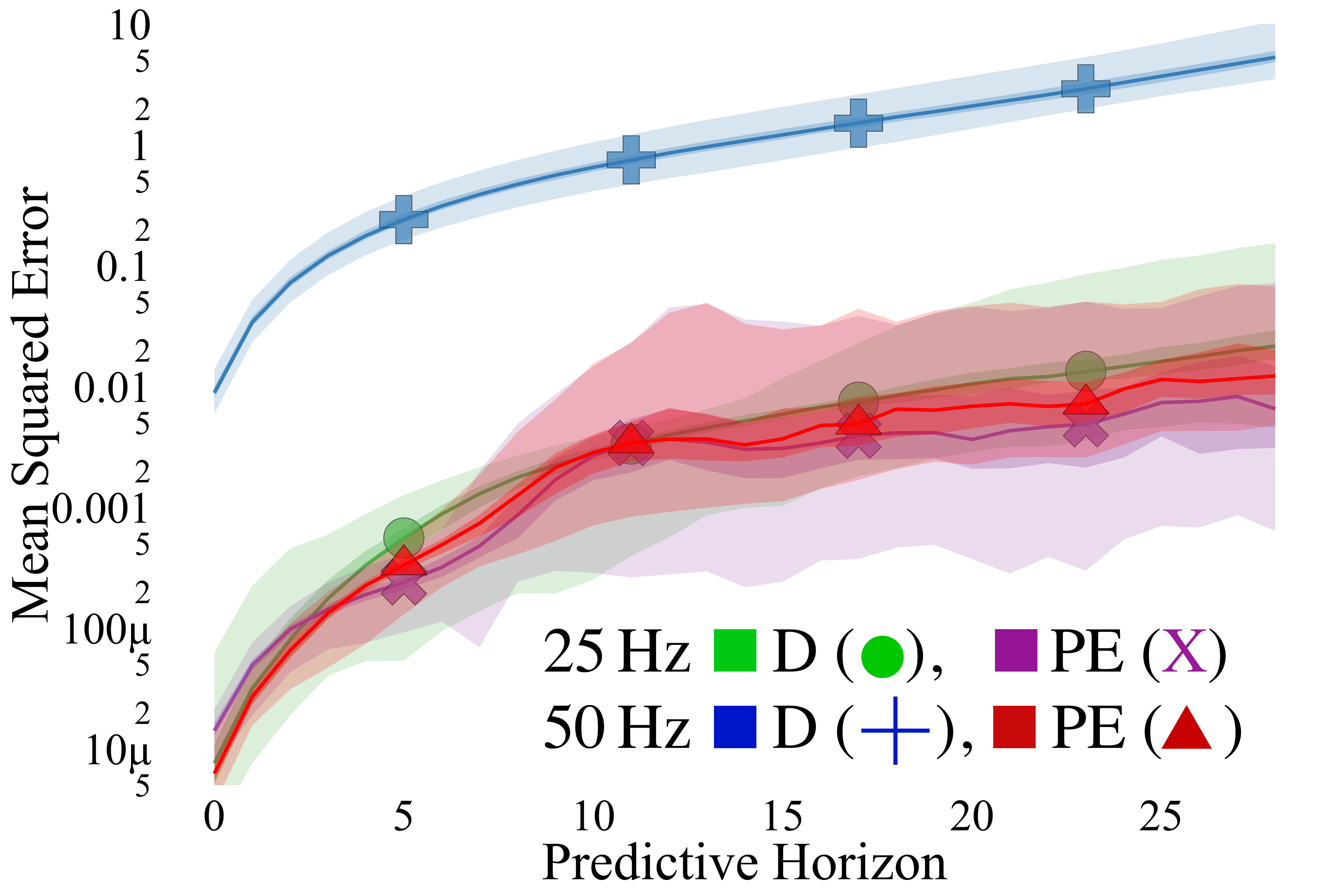}
        \label{fig:cf}
    }
    \quad
    \subfigure[Unitree A1 Quadruped. 
]{
    \includegraphics[width=0.39\linewidth]{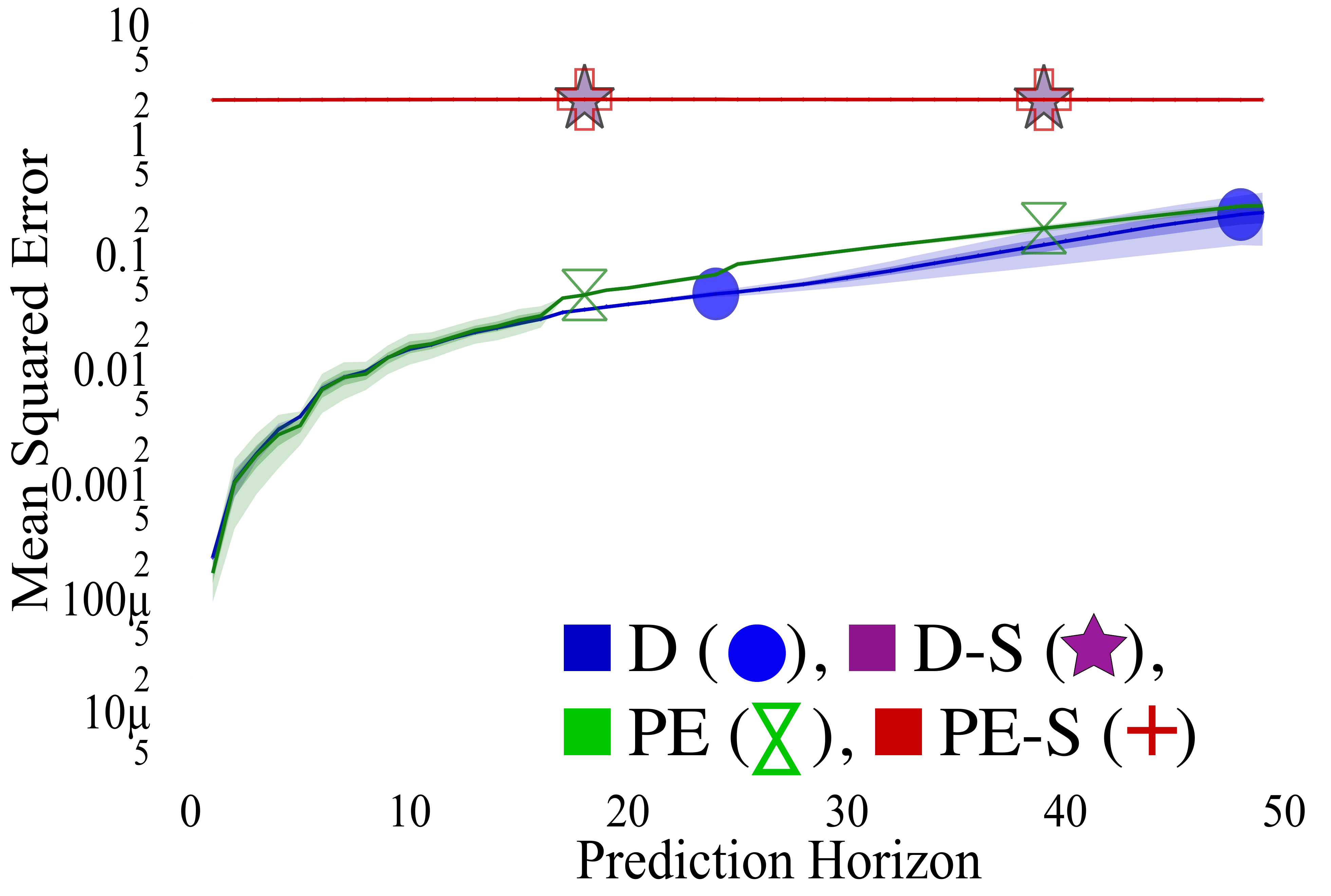}
    \label{fig:quadruped}
    }
    \caption{Looking at the real-world prediction divergence on a flying (a) and walking robot (b) 
    two training sets for a Euler angles of a flying robot (median, $65^\text{th}$, and $95^\text{th}$ percentiles).
    }
    \label{fig:realworld}
    \vspace{-10pt}
\end{figure}

\section{Case Studies of Compounding Error with Real World Data}
\label{sec:realworld}

In this section we showcase the prediction accuracy of models trained on real-world dynamics data from a quadrotor and a quadruped.
The quadrotor is a high-speed, high-noise system with datasets from two different control frequencies (showcasing the potential prediction challenge with low state-signal to noise ratio, which is studied further in the Appendix).
The quadruped shows how high-dimensional state-action spaces can reduce the prediction accuracy of true-state models.
\ifx\arxivversion\undefined
\else
\footnote{To contribute additional data to study compounding error on another platform, contact \href{mailto:nol@berkeley.edu}{\texttt{nol@berkeley.edu}}.}
\fi

\textbf{Quadrotor}: 
The Crazyflie is a micro-aerial vehicle that masses only \SI{27}{\gram}, is \SI{9}{\centi \meter^2} and performs on-board sensor fusion with an MPU-9250 inertial measurement unit ($\mdpstate\in\mathbb{R}^3$, $\mdpaction\in\mathbb{R}^4$).
The dataset used corresponds primarily to episodes of flights from 1 to \SI{5}{\second} attempting to stabilize the Euler angles of the robot (data is from~\cite{lambert2019low}).
Most of these sequences are unstable and end with failed control where the Euler angles diverge or the robot collided with a wall due to drift of unmeasured states.
The prediction accuracy is shown for a training set in \fig{fig:realworld}(a).
Note that a prediction of the same horizon in steps translates to a longer prediction in \textit{time} when the model is trained on data with a lower frequency.
The results show that there is a clear gain in prediction accuracy with the lower frequency. %

\textbf{Quadruped}: As another evaluation of compounding data, we have randomly created episodes of length 50 from a batch of 3800 points of state-action data from the Unitree A1 Quadruped ($a_t \in \mathbb{R}^{60}$, $s_t \in \mathbb{R}^{52}$).
This data comes from non-episodic data of the quadruped walking with a trot gait.
The action space is a multi-modal controller representing a unstudied problem in MBRL for control.%
When a leg corresponding to one of the 12 motors (3 per leg) is in contact with the ground, all actions are set to 0 except torque indicator, and vice-versa when the leg is in the air. %
The error shown when predicting a high-dimensional input-output relation is shown in \fig{fig:realworld}(b).

\ifx\arxivversion\undefined

\else

\section{Takeaways}
Here, we outline a few key observations that should be considered when understanding the compounding prediction error on a new system. 
These should be the points of focus when model-learning for control is applied on new systems:
\begin{itemize}
    \item \textbf{``\textit{No Free Lunch}'' Applies to Model Accuracy}: Given a fixed dataset, changing between different models will shift where error is present in the state-action space and over different predictive horizons. 
    There will be no model that is perfect for one task, so designers should match their model to the desired controller.
    \item \textbf{Dynamics Dominates the Model Accuracy}: 
    The properties of the dynamic system being modeled often has substantially greater effect on the prediction accuracy compared to model parametrization or training parameters. This point covers the accepted optimization procedures in the literature, but more complex optimizations, such as Automatic Machine Learning of dynamics models for MBRL~\citep{zhang2021importance}, is an exception that shifts modeling from an accuracy problem to one of maximization of reward.
    \item \textbf{Long-horizon Errors Can Level-off}: The results show that for many simulated applications, the error only compounds over an initial horizon~$\hor$ after which the error levels off or grows slowly. In most cases, this levelling happens when predictions are already useless for control purposes. However, we can postulate that might exist cases when the levelling off of error is sufficiently low that long-horizon predictions could be leveraged to solve sparse tasks.
    \item \textbf{Simple Models for Simple Systems}: 
    In our low-dimensional experiments, simple linear models and deterministic neural networks provide a strong baseline that should be considered in real-world applications.
    \item \textbf{Low-to-Zero Noise is not a Accuracy Guarantee}: 
    Transitioning from moderate to low to zero noise has diminishing returns on prediction accuracy, indicating that even simulated environments with no noise can still be difficult modelling tasks.
\end{itemize}

\section{Future Work}

 \paragraph{Other Prediction Modalities}
    Many other model types and trajectory propagation techniques exist that are well suited to a more narrow spectrum of problems than deep neural networks.
    Linear models are suited to linear systems~\citep{fu2016one, bansal2017goal}, Gaussian processes are useful for lower dimensionalities and dataset sizes~\citep{wang2006gaussian, Deisenroth2011PILCO, mckinnon2017learning}, trajectory-based neural networks are useful with closed form control laws~\citep{lambert2020learning}, and new physics-based neural networks are yet to be deployed for control~\citep{krishnapriyan2021characterizing,jiahao2021knowledge}. 
    When dynamics models learn distributions instead of specific transitions, the method for propagation of the imagined trajectory can heavily impact both compounding error and downstream control.
    In this work, only expectation-based propagation is used for probabilistic models and probabilistic ensembles.
    Crucially, careful understanding of the various model types' strengths and weaknesses with respect to compounding error will yield improved performance when paired with a suitable controller.

\paragraph{Dynamics Modeling with Distribution Shift}
    In model-based reinforcement learning, the data used to train the model changes with each step. 
    This can take two forms: the amount of data in the replay buffer and the relative shape of the data distribution.
    There are complicated relationships in model-based reinforcement learning between model accuracy, task-performance, and data-distribution that are not studied in this paper.
    Recent work suggests that optimizing solely for prediction accuracy does not result in maximum task-performance~\citep{lambert2020objective, zhang2021importance}.
    These data-properties are very difficult to quantify but crucial for performance -- for example, the relative density of labeled data points and the underlying difficulty of a transition to model both effect prediction accuracy and are not well understood. 

\fi

\section{Conclusion}
\label{sec:conclusion}
Accurately understanding the dynamics of a robotic system with finite data can enable many forms of control.
In this work we characterize the compounding error problem that emerges in long-horizon prediction with learned one-step, neural network models.
In hopes of the improvement of tools used to model these systems, we show that underlying system dynamics have a dramatic effect on the short- and long-term prediction accuracy for any system.
With this understanding, model-learning for control can design advanced controllers with a more precise relationship between the prediction and action-decision error.

\section*{Acknowledgements}
The authors would like to thank Tianyu Li for providing data from the Unitree A1 Quadruped and Howard Zhang for participating in useful discussions.

\clearpage

\ifjmlrutilsmaths
\else
    \bibliographystyle{unsrt}  
\fi
\bibliography{models}  %

\ifx\arxivversion\undefined
\else

\clearpage
\newpage

\appendix
\paragraph{\textbf{\Large{Appendices}}} 
\paragraph{}The Appendices contain the following content:
\begin{itemize}
\item \sect{sec:exp_details} contains additional environment (see \sect{sec:env_extra}) details and the model training parameters (see \sect{sec:training}).
\item \sect{sec:additional} contains additional experiments investigating compounding error. The additional experiments are broken up into two sections: 1) additional effects of model properties are discussed in \sect{sec:model_appndx} and 2) other impacts on and lenses for studying compounding error are illustrated in \sect{sec:other_causes}.

\end{itemize}
\section{Extra Experimental Details}
\label{sec:exp_details}
\subsection{Additional Environment Details}
\label{sec:env_extra}
\paragraph{State-space System}
For a discrete-time state-space system, the time evolution of the state over time can be solved for explicitly. 
This time evolution, shown in \eq{eq:time_evolution}, is the goal of what these composed one-step systems attempt to model, yet result in compounding errors.
The solution to the discrete transition dynamics takes the form an transient response from the initial state and a forced response corresponding to the applied action sequence:
\begin{equation}
    s_t = \underbrace{\vec{A}^t\vec{s}_0}_{\text{Transient}} + \underbrace{\sum_{l=0}^{k-1}A^{k-l-1}Ba(l)}_{\text{Forced Response}}
    \label{eq:time_evolution}
\end{equation}
For the state-space system, the time by which the prediction error converges to its minimum for a given pole is proportional to the number of steps by which the transient of the initial state will decay.
The transient is proportional to powers of the dynamics matrix, $\|A\|^k$, which is proportional to the powers of the eigenvalues.
For the poles in $\{0.01,0.05,0.1,0.25,0.5\}$ the number of time-steps until the steady state error is reached is proportional to the mean transient decay in \tab{tab:data}.
Example trajectories and one-step predictions are shown in \fig{fig:exampletraj}.
In this parametrization, the input at any time is randomly sampled, so the forced response becomes a source of noise.

\begin{wrapfigure}{r}{0.6\textwidth}
    \vspace{-10pt}
    \centering
    \ifjmlrutilsmaths
        \subfigure[Cartpole.]{
        \includegraphics[width=0.3\linewidth]{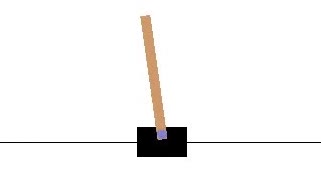}
        \label{fig:cp}}
        \hfill
        \subfigure[Reacher.]{
        \includegraphics[width=0.3\linewidth]{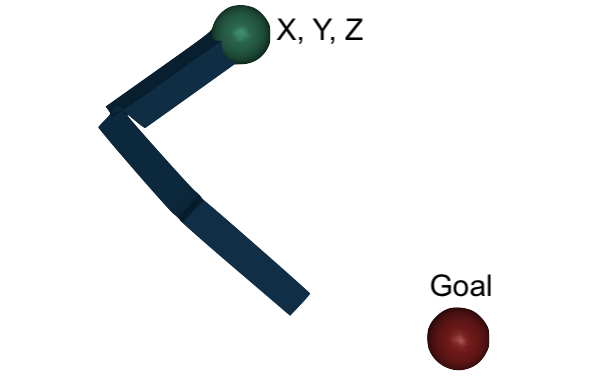}
        \label{fig:rch}}
        \hfill
        \subfigure[Quadrotor.]{
        \includegraphics[width=0.3\linewidth]{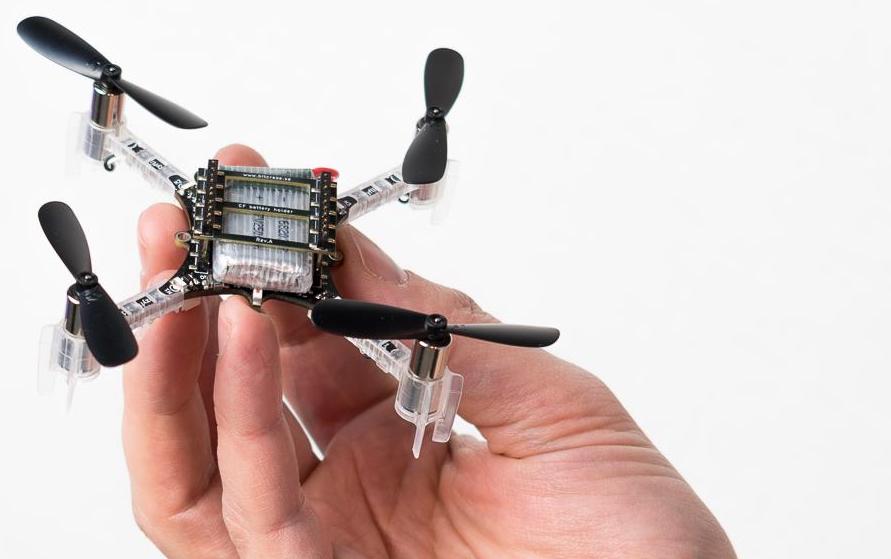}
        \label{fig:quad}}
        \hfill
    \else
        \begin{subfigure}[t]{0.32\linewidth}
            \centering
            \includegraphics[width=\linewidth]{figures/robots/cp.jpg}
            \caption{Cartpole.}    
            \label{fig:cp}
        \end{subfigure}
        \hfill
        \begin{subfigure}[t]{0.32\linewidth}  
            \centering 
            \includegraphics[width=\linewidth]{figures/robots/reacher_white.png}
            \caption{Reacher.}    
            \label{fig:rch}
        \end{subfigure}
        \hfill
        \begin{subfigure}[t]{0.32\linewidth}  
            \centering 
            \includegraphics[width=\linewidth]{figures/robots/cf-official.jpg}
            \caption{Quadrotor.}    
            \label{fig:quad}
        \end{subfigure}
    \fi
    \caption{Experimental platforms.}
    \label{fig:robots}
    \vspace{-15pt}
\end{wrapfigure}

\paragraph{Cartpole}
We evaluate predictions of state and reward of cartpole agents conditioned on a varied Linear Quadratic Regulator (LQR)~\citep{callier2012linear} (\fig{fig:examples-compare}). %

\paragraph{Quadrotor - Simulated}
The quadrotor model is based off the Crazyflie ~\citep{giernacki2017crazyflie} -- an \SI{27}{\gram}, open-source micro-aerial vehicle.
The 12 state Euler-step simulation follows \citep{mahony2012multirotor}. 
The simulated controller is a linear, pitch and roll Proportional-Derivative controller with randomly sampled parameters.

\paragraph{Reacher}
The task associated with the environment is to maneuver the end-effector of the arm from an initial position state to an end position state.
Our experiments control the agent using a Proportional-Integral-Derivative (PID) controller with randomly generated parameter vectors $\vec{K}\in \mathbb{R}^{15}$.

\paragraph{Quadrotor - Real World}
Due to the high noise on the accelerations measured by on-board sensors, we evaluate predicting a restricted state of Euler angles from direct motor voltages as:
\begin{equation}
    s_t = \begin{bmatrix}
  \text{Yaw:} \psi & \text{Pitch:} \theta & \text{Roll:}\phi
    \end{bmatrix},
    \quad 
    a_t = \begin{bmatrix}
    V_1 & V_2 & V_3 & V_4
    \end{bmatrix}
    \label{eq:cf_state}
\end{equation}
When discretizing dynamics, lower sample rates yield more unstable eigenvalues, but the system can also gain by having relatively lower signal-to-noise ratio on the measured states.

\paragraph{Quadruped - Real World}
The state, $s_t \in \mathbb{R}^{52}$, corresponds to the following:
\begin{equation}
    s_t = \begin{bmatrix}
    \vec{x} & \dot{\vec{x}}& \vec{\omega} & \vec{x}^f_k & \psi & \theta & \phi & \alpha_i & \dot{\alpha}_i & c^f_k
    \end{bmatrix}
    \label{eq:quadruped_state}
\end{equation}
Here, $\vec{x}\in \mathbb{R}^{3}$ is the position of the robot base, $\vec{\omega}\in\mathbb{R}^3$ is the angular rates of the robot base, $\vec{x}^f_k\in \mathbb{R}^{3}$ is the position of the kth foot, $\psi, \theta, \phi$ are the Euler angles, $\alpha_i$ is the joint angle for each of the 12 motors, and $c^f$ is an indicator if each foot is in contact with the ground.
The action, $a_t \in \mathbb{R}^{60}$, is a bimodal input, where for each of the 12 motors on the robot has 5 dimensional action space of desired joint position and velocities ($\alpha^*_i, \dot{\alpha}^*_i$), low-level PID control coefficients ($K^p_i, K^d_i$), and set torque ($\tau_i$):
\begin{equation}
    a_t = \begin{bmatrix}
    \alpha^*_i & \dot{\alpha}^*_i & K^p_i & K^d_i & \tau_i
    \end{bmatrix}
    \label{eq:quadruped_action}
\end{equation}

\begin{table}[t]
\centering
\begin{tabular}{ r | c c c }
\toprule %
& Transient decay & Delta-state labels & True-state labels \\
  Poles ($\rho$) & $k$ : $\|\vec{A}^k \vec{x}_0 \| < \num{1E-4}$ &  $\text{mean}(\|s_{t+1} - s_{t}\|) \pm \sigma(\cdot)$ & $\text{mean}(\|s_{t+1}\|)\pm \sigma(\cdot)$ \\
  \midrule%
 0.01 & 4.7  & $\num{0.019}\pm\num{0.063}$ & $\num{0.011}\pm\num{0.021}$\\
 0.05  & 6.1 & $\num{0.020}\pm\num{0.066}$ & $\num{0.012}\pm\num{0.020}$\\
  0.10 & 7.4  & $\num{0.018}\pm\num{0.052}$ & $\num{0.012}\pm\num{0.026}$\\
 0.25 & 11.2 & $\num{0.017}\pm\num{0.044}$ & $\num{0.016}\pm\num{0.048}$\\
 0.50 & 21.8  & $\num{0.020}\pm\num{0.042}$ & $\num{0.033}\pm\num{0.087}$\\
  0.75 & 55.5& $\num{0.029}\pm\num{0.051}$ & $\num{0.134}\pm\num{0.299}$\\
 0.90 & 168.3 & $\num{0.086}\pm\num{0.156}$ & $\num{1.284}\pm\num{2.610}$\\
 0.95 & N.A. & $\num{0.225}\pm\num{0.351}$ & $\num{8.511}\pm\num{14.30}$\\
 1.00 & N.A. & $\num{7.442}\pm\num{11.164}$ & $\num{201.2}\pm\num{450.5}$\\
 1.10 & N.A. & $\num{6.5E4}\pm\num{1.9E5}$ & $\num{7.0E5}\pm\num{1.9E6}$ \\
  \bottomrule
 \end{tabular}
 \caption{Dataset properties for state-space systems with different eigenvalues.
 The transient decay is the number of discrete transitions on average by which the transient term in \eq{eq:time_evolution} decays below $\num{1E-4}$ (chosen based on the steady state prediction error that the most stable poles converge to on average).
 The mean and standard deviations of the data labels represent a relative challenge for the models -- as the input and output normalizers need to compress a wider range of data to $\mathcal{N}(0,1)$, the more sensitive the learning process becomes (for more on normalization, see \sect{sec:normalization}).
 }
\label{tab:data}
\end{table}

\begin{figure}[h]
    \centering
    \ifjmlrutilsmaths
        \subfigure[Poles at $0.1$, state index 0.]{
        \includegraphics[width=0.3\linewidth]{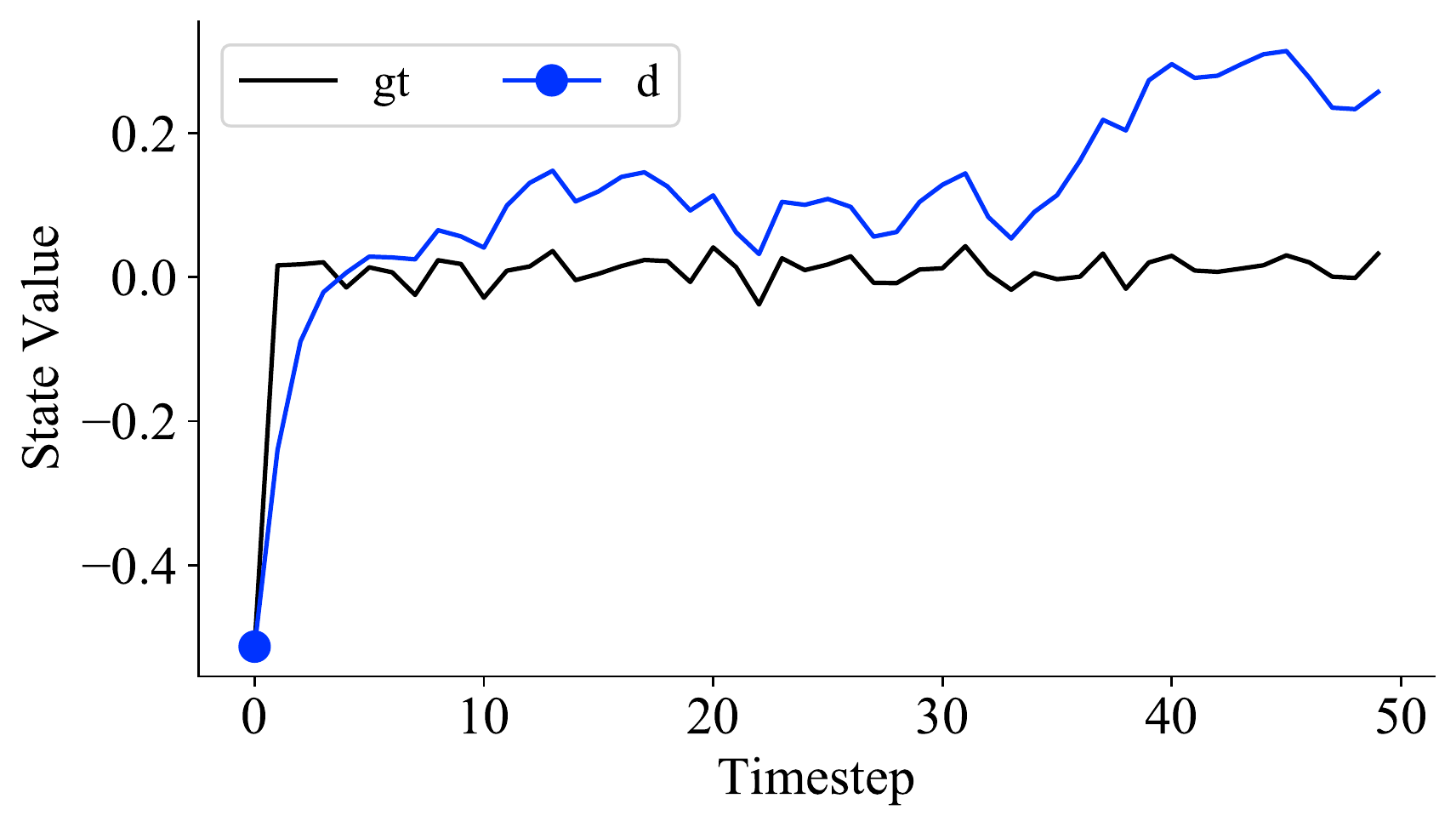}
        }
        \subfigure[Poles at $0.1$, state index 1.]{
        \includegraphics[width=0.3\linewidth]{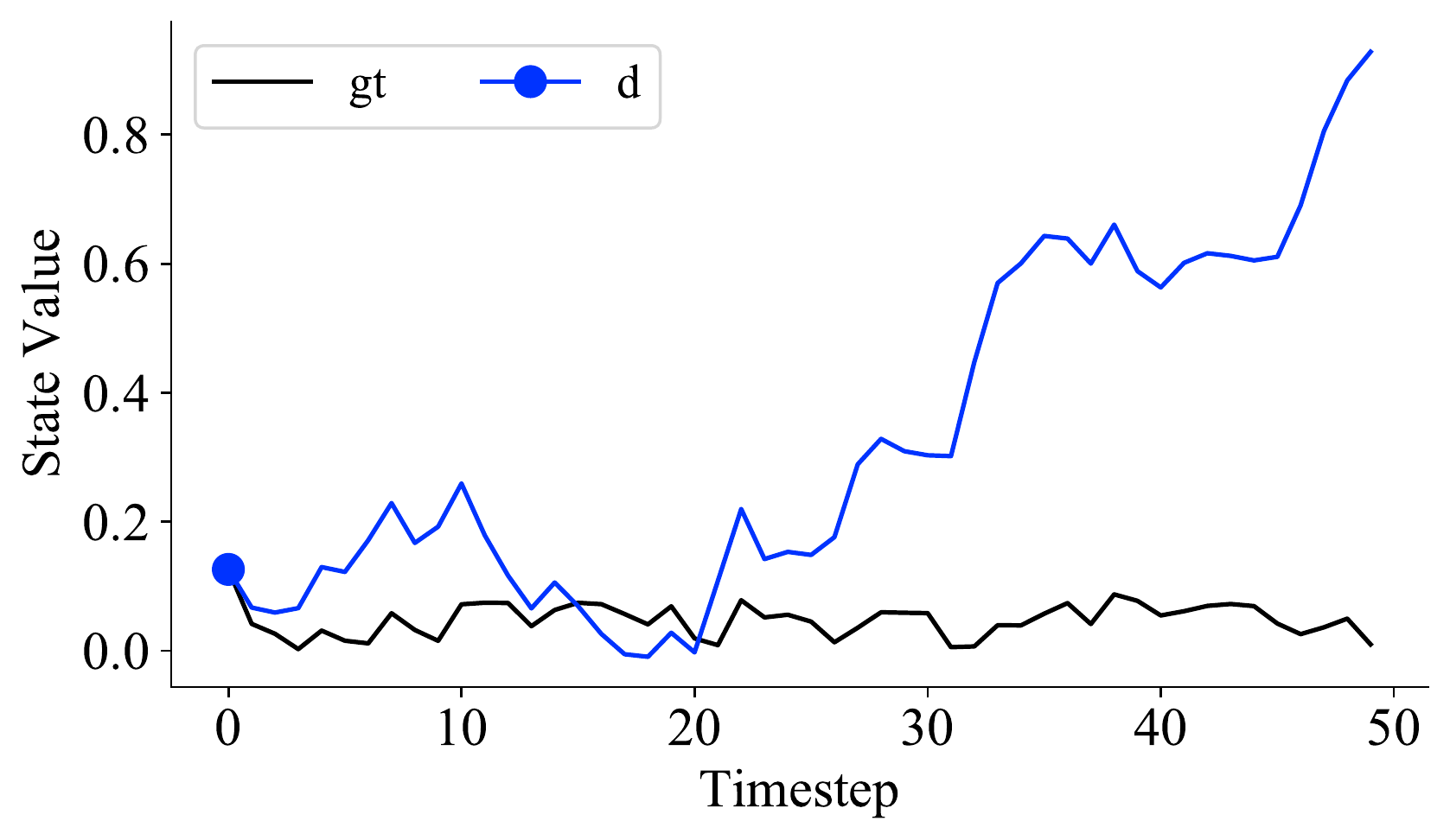}
        }
        \subfigure[Poles at $0.1$, state index 2.]{
        \includegraphics[width=0.3\linewidth]{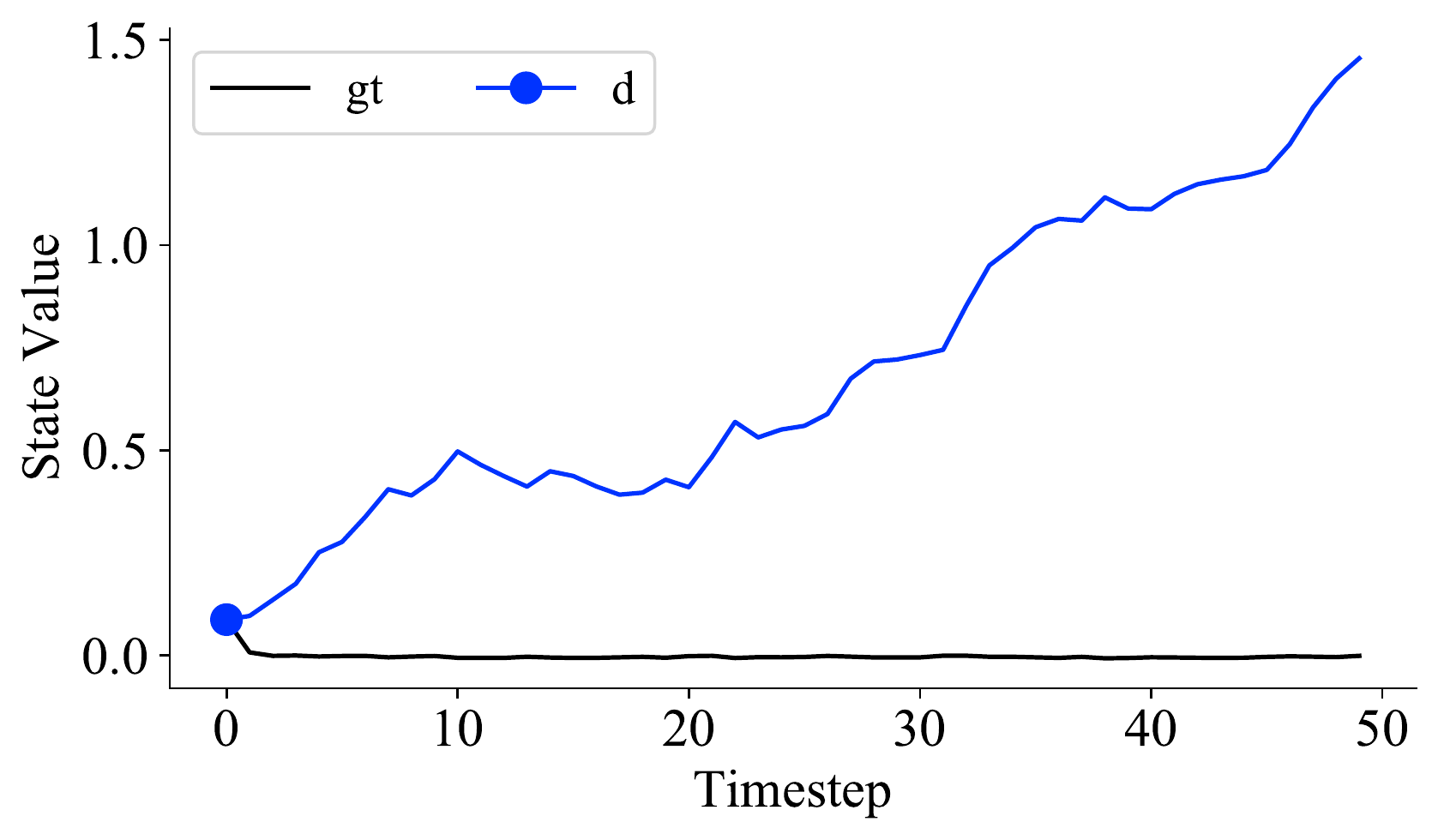}
        }
        \\
        \subfigure[Poles at $0.25$, state index 0.]{
        \includegraphics[width=0.3\linewidth]{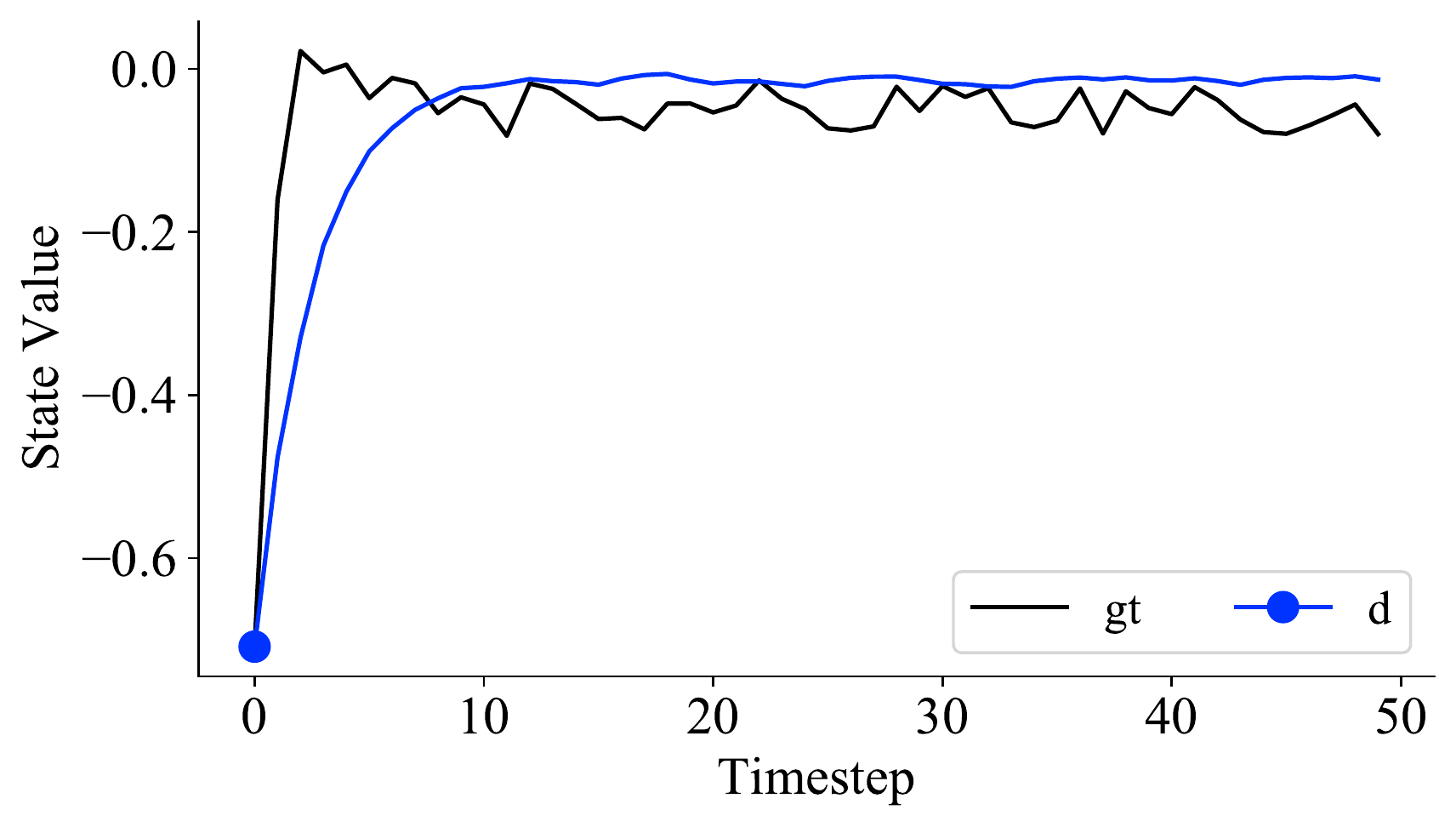}
        }
        \subfigure[Poles at $0.25$, state index 1.]{
        \includegraphics[width=0.3\linewidth]{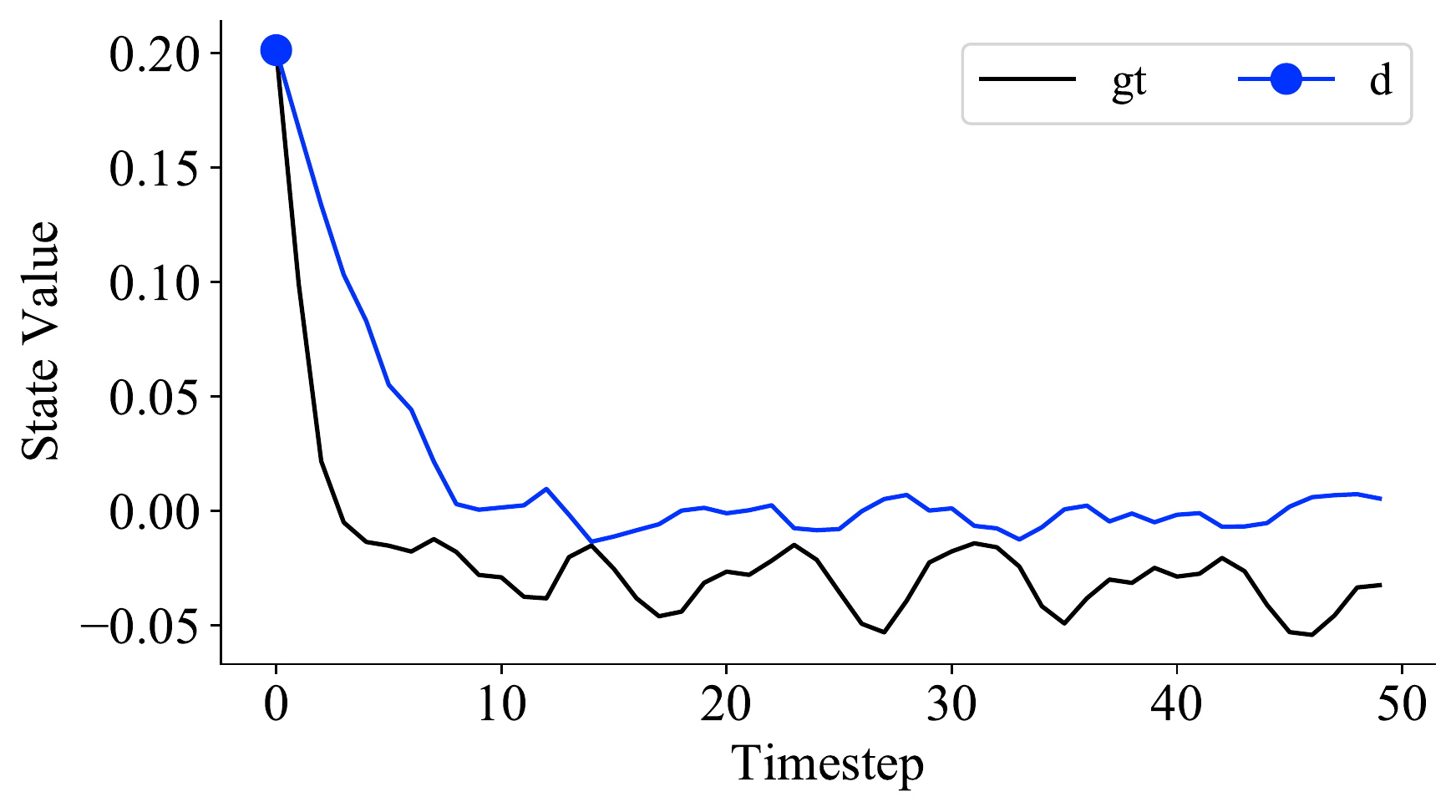}
        }
        \subfigure[Poles at $0.25$, state index 2.]{
        \includegraphics[width=0.3\linewidth]{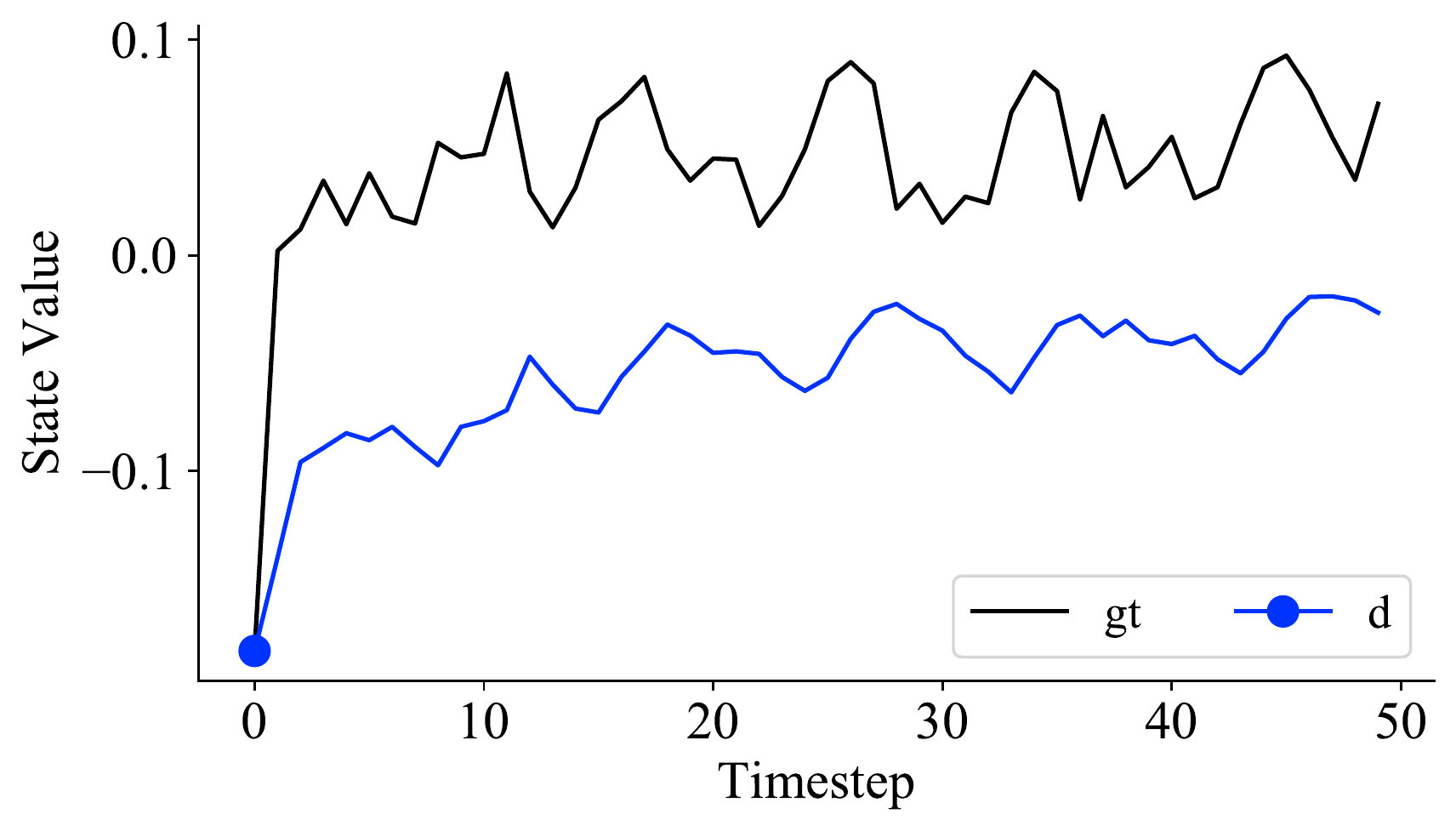}
        }
        \\
        \subfigure[Poles at $0.5$, state index 0.]{
        \includegraphics[width=0.3\linewidth]{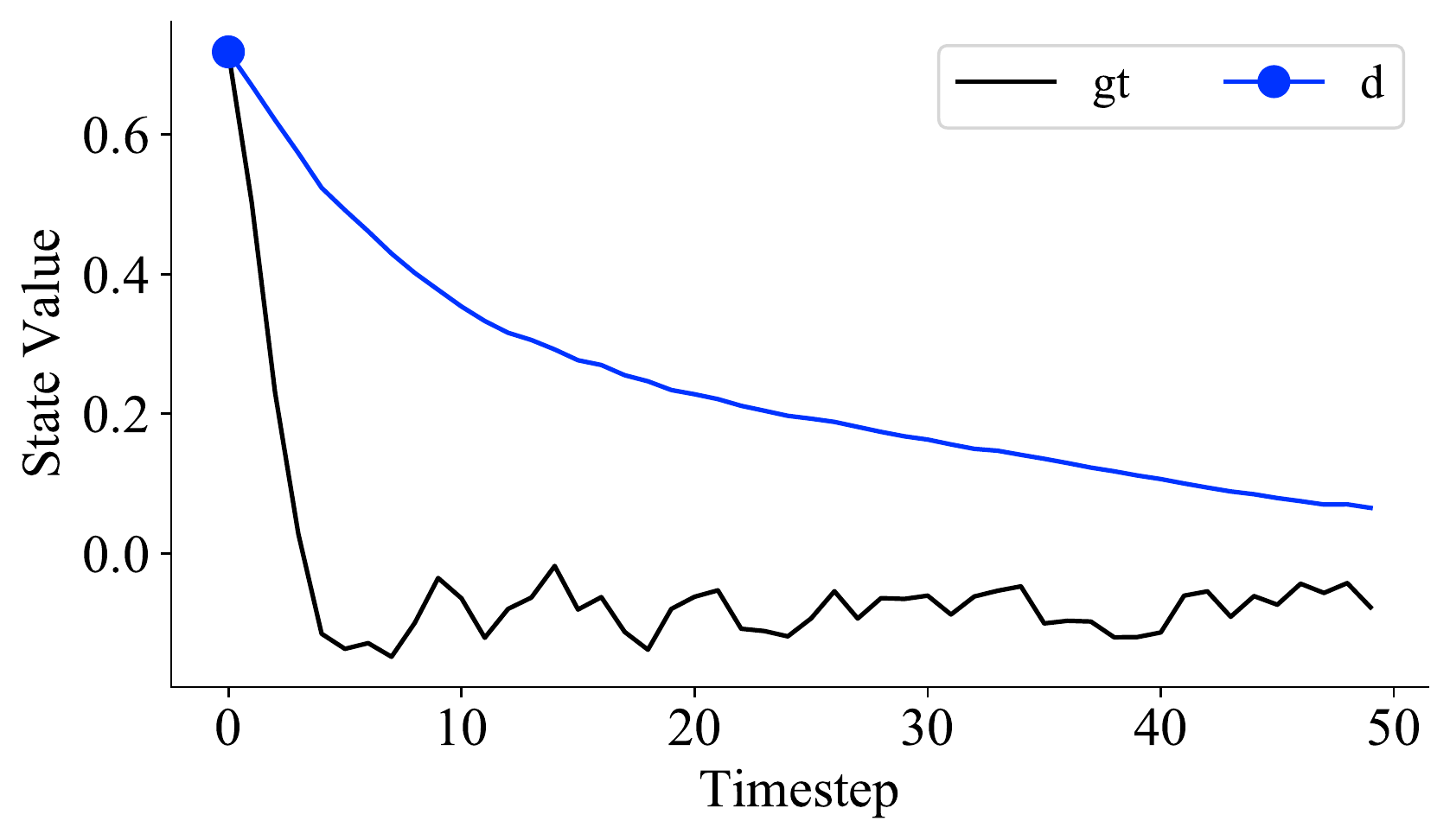}
        }
        \subfigure[Poles at $0.5$, state index 1.]{
        \includegraphics[width=0.3\linewidth]{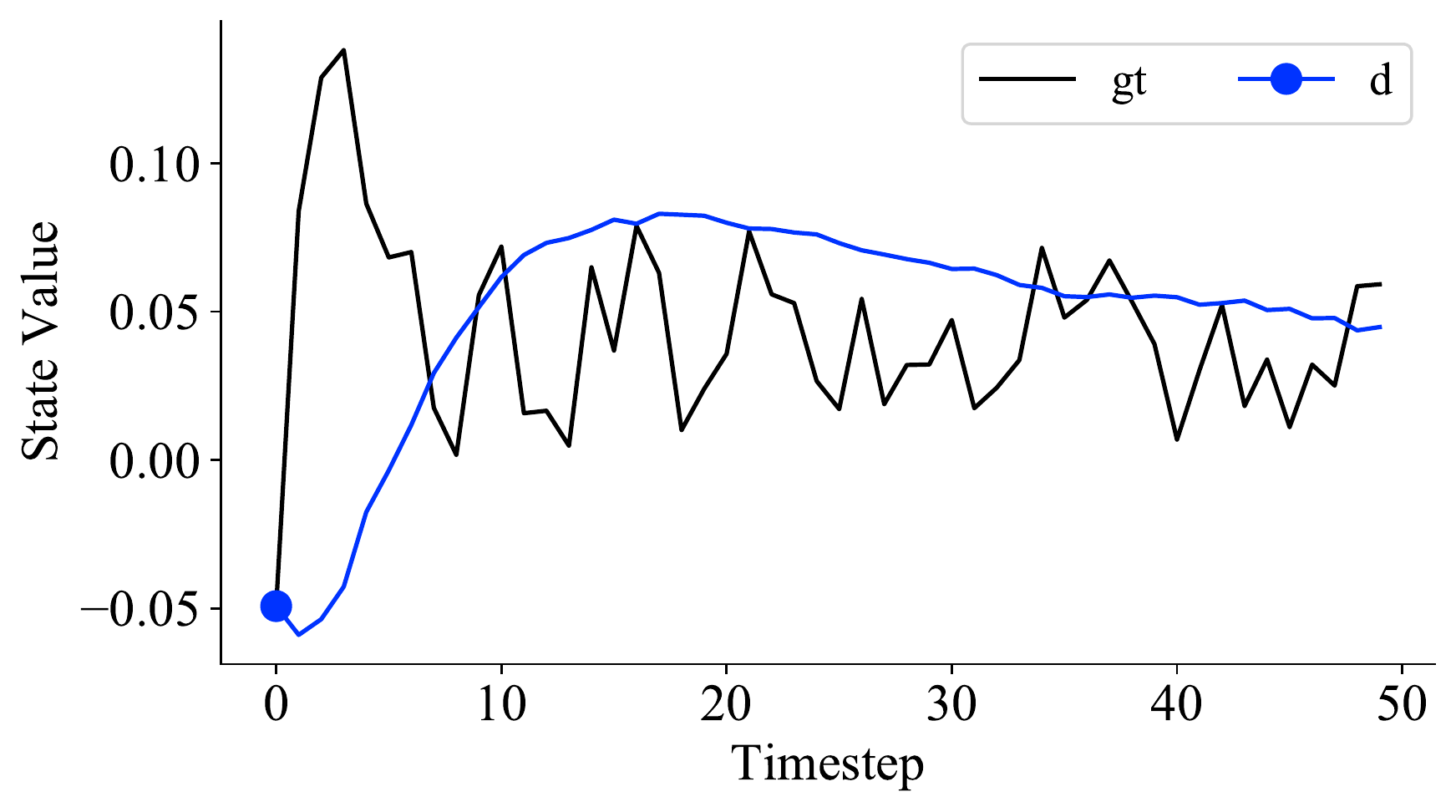}
        }
        \subfigure[Poles at $0.5$, state index 2.]{
        \includegraphics[width=0.3\linewidth]{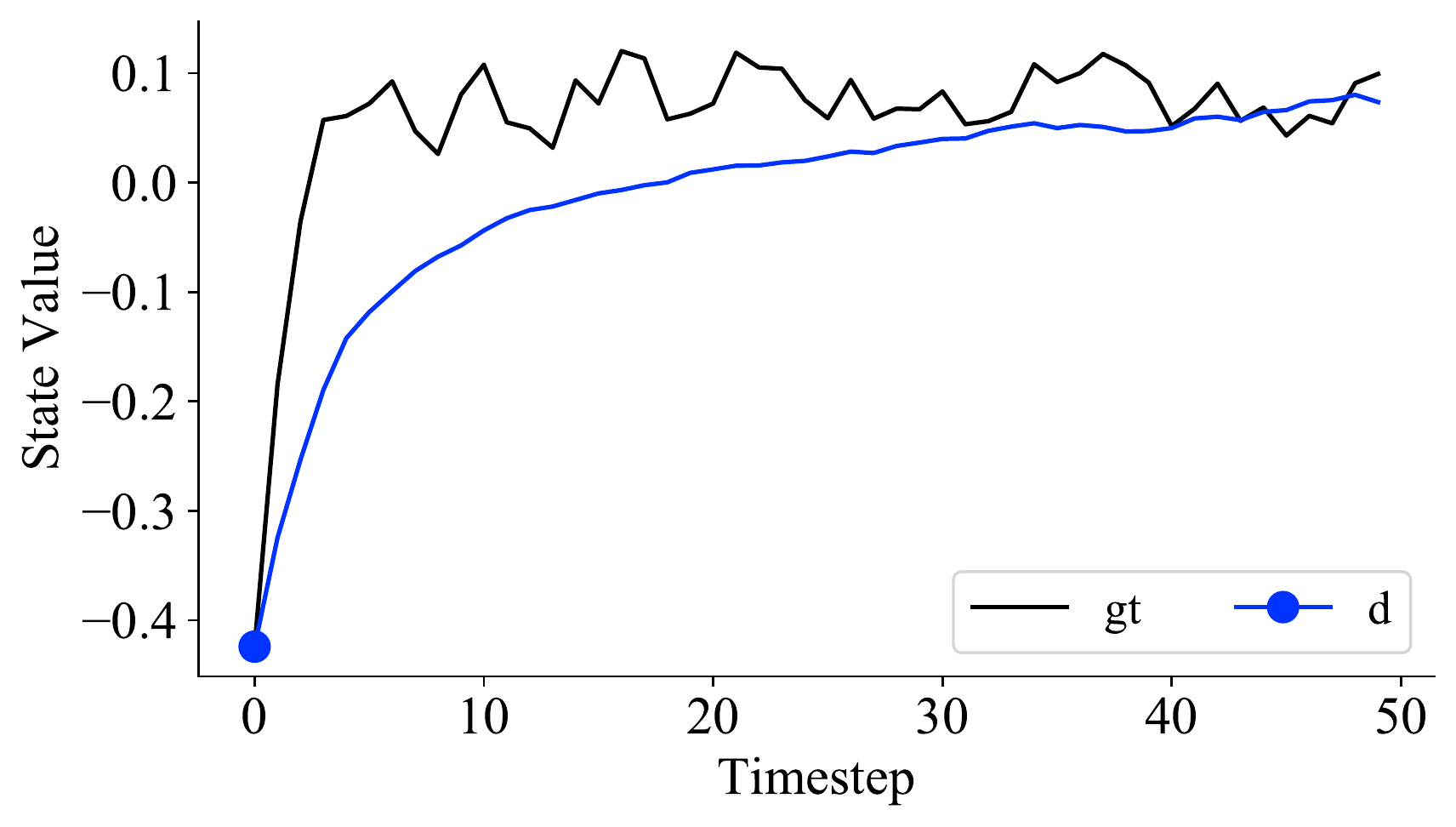}
        }
        \\
    \else
        \begin{subfigure}[t]{0.25\linewidth}
            \centering
            \includegraphics[width=\linewidth]{figures/p010/predictions-state0.pdf}
            \caption{Poles at $0.1$, state index 0.}    
            \label{fig:256hidp001}
        \end{subfigure}
        \quad %
        \begin{subfigure}[t]{0.25\linewidth}  
            \centering 
            \includegraphics[width=\linewidth]{figures/p010/predictions-state1.pdf}
            \caption{Poles at $0.1$, state index 1.}    
            \label{fig:256hidp005}
        \end{subfigure}
        \quad %
        \begin{subfigure}[t]{0.25\linewidth}
            \centering
            \includegraphics[width=\linewidth]{figures/p010/predictions-state2.pdf}
            \caption{Poles at $0.1$, state index 2.}    
            \label{fig:256hidp010}
        \end{subfigure}
        \\
        \begin{subfigure}[t]{0.25\linewidth}
            \centering
            \includegraphics[width=\linewidth]{figures/p025/predictions-state0.pdf}
            \caption{Poles at $0.25$, state index 0.}    
            \label{fig:256hidp001}
        \end{subfigure}
        \quad %
        \begin{subfigure}[t]{0.25\linewidth}  
            \centering 
            \includegraphics[width=\linewidth]{figures/p025/predictions-state1.pdf}
            \caption{Poles at $0.25$, state index 1.}    
            \label{fig:256hidp005}
        \end{subfigure}
        \quad %
        \begin{subfigure}[t]{0.25\linewidth}
            \centering
            \includegraphics[width=\linewidth]{figures/p025/predictions-state2.pdf}
            \caption{Poles at $0.25$, state index 2.}    
            \label{fig:256hidp010}
        \end{subfigure}
        \\
        \begin{subfigure}[t]{0.25\linewidth}
            \centering
            \includegraphics[width=\linewidth]{figures/p050/predictions-state0.pdf}
            \caption{Poles at $0.5$, state index 0.}    
            \label{fig:256hidp001}
        \end{subfigure}
        \quad %
        \begin{subfigure}[t]{0.25\linewidth}  
            \centering 
            \includegraphics[width=\linewidth]{figures/p050/predictions-state1.pdf}
            \caption{Poles at $0.5$, state index 1.}    
            \label{fig:256hidp005}
        \end{subfigure}
        \quad %
        \begin{subfigure}[t]{0.25\linewidth}
            \centering
            \includegraphics[width=\linewidth]{figures/p050/predictions-state2.pdf}
            \caption{Poles at $0.5$, state index 2.}    
            \label{fig:256hidp010}
        \end{subfigure}
    \fi
    \caption{
    Example trajectories for the state-space system in three dimensions showcasing the predictions of deterministic models for $\rho=0.1,0.25,0.5$. %
    The three states for each pole correspond to one example trajectory, and the matrices and initial states are different for each of the representative poles shown.
    \textit{gt} is the true state dynamics and \textit{d} is the one-step model.
    }
    \label{fig:exampletraj}
\end{figure}

\subsection{Model Training}
\label{sec:training}
To learn a model of the dynamics, we use a feedforward neural network with two hidden layers of width 256. 
Ensemble models use $E=5$ members.
The models are trained on 100 trajectories -- all with different matrices $\vec{A},\vec{B}$ with the same poles for the state-space systems.
For the other environments, control policies are randomly sampled to create diverse training and testing data.
The models are trained for 20 epochs with a learning rate of 0.0003 for deterministic and 0.000025 for probabilistic models with the Adam optimizer.
Deterministic models use a batch size of 32 and probabilistic models use a batch size of 64.
All state data is normalized to a unit Gaussian and action data is normalized to $[-1,1]$ for training.
The equations for computing the loss during training are shown for MSE and NLL:
\begin{align}
l_\text{MSE} &= \sum_{n=1}^N \| f_\theta(\mdpstate_n, \mdpaction_n) - \mdpstate_{n+1} \|^2_2 \ ,  \label{eq:loss_mse} \\
l_\text{NLL} &=  \sum_{n=1}^N  [\mu_\theta (s_n, a_n) - s_{n+1}]^T \Sigma_{\theta}^{-1} (s_n, a_n) [\mu_\theta (s_n, a_n) - s_{n+1}] + \text{log det } \Sigma_\theta (s_n, a_n)\,.  \label{pll}
\end{align}

Important to the convergence of supervised learning is the shape and magnitude of the training data.
In \tab{tab:data}, we compare the training data shape for the different state-space system eigenvalues when using true- and delta-state labels for the one-step dynamics model.
As the eigenvalues become more unstable, the variation in the training labels grows exponentially. 
This effect can be counteracted with normalization if it is uniform, but the model loses the ability to differentiate between elements at fine scales, which could render the usefulness of the model low.

\ifx \arxivversion \undefined
\else
The linear model is the result of solving the least squared problem, $\argmin_\omega || X\omega-b ||^2$, where $b=(\mdpstate_{t+1}-\mdpstate_t), \omega = \big[ \vec{\hat{A}} \ \ \vec{\hat{B}}\big], X = \big[S \ \ U \big] $ ($S$ and $U$ are stacked  state and action vectors). 
The next state is then predicted with $\hat{s}_{t+1} = \vec{\hat{A}}\mdpstate_t + \vec{\hat{B}}\mdpaction_t$.
\fi

\section{Additional Experiments}
\label{sec:additional}

\subsection{Further Investigation of the Effect of Model Properties on Compounding Error}

\label{sec:model_appndx}

\begin{figure}
    \centering
    \ifjmlrutilsmaths
        \subfigure[\centering Training set:  1 Trajectory.]{
        \includegraphics[width=0.4\linewidth]{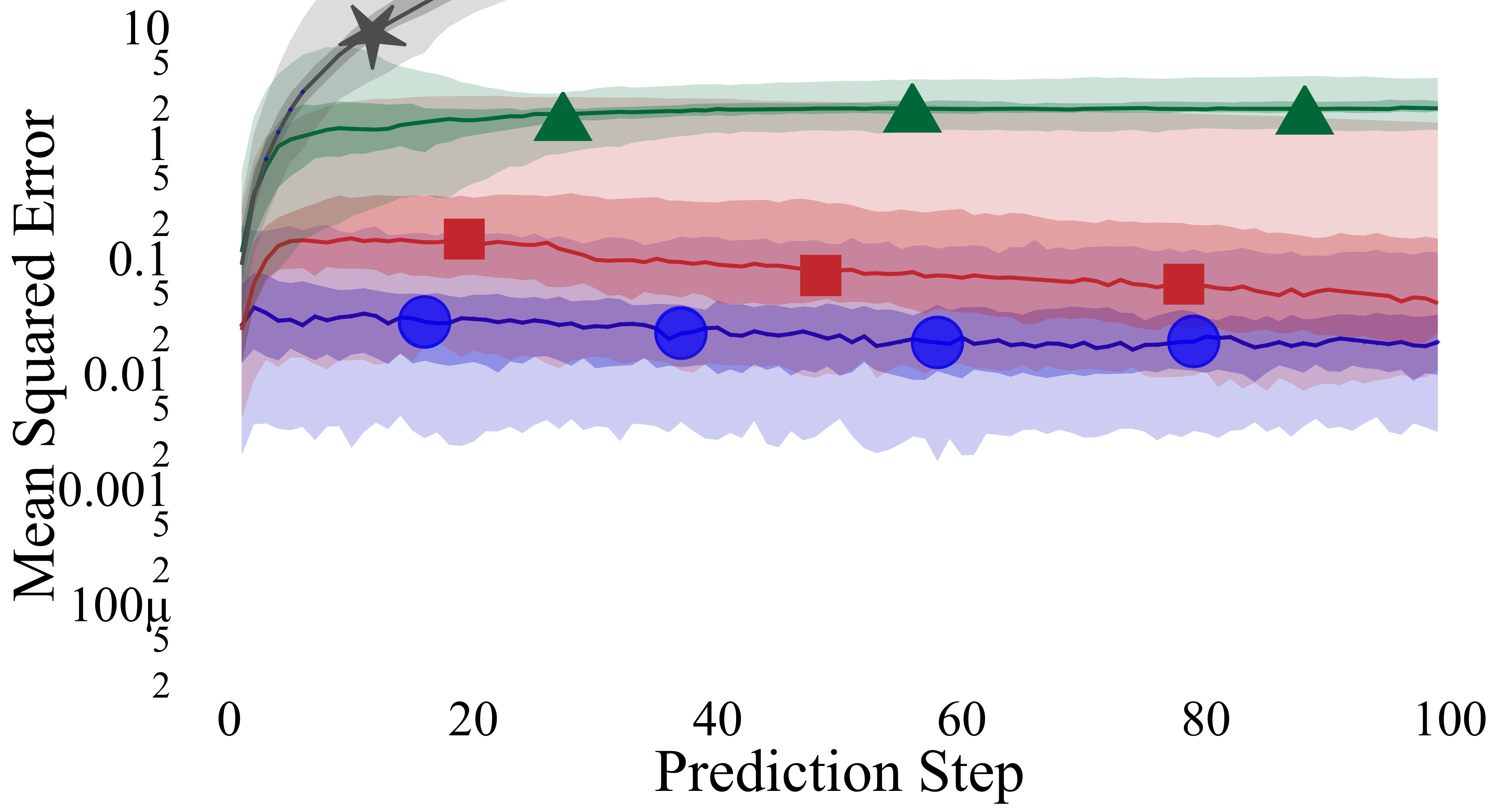}
        }
        \quad
        \subfigure[\centering Training set:  5 Trajectories.]{
        \includegraphics[width=0.4\linewidth]{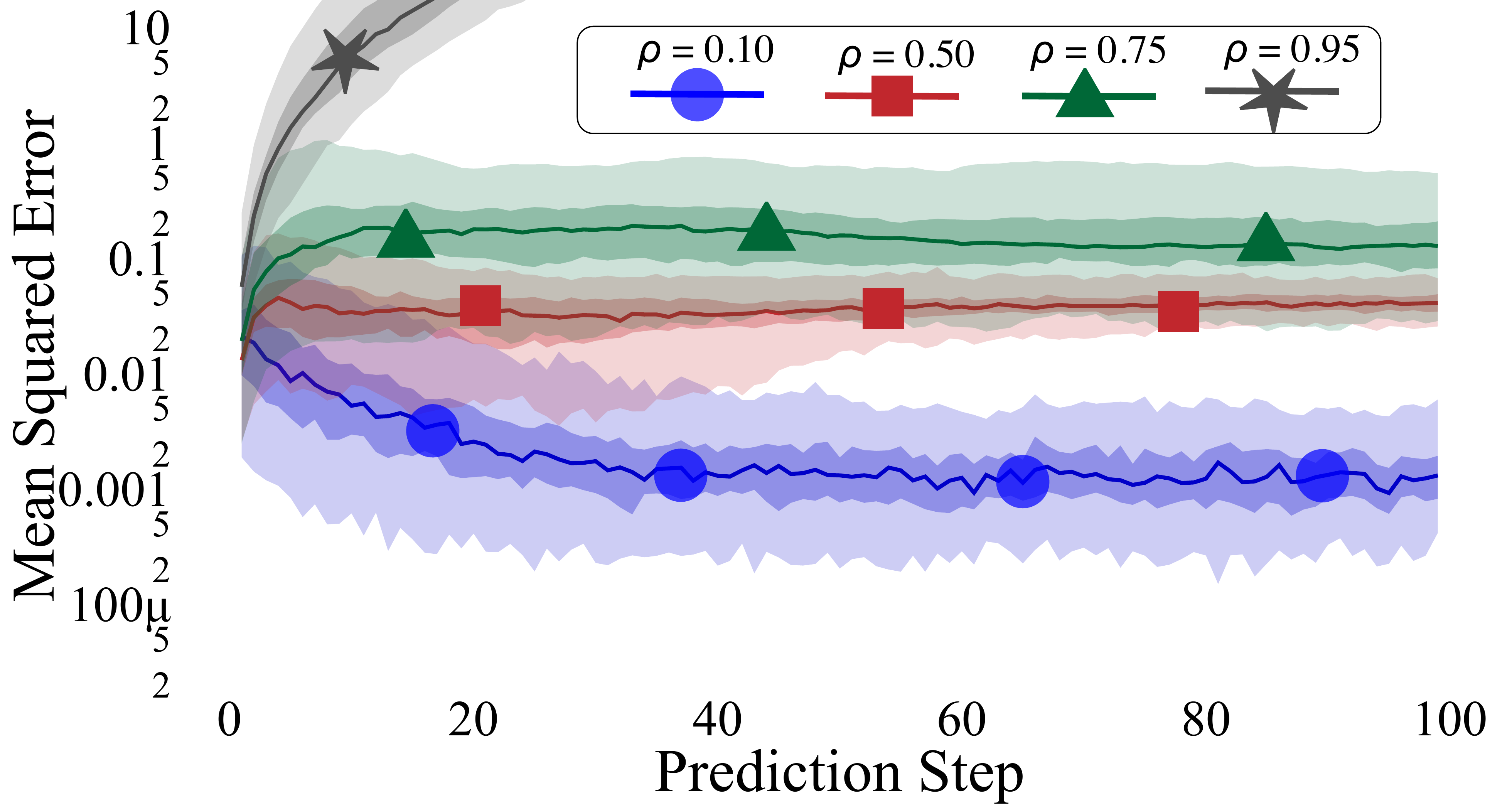}
        }
        \\
        \subfigure[\centering Training set: 10 Trajectories.]{
        \includegraphics[width=0.4\linewidth]{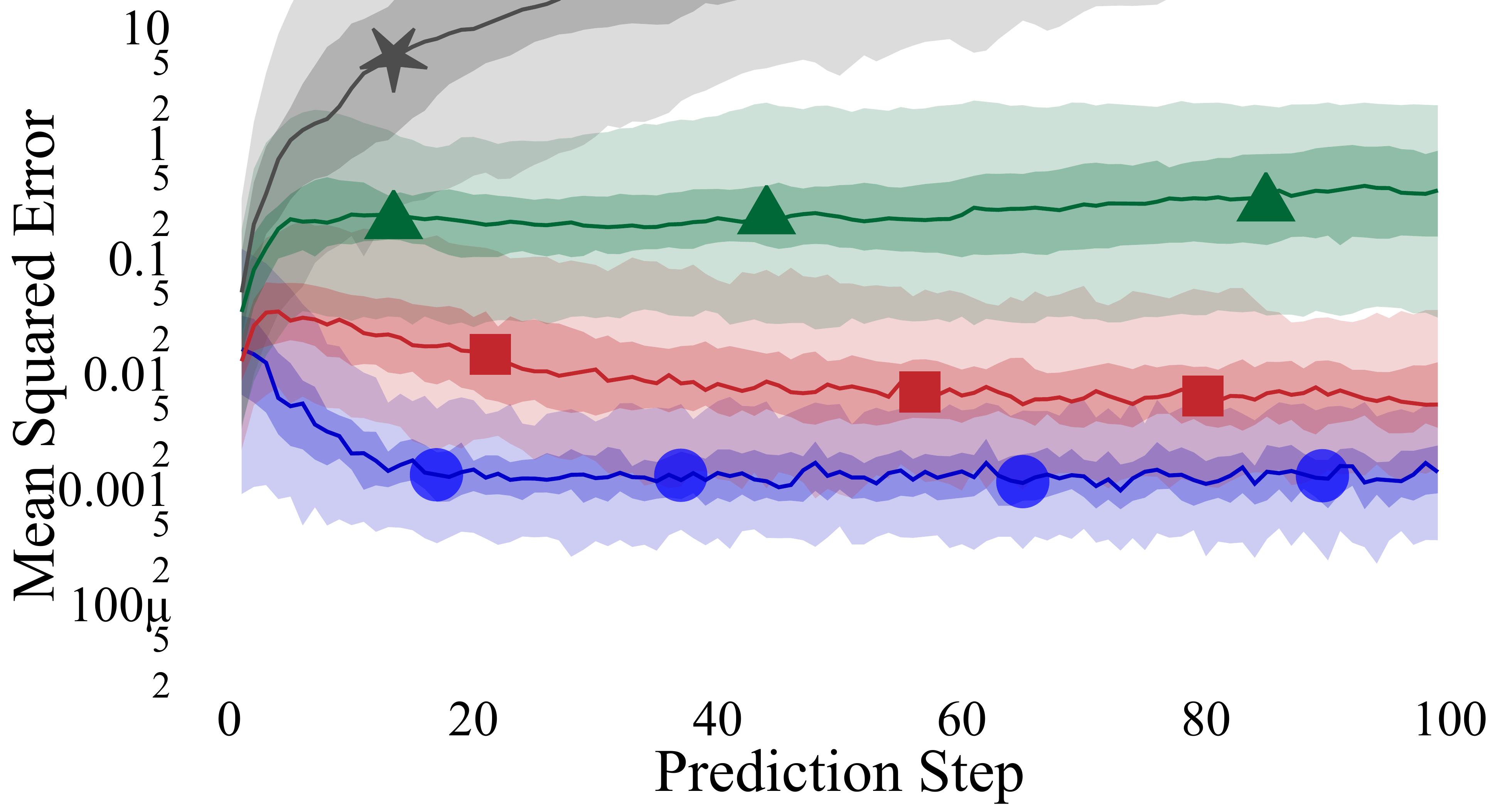}
        }
        \quad
        \subfigure[\centering Training set:  100 Trajectories.]{
        \includegraphics[width=0.4\linewidth]{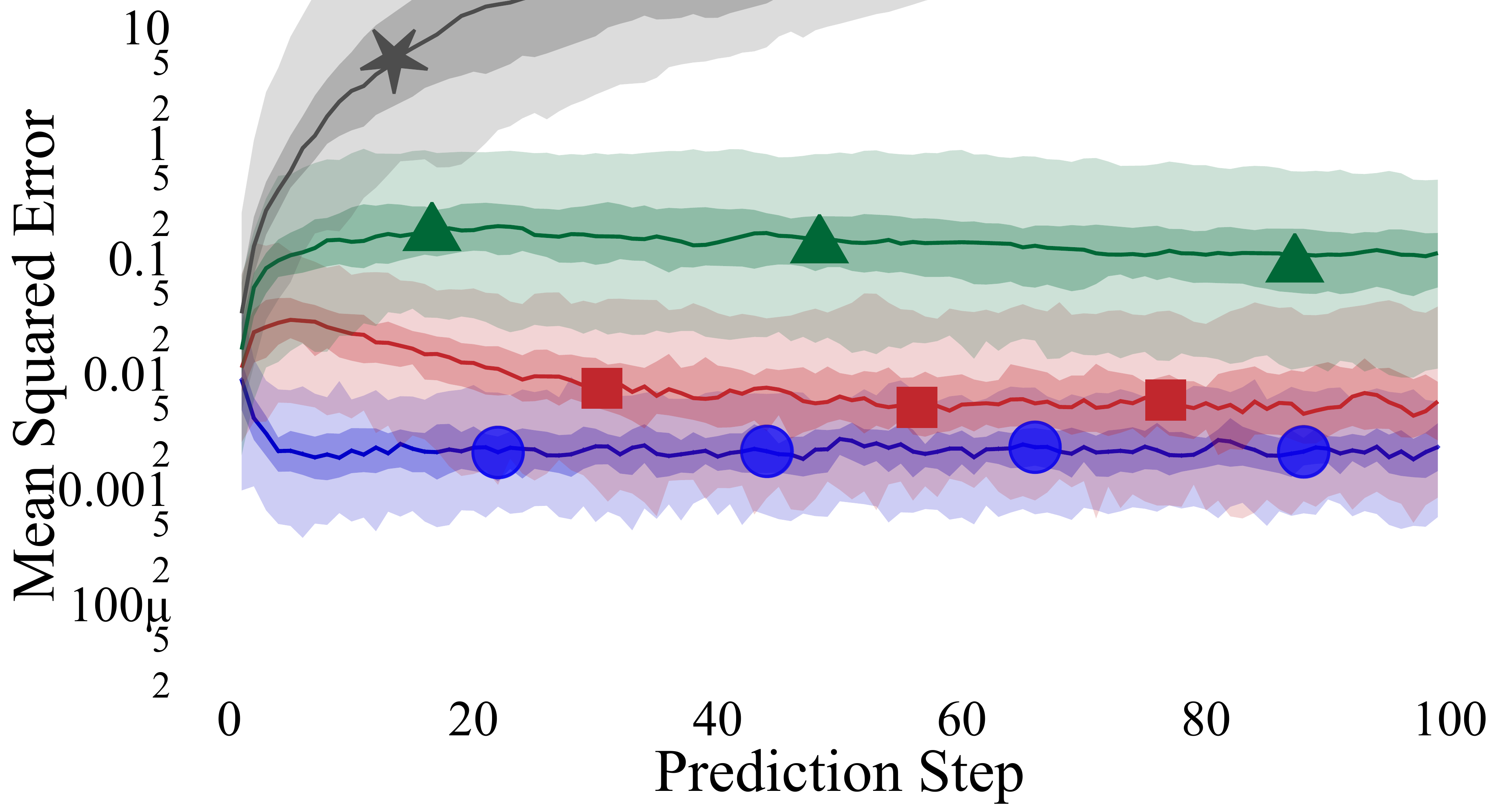}
    }
    \else

     \begin{subfigure}[t]{0.32\linewidth}  
        \centering 
        \includegraphics[width=\linewidth]{figures/training-data/combine-1train.pdf}
        \caption{1 Training Trajectory.}  
        \label{fig:training-data-a}
    \end{subfigure}
    ~
    \begin{subfigure}[t]{0.32\linewidth}  
        \centering 
        \includegraphics[width=\linewidth]{figures/training-data/combine-5train.pdf}
        \caption{5 Training Trajectories. }    
        \label{fig:training-data-b}
    \end{subfigure}
    ~
    \begin{subfigure}[t]{0.32\linewidth}  
        \centering 
        \includegraphics[width=\linewidth]{figures/training-data/combine-10train.pdf}
        \caption{10 Training Trajectories.}  
        \label{fig:training-data-c}
    \end{subfigure}
    \fi
    \caption{
    Prediction error across representative poles for different training set sizes given a constant training set of 100 trajectories for each pole.
    The models quickly converge with only 10 training trajectories.}
    \label{fig:training-set}
\end{figure}
\subsubsection{Model: Capacity}
\label{sec:capacity}
Given recent advancements in deep learning driven by large datasets with evolving model architectures, one-step dynamics models operate with simpler and smaller models and datasets.
The results shown in \fig{fig:compound} show the model prediction accuracy with a field-standard model capacity of 2 hidden layers and 250 neurons. 
To further examine the effects of model capacity, we also test the rate of divergence for deep predictive models on state-space systems with models with substantially fewer or greater parameters.
The results of model predictive error with models of a hidden layer of size 32, shown in \fig{fig:model_paramet}(a-d), and of a model with 3 hidden layers of width 512, shown in \fig{fig:model_paramet}(e-h) show that for a simple task, changing the model size has little impact on prediction accuracy. 
The smaller model has slightly higher prediction error and variance among errors and the larger model has slightly improved prediction accuracy, though the effect is substantially less than the effects of system properties studied in \sect{sec:dynamics}.
Additionally, the model accuracy with different training set sizes is shown in \fig{fig:training-set}, where there is not a substantial effect beyond the first few trajectories.

\begin{figure}[t]
    \centering
    \ifjmlrutilsmaths
        \subfigure[\centering Reduced Model Capacity; $\rho=0.1$.]{
        \includegraphics[width=0.23\linewidth]{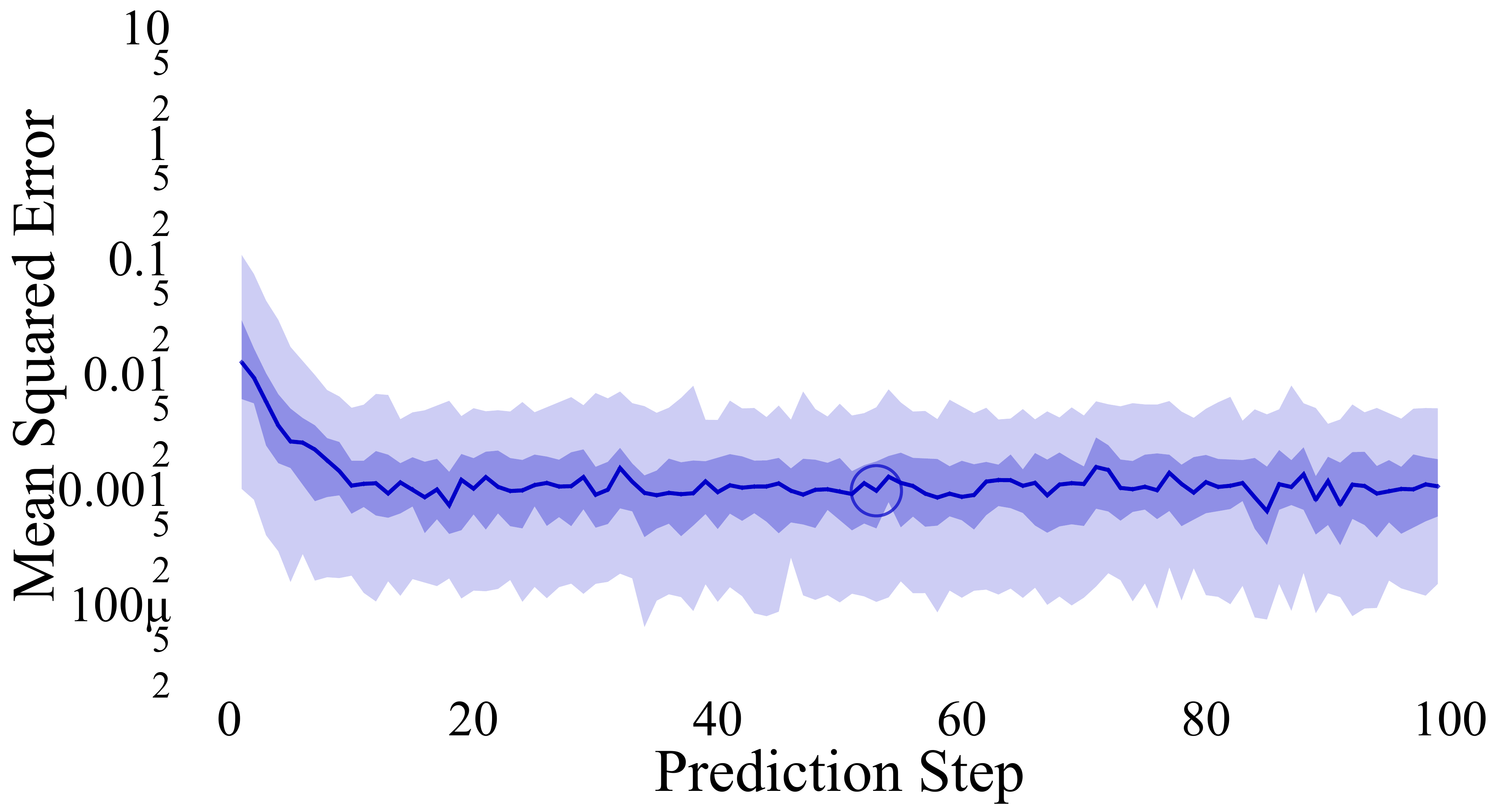}
        }
        \hfill
        \subfigure[\centering Reduced Model Capacity; $\rho=0.5$.]{
        \includegraphics[width=0.23\linewidth]{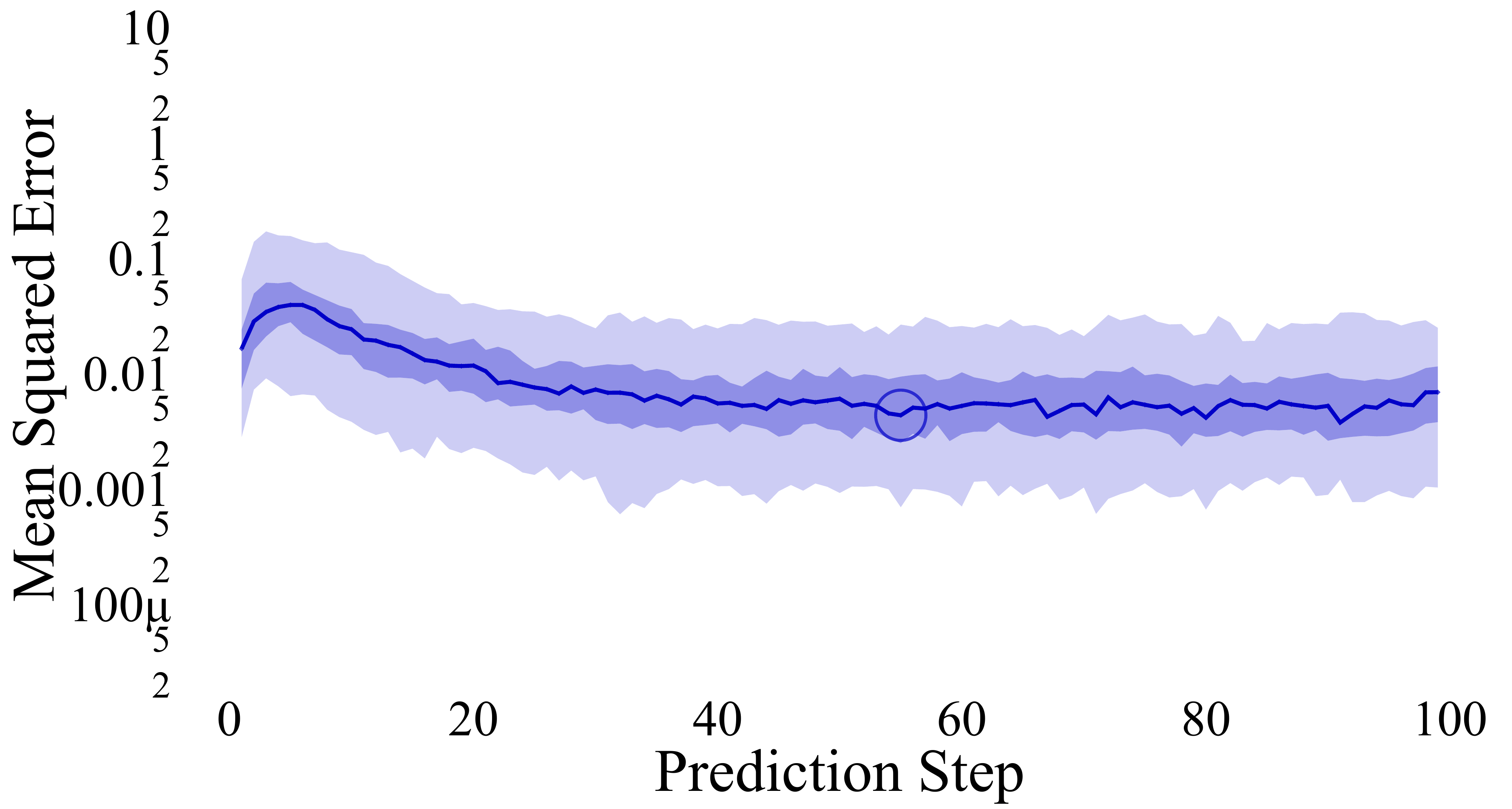}
        }
        \hfill
        \subfigure[\centering Reduced Model Capacity; $\rho=0.75$.]{
        \includegraphics[width=0.23\linewidth]{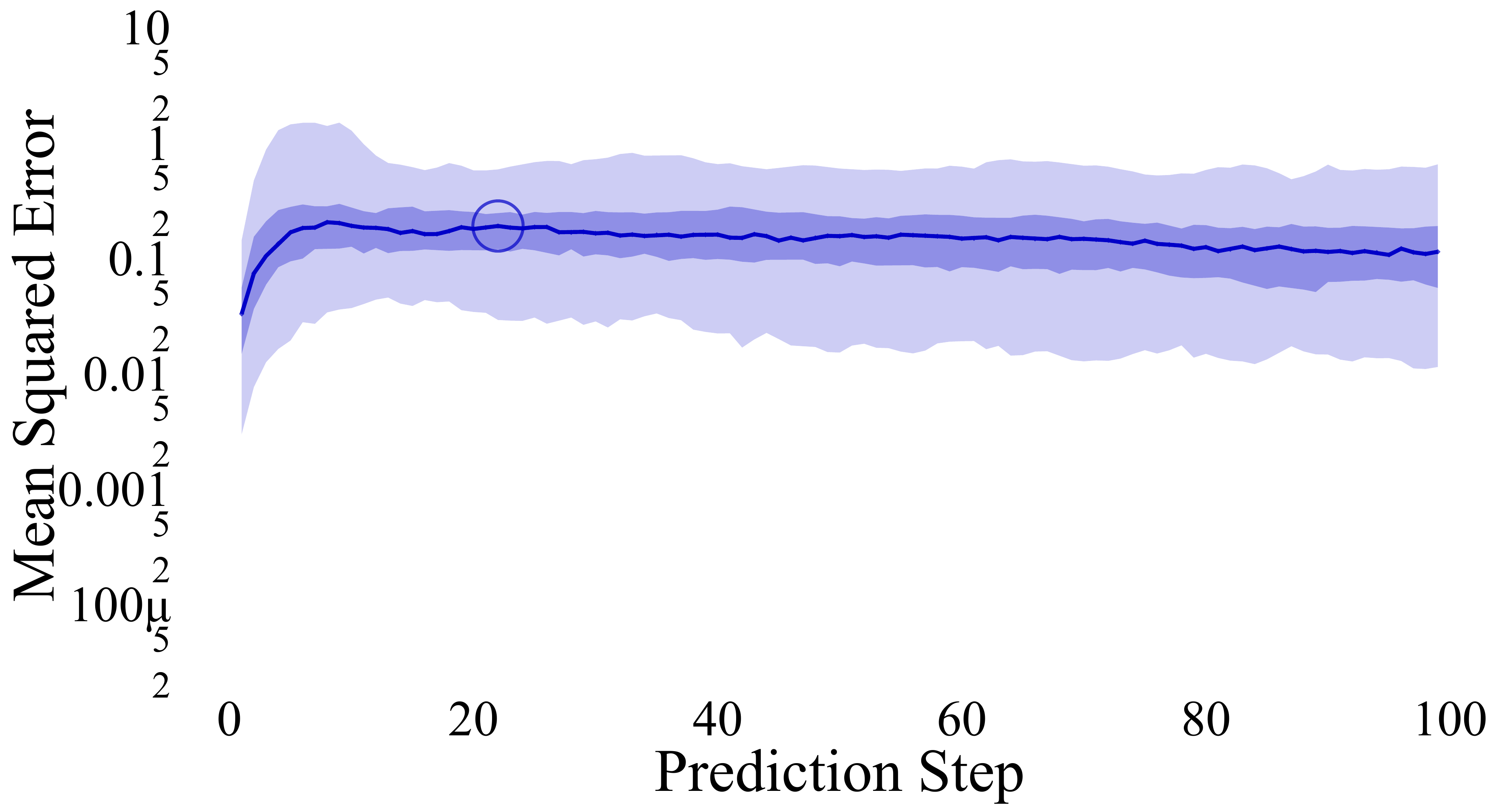}
        }
        \hfill
        \subfigure[\centering Reduced Model Capacity; $\rho=0.95$.]{
        \includegraphics[width=0.23\linewidth]{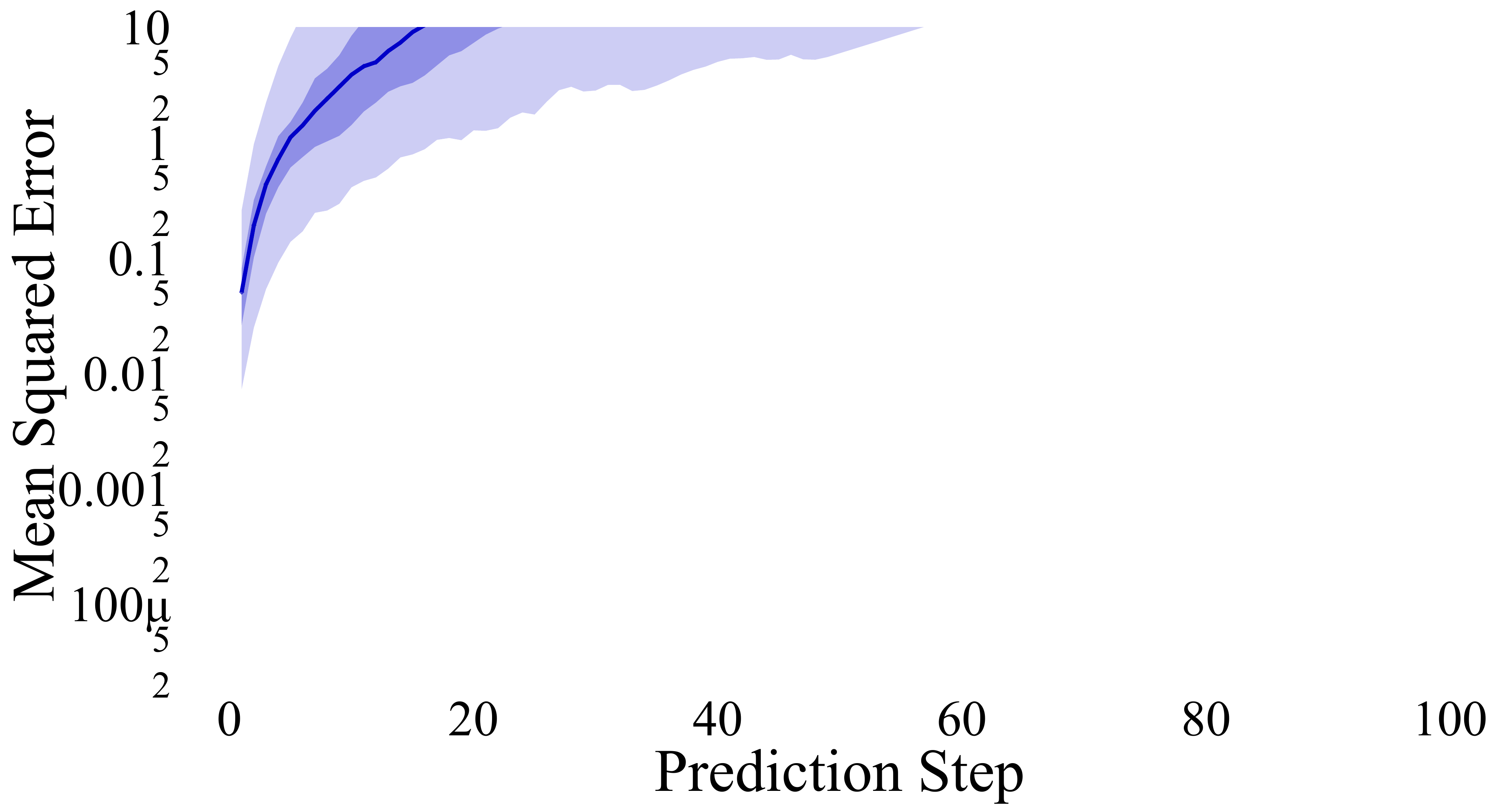}
        }
        \\
        \subfigure[\centering Increased Model Capacity; $\rho=0.1$.]{
        \includegraphics[width=0.23\linewidth]{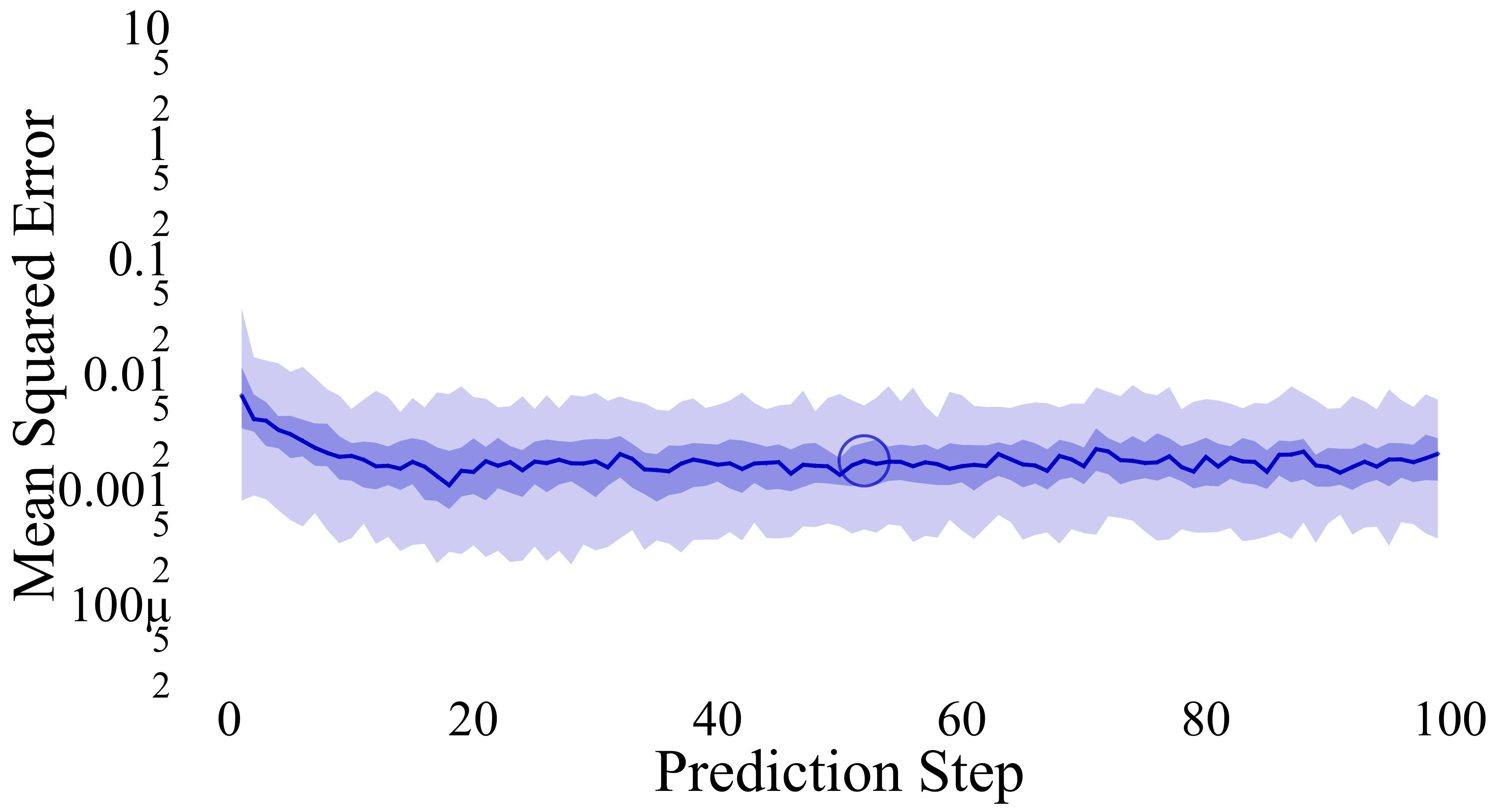}
        }
        \hfill
        \subfigure[\centering Increased Model Capacity; $\rho=0.5$.]{
        \includegraphics[width=0.23\linewidth]{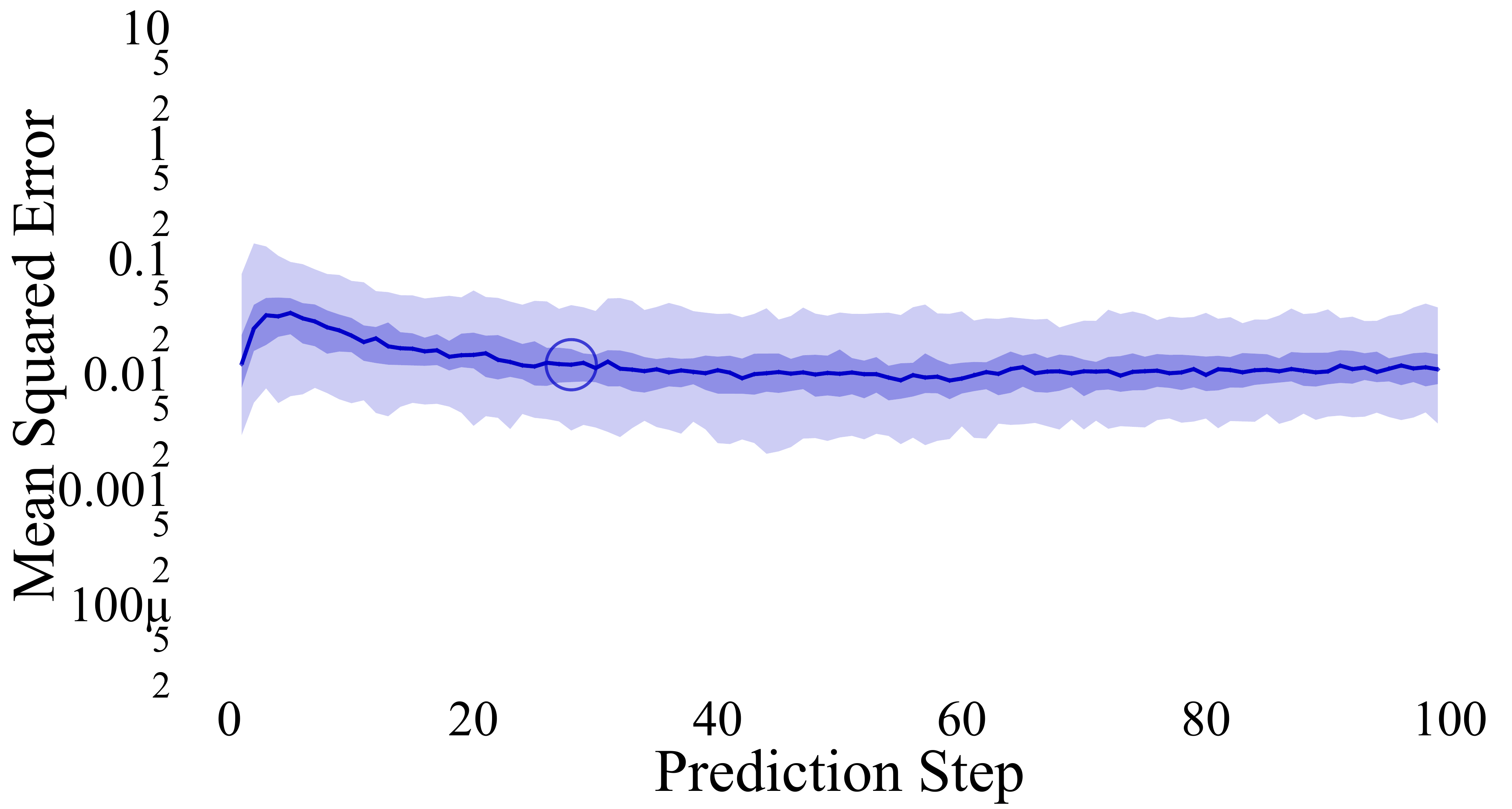}
        }
        \hfill
        \subfigure[\centering Increased Model Capacity; $\rho=0.75$.]{
        \includegraphics[width=0.23\linewidth]{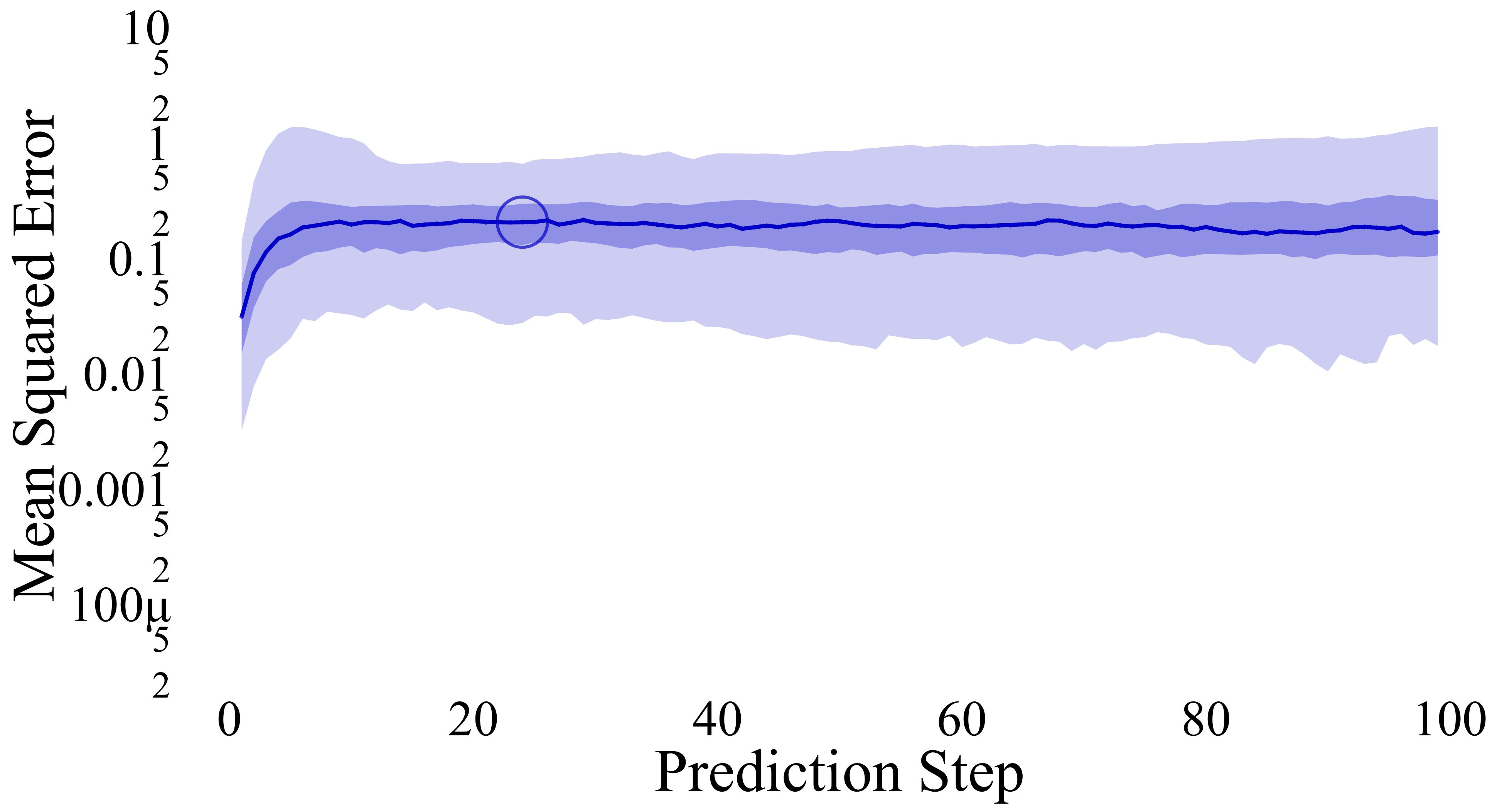}
        }
        \hfill
        \subfigure[\centering Increased Model Capacity; $\rho=0.95$.]{
        \includegraphics[width=0.23\linewidth]{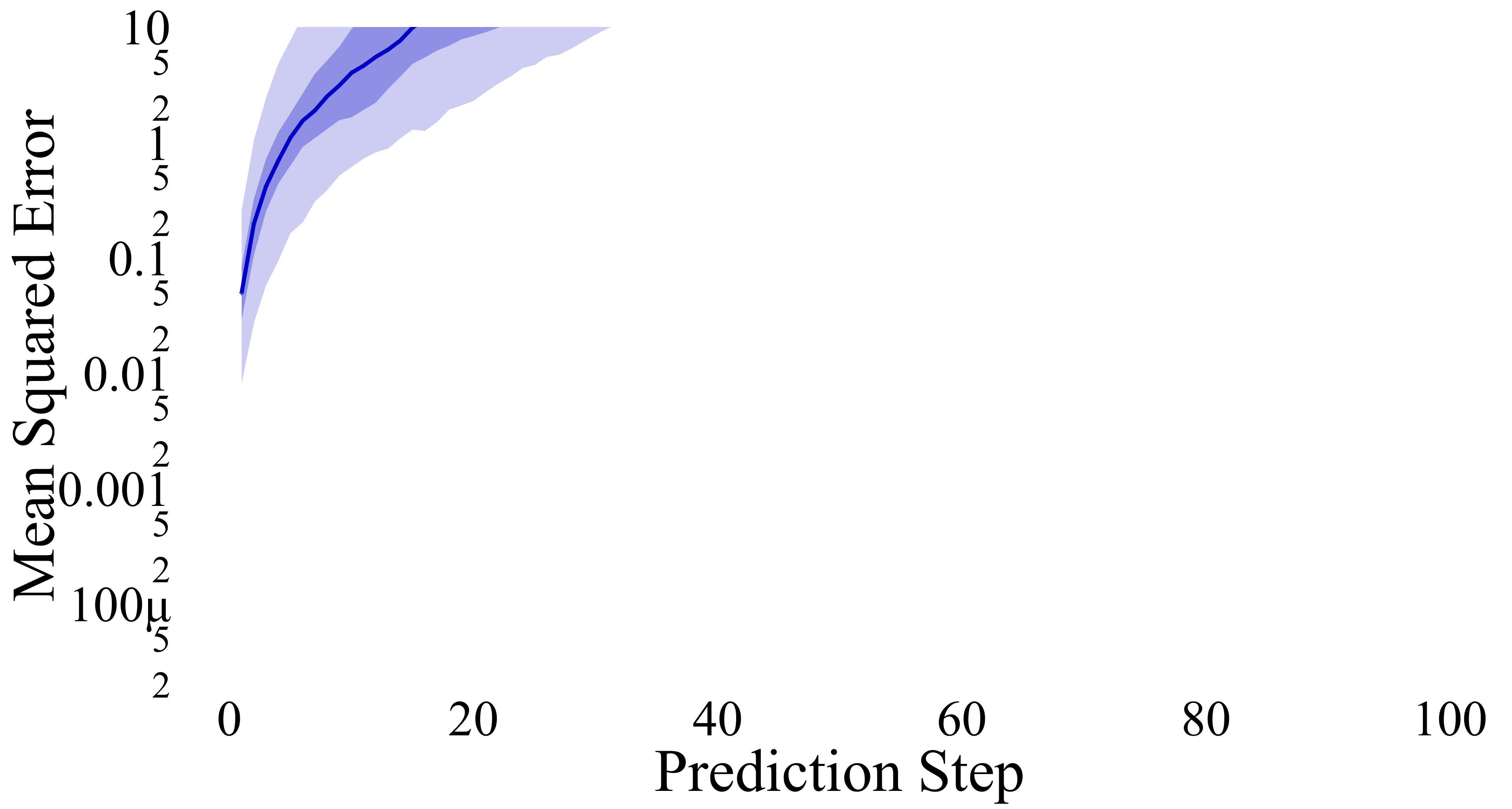}
        }
        \\
        \subfigure[\centering No Normalization; $\rho=0.1$.]{
        \includegraphics[width=0.23\linewidth]{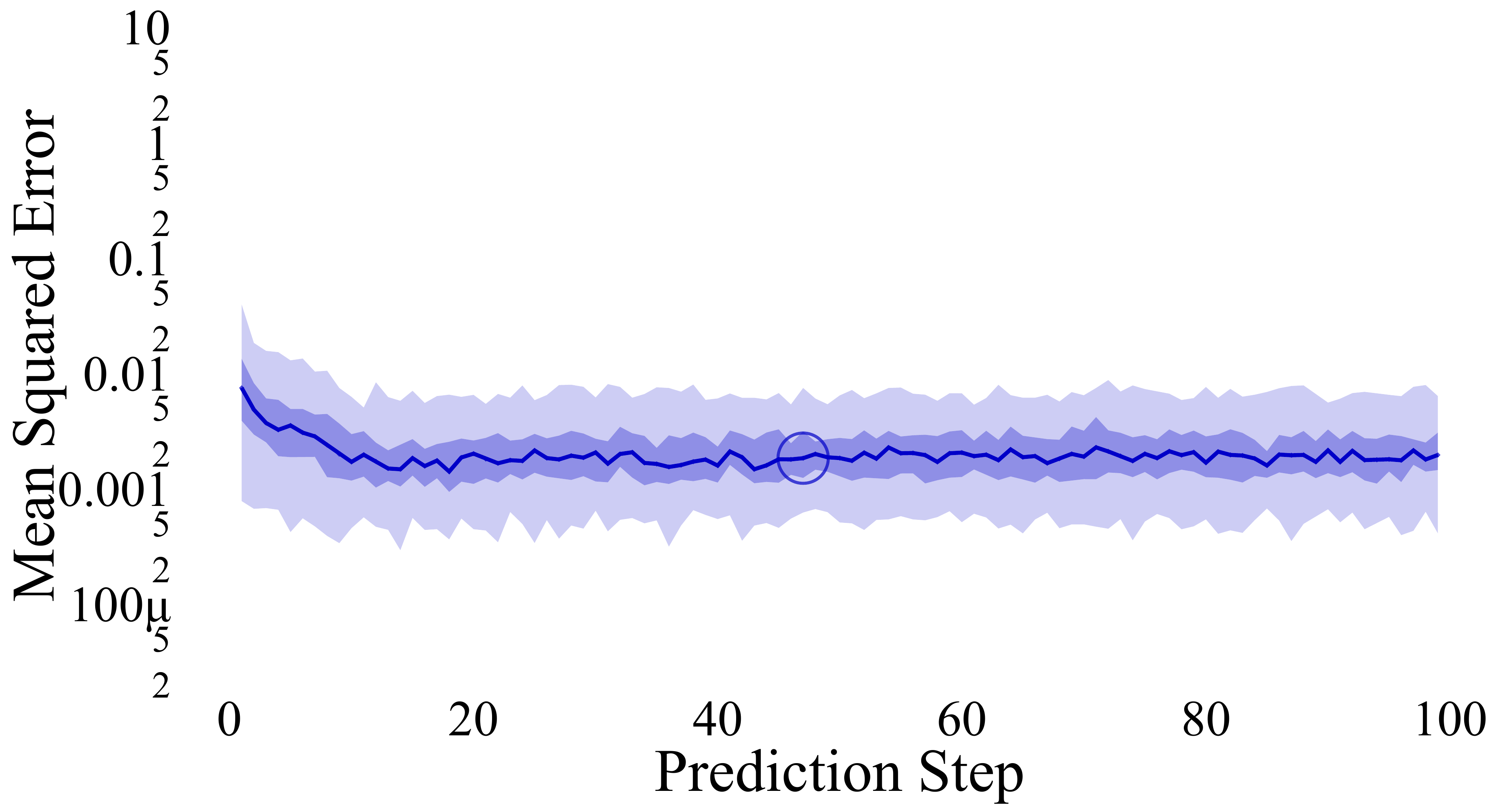}
        }
        \hfill
        \subfigure[\centering No Normalization; $\rho=0.5$.]{
        \includegraphics[width=0.23\linewidth]{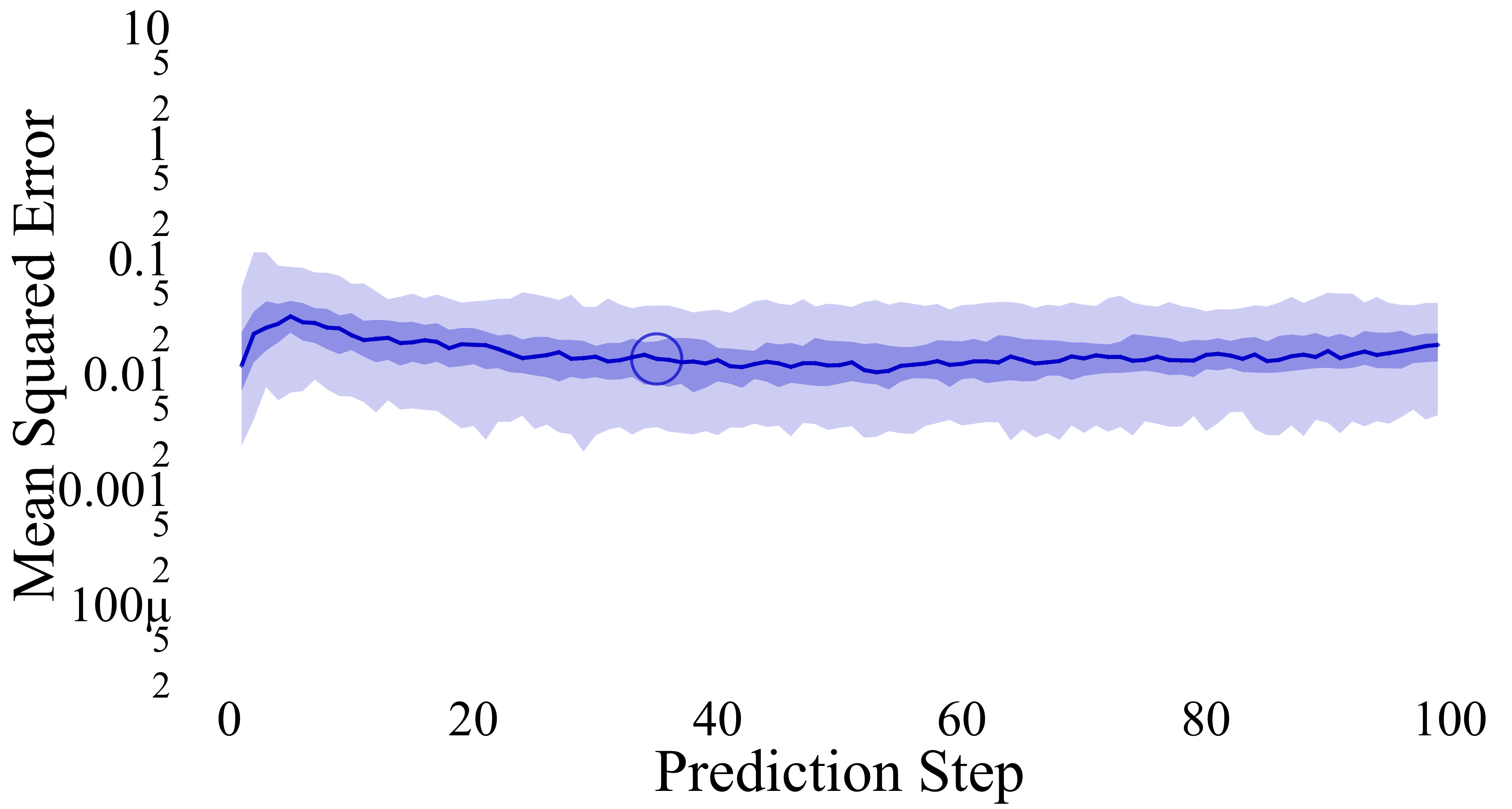}
        }
        \hfill
        \subfigure[\centering No Normalization; $\rho=0.75$.]{
        \includegraphics[width=0.23\linewidth]{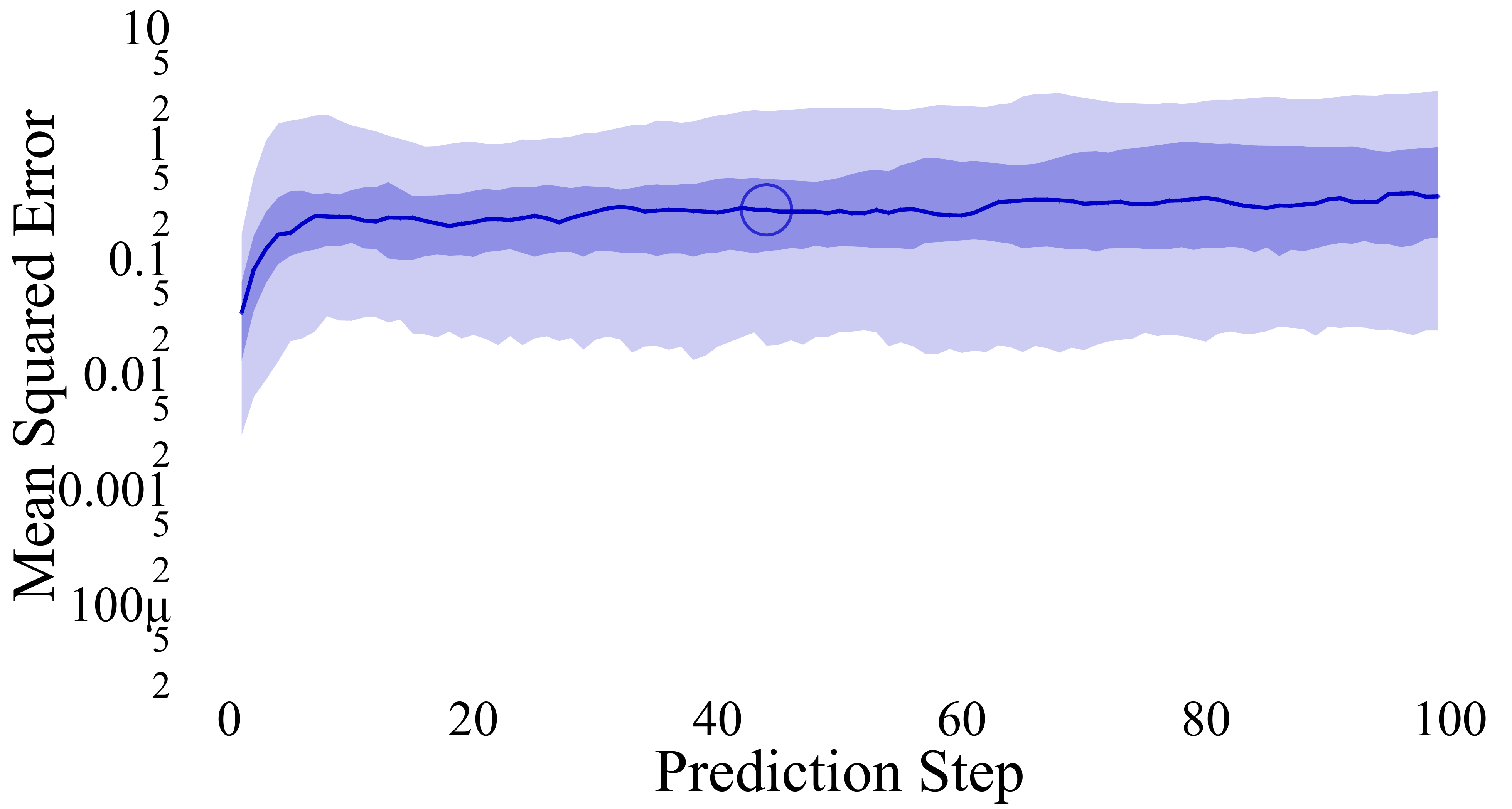}
        }
        \hfill
        \subfigure[\centering No Normalization; $\rho=0.95$.]{
        \includegraphics[width=0.23\linewidth]{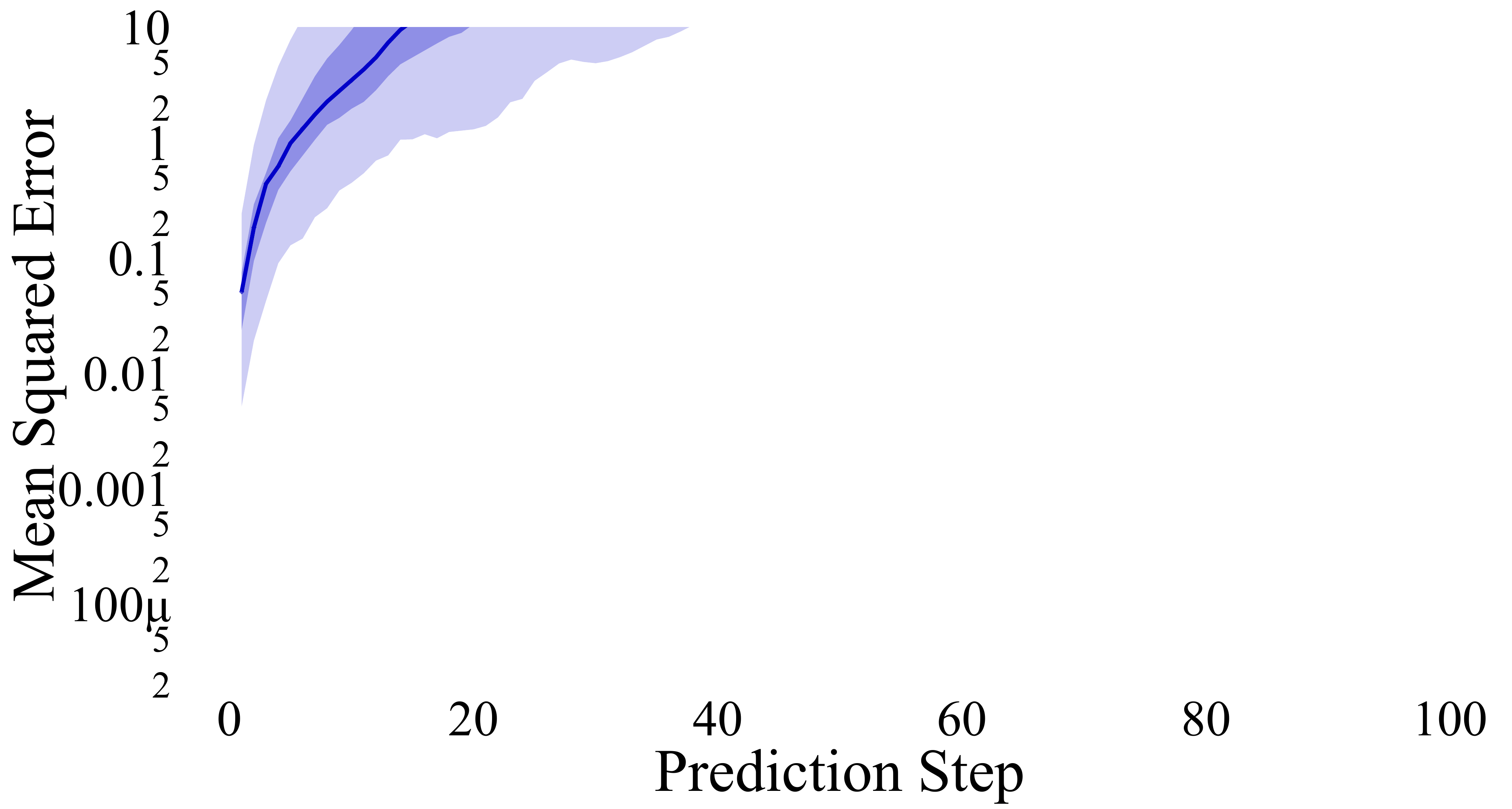}
        }
        \\
        \subfigure[\centering Other Models; $\rho=0.1$.]{
        \includegraphics[width=0.23\linewidth]{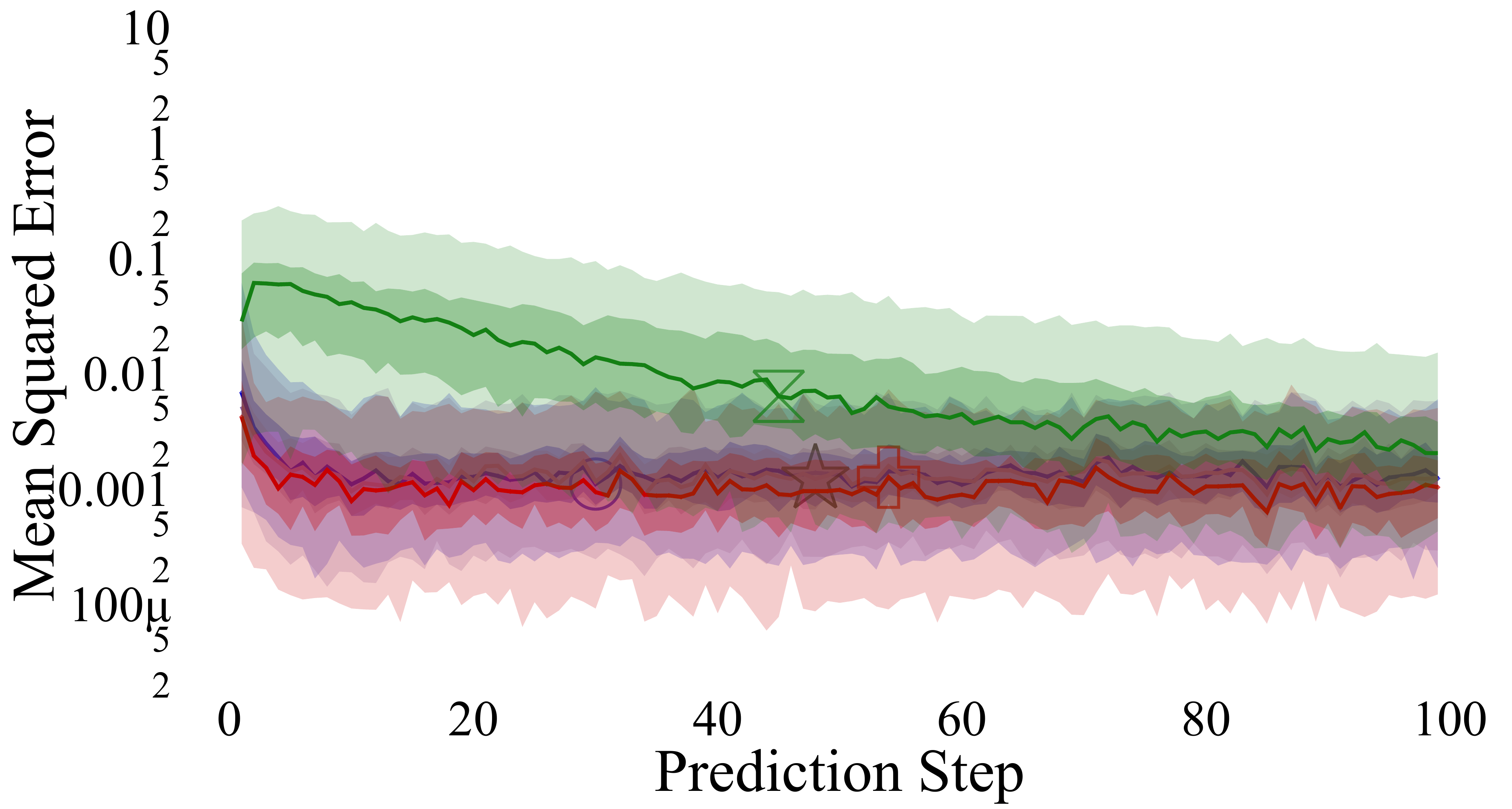}
        }
        \hfill
        \subfigure[\centering Other Models; $\rho=0.5$.]{
        \includegraphics[width=0.23\linewidth]{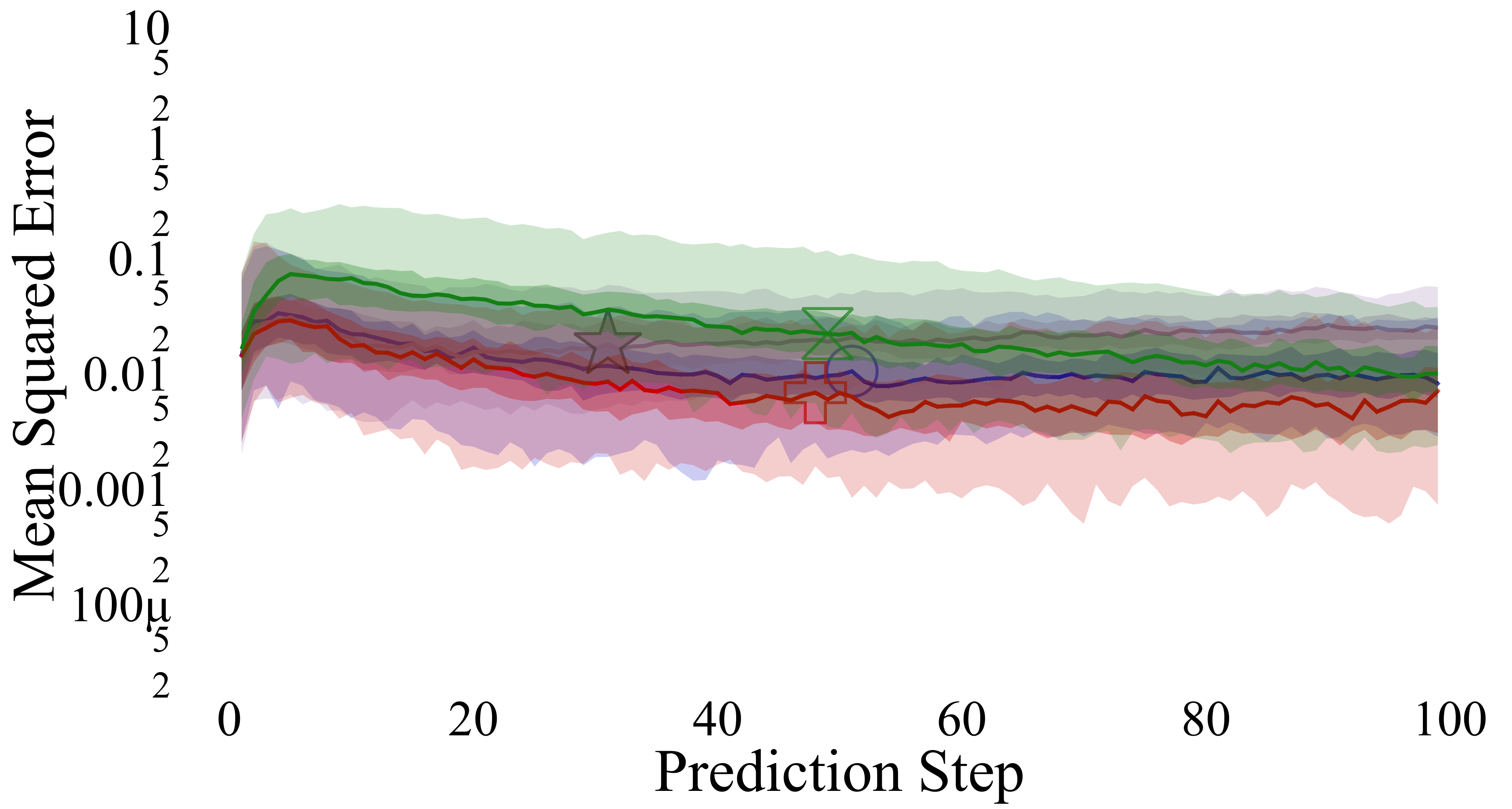}
        }
        \hfill
        \subfigure[\centering Other Models; $\rho=0.75$.]{
        \includegraphics[width=0.23\linewidth]{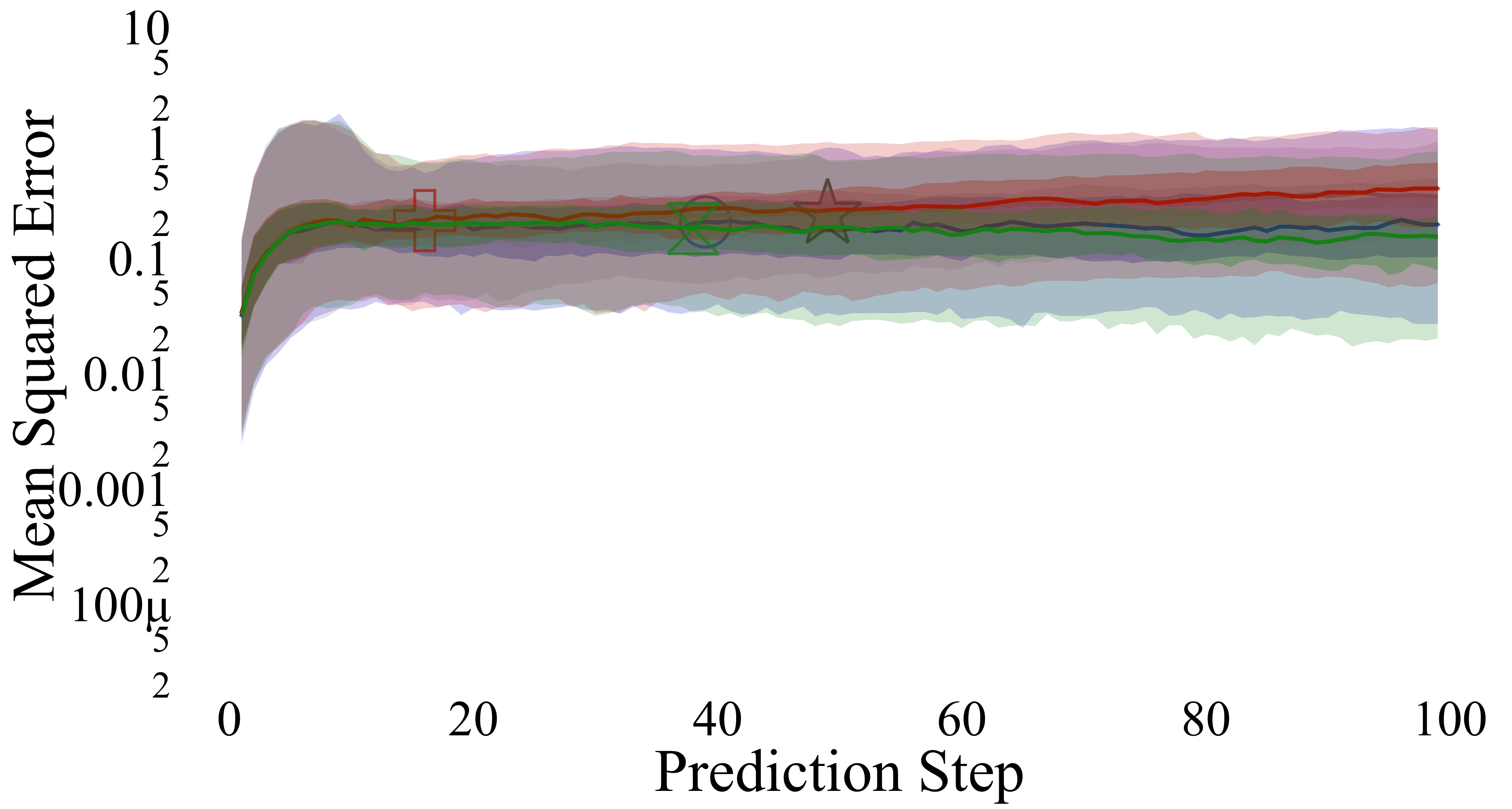}
        }
        \hfill
        \subfigure[\centering Other Models; $\rho=0.95$.]{
        \includegraphics[width=0.23\linewidth]{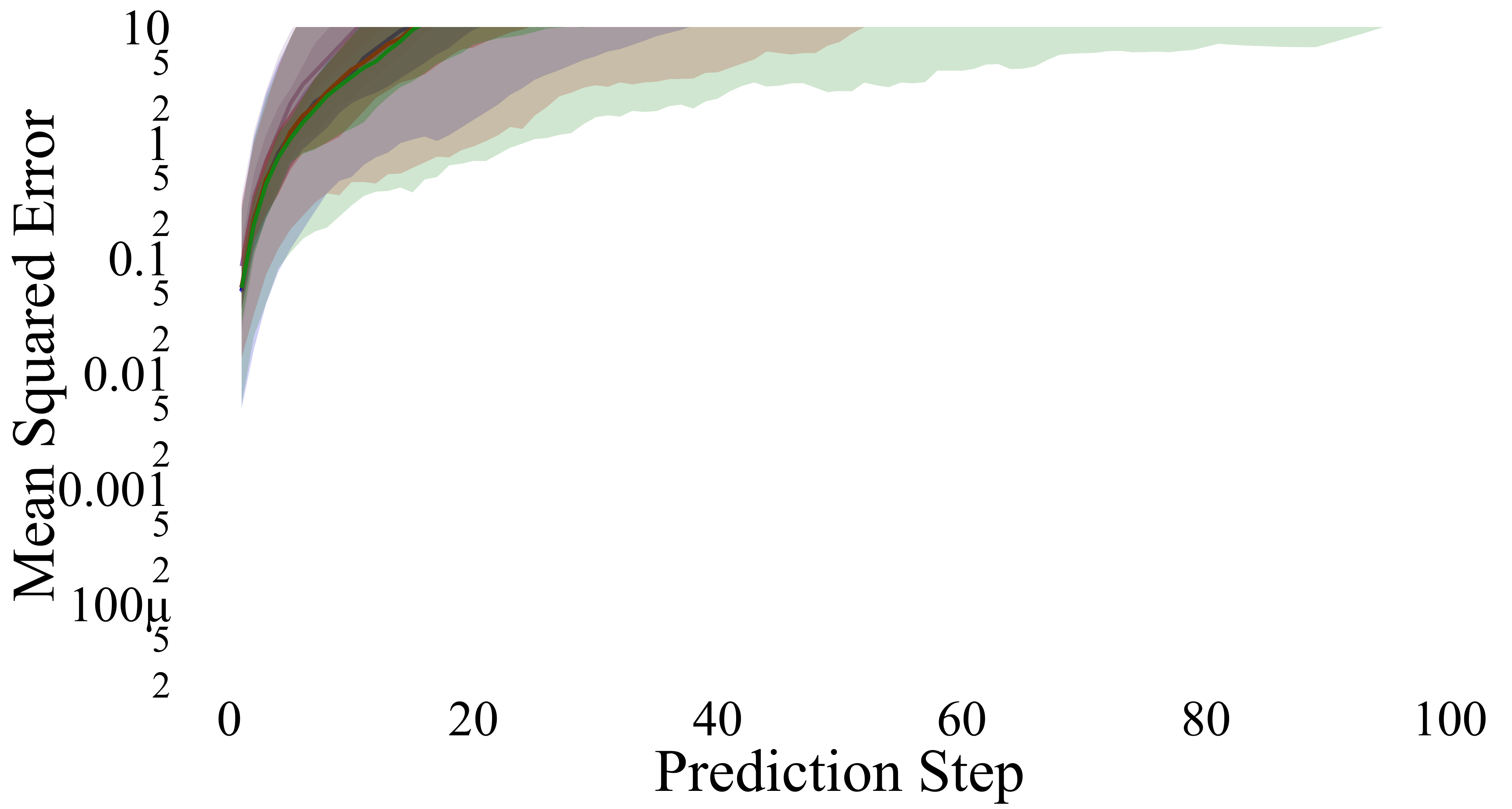}
        }
        \\
    \else
        \begin{subfigure}[t]{0.24\linewidth}
            \centering
            \includegraphics[width=\linewidth]{figures/p010/256hid.pdf}
            \caption{Default model. 
            $\rho=0.1$.}    
            \label{fig:comparep0.1}
        \end{subfigure}
        \hfill
        \begin{subfigure}[t]{0.24\linewidth}  
            \centering 
            \includegraphics[width=\linewidth]{figures/p050/256hid.pdf}
            \caption{Default model. %
            $\rho=0.5$.}    
            \label{fig:comparep0.5}
        \end{subfigure}
        \hfill
        \begin{subfigure}[t]{0.24\linewidth}
            \centering
            \includegraphics[width=\linewidth]{figures/p075/256hid.pdf}
            \caption{Default model. %
            $\rho=0.75$.}    
            \label{fig:comparep0.75}
        \end{subfigure}
        \hfill
        \begin{subfigure}[t]{0.24\linewidth}  
            \centering 
            \includegraphics[width=\linewidth]{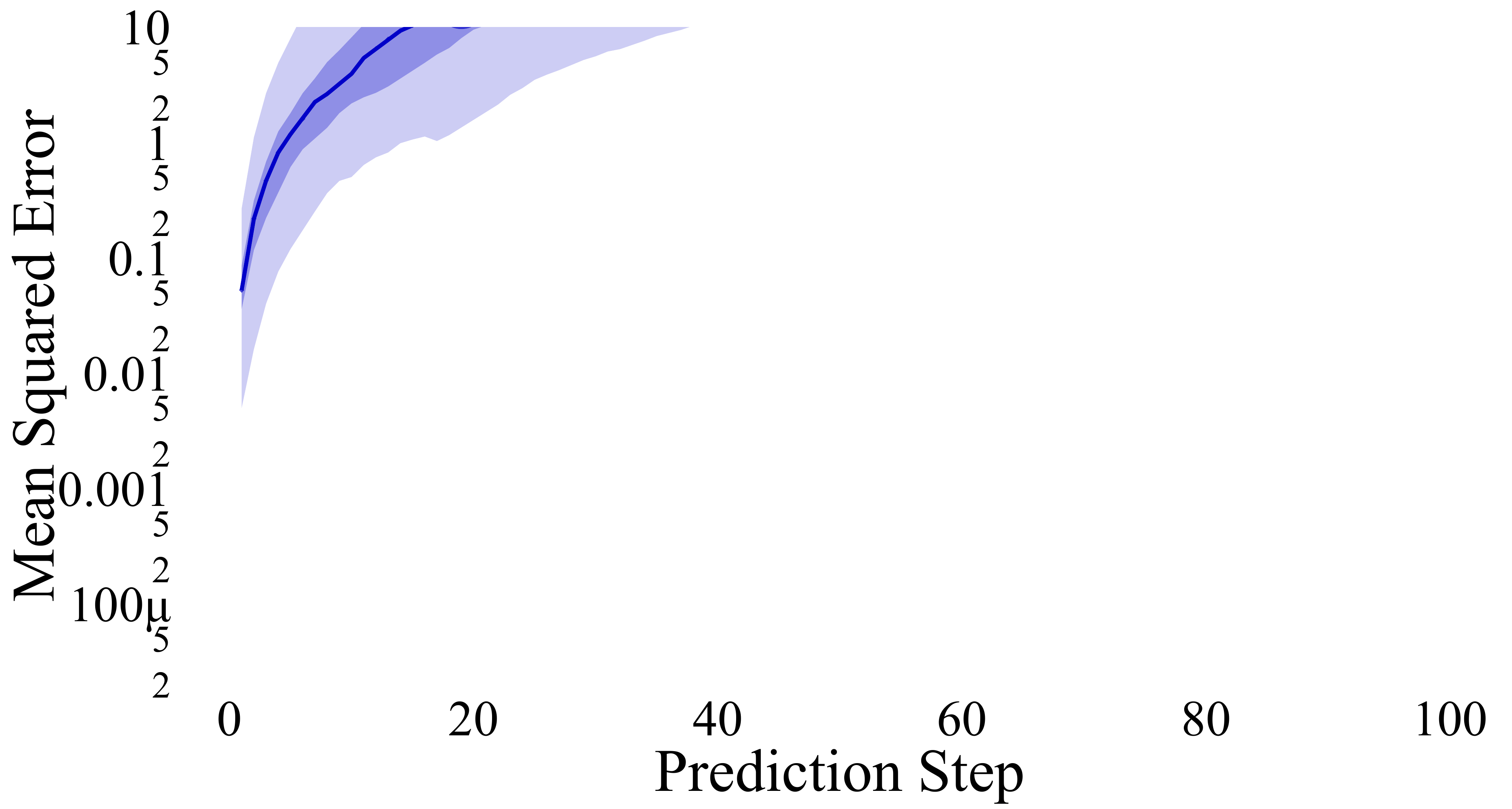}
            \caption{Default model. %
            $\rho=0.95$.}    
            \label{fig:comparep1.0}
        \end{subfigure}
        \\
        \begin{subfigure}[t]{0.24\linewidth}
            \centering
            \includegraphics[width=\linewidth]{figures/p010/32hid.pdf}
            \caption{\centering Hidden width 32. 
            }    
            \label{fig:comparep0.1}
        \end{subfigure}
        \hfill
        \begin{subfigure}[t]{0.24\linewidth}  
            \centering 
            \includegraphics[width=\linewidth]{figures/p050/32hid.pdf}
            \caption{\centering Hidden width 32. 
            }    
            \label{fig:comparep0.5}
        \end{subfigure}
        \hfill
        \begin{subfigure}[t]{0.24\linewidth}
            \centering
            \includegraphics[width=\linewidth]{figures/p075/32hid.pdf}
            \caption{\centering Hidden width 32. 
            }    
            \label{fig:comparep0.75}
        \end{subfigure}
        \hfill
        \begin{subfigure}[t]{0.24\linewidth}  
            \centering 
            \includegraphics[width=\linewidth]{figures/p095/32hid.pdf}
            \caption{\centering Hidden width 32. 
            }    
            \label{fig:comparep1.0}
        \end{subfigure}
        \\
        \begin{subfigure}[t]{0.24\linewidth}
            \centering
            \includegraphics[width=\linewidth]{figures/p010/512hid.pdf}
            \caption{\centering Hidden width 512, 3 hidden layers. 
            }    
            \label{fig:comparep0.1}
        \end{subfigure}
        \hfill
        \begin{subfigure}[t]{0.24\linewidth}  
            \centering 
            \includegraphics[width=\linewidth]{figures/p050/512hid.pdf}
            \caption{\centering Hidden width 512, 3 hidden layers. 
            }    
            \label{fig:comparep0.5}
        \end{subfigure}
        \hfill
        \begin{subfigure}[t]{0.24\linewidth}
            \centering
            \includegraphics[width=\linewidth]{figures/p075/512hid.pdf}
            \caption{\centering Hidden width 512, 3 hidden layers. 
            }    
            \label{fig:comparep0.75}
        \end{subfigure}
        \hfill
        \begin{subfigure}[t]{0.24\linewidth}  
            \centering 
            \includegraphics[width=\linewidth]{figures/p095/512hid.pdf}
            \caption{\centering Hidden width 512, 3 hidden layers. 
            }    
            \label{fig:comparep1.0}
        \end{subfigure}
        \\
        \begin{subfigure}[t]{0.24\linewidth}
            \centering
            \includegraphics[width=\linewidth]{figures/p010/no_norm.pdf}
            \caption{\centering No normalization. 
            }    
            \label{fig:comparep0.1}
        \end{subfigure}
        \hfill
        \begin{subfigure}[t]{0.24\linewidth}  
            \centering 
            \includegraphics[width=\linewidth]{figures/p050/no_norm.pdf}
            \caption{\centering No normalization. 
            }    
            \label{fig:comparep0.5}
        \end{subfigure}
        \hfill
        \begin{subfigure}[t]{0.24\linewidth}
            \centering
            \includegraphics[width=\linewidth]{figures/p075/no_norm.pdf}
            \caption{\centering No normalization. 
            }    
            \label{fig:comparep0.75}
        \end{subfigure}
        \hfill
        \begin{subfigure}[t]{0.24\linewidth}  
            \centering 
            \includegraphics[width=\linewidth]{figures/p095/no_norm.pdf}
            \caption{\centering No normalization. 
            }    
            \label{fig:comparep1.0}
        \end{subfigure}
        \\
        \begin{subfigure}[t]{0.24\linewidth}
            \centering
            \includegraphics[width=\linewidth]{figures/p010/model_type.pdf}
            \caption{\centering More model types. 
            }    
            \label{fig:comparep0.1}
        \end{subfigure}
        \hfill
        \begin{subfigure}[t]{0.24\linewidth}  
            \centering 
            \includegraphics[width=\linewidth]{figures/p050/model_type.pdf}
            \caption{\centering More model types. 
            }    
            \label{fig:comparep0.5}
        \end{subfigure}
        \hfill
        \begin{subfigure}[t]{0.24\linewidth}
            \centering
            \includegraphics[width=\linewidth]{figures/p075/model_type.pdf}
            \caption{\centering More model types. 
            }    
            \label{fig:comparep0.75}
        \end{subfigure}
        \hfill
        \begin{subfigure}[t]{0.24\linewidth}  
            \centering 
            \includegraphics[width=\linewidth]{figures/p095/model_type.pdf}
            \caption{\centering More model types. 
            }    
            \label{fig:comparep1.0}
        \end{subfigure}
    \fi
    \input{figure_latex/legend}
    \caption{
    Comparing the effects of common modeling tools on MSE (median, $65^\text{th}$, and $95^\text{th}$ percentiles) -- changing models from top to bottom with increasing poles from right to left.
    Reducing the model capacity by lowering layer width from 256 to 32 units or by removing data normalization does not substantially affect prediction accuracy on the simple state-space system.
    The probabilistic ensemble is able to improve on prediction accuracy, though contrary to common practices, only when using the state-based predictions.
    }
    \label{fig:model_paramet}
\end{figure}

\subsubsection{Model: Normalization}
\label{sec:normalization}
Tools for designing and optimizing neural networks are designed to work on data centered around unit normal distributions -- \textit{i.e.} identical and independently distributed data closely centered around 0. 
In robotics data where one-step models are deployed, this is often not the case, which leaves it up the user to maintain data cleaning practices for dynamics model training.

Normalization techniques map state and action variables over different ranges (bounds) and shapes (relative density) to well-behaved distributions to aid model training. 
The model normalizes the inputs and targets at training, and at prediction time utilizes these distributions to map new inputs to the latent space of the model and then back into the true distribution.
Such mappings can also contribute to compounding error by pushing both inputs and targets of some validation data further outside of a training distribution.
In this work, we map continuous variables to a normal distribution $\mathcal{N}(0,1)$ and bounded variables (such as actions) to a uniform distribution $\mathcal{U}(-1,1)$.
The effects of turning this normalization off is a small increase in prediction error, shown in \fig{fig:model_paramet}(i-l).
Normalization is heavily sensitive to outliers because if some training points are substantially outside the distribution, it will further concentrate the data of interest onto a small region of the input space, resulting in a harder learning problem and one that is more sensitive to model bias.

\subsection{Other Factors Impacting Compounding Error}
\label{sec:other_causes}
In this section, we build upon our study of system and model properties to show how some of these variables can interweave in complex manners, resulting in difficultly to forecast model performance. 

\begin{figure}
    \centering
    \ifjmlrutilsmaths
        \subfigure[Cartpole. $d_s + d_a = 5$.]{
        \includegraphics[width=0.31\linewidth]{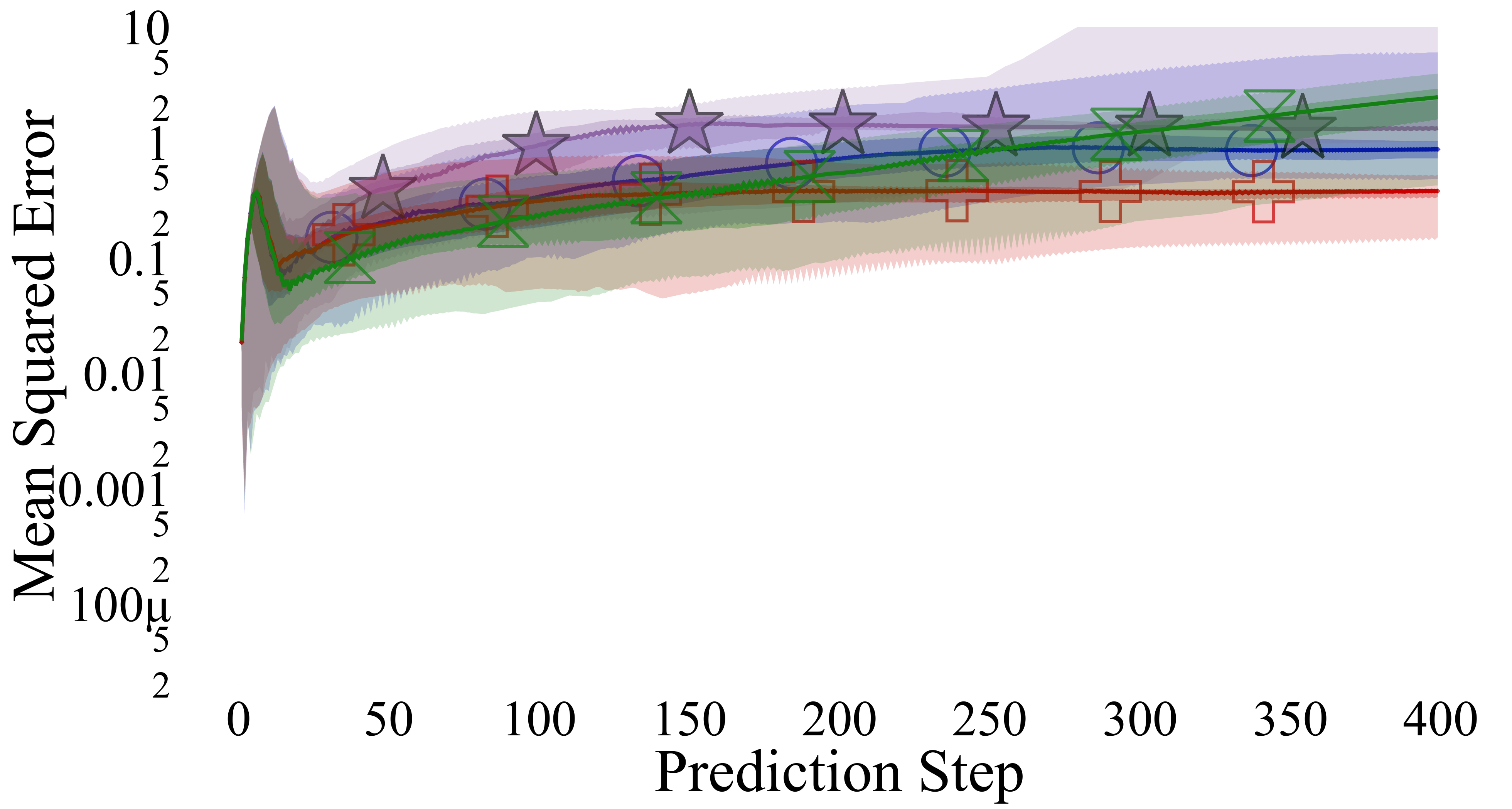}
        }
        \hfill
        \subfigure[Quadrotor. $d_s + d_a = 13$.]{
        \includegraphics[width=0.31\linewidth]{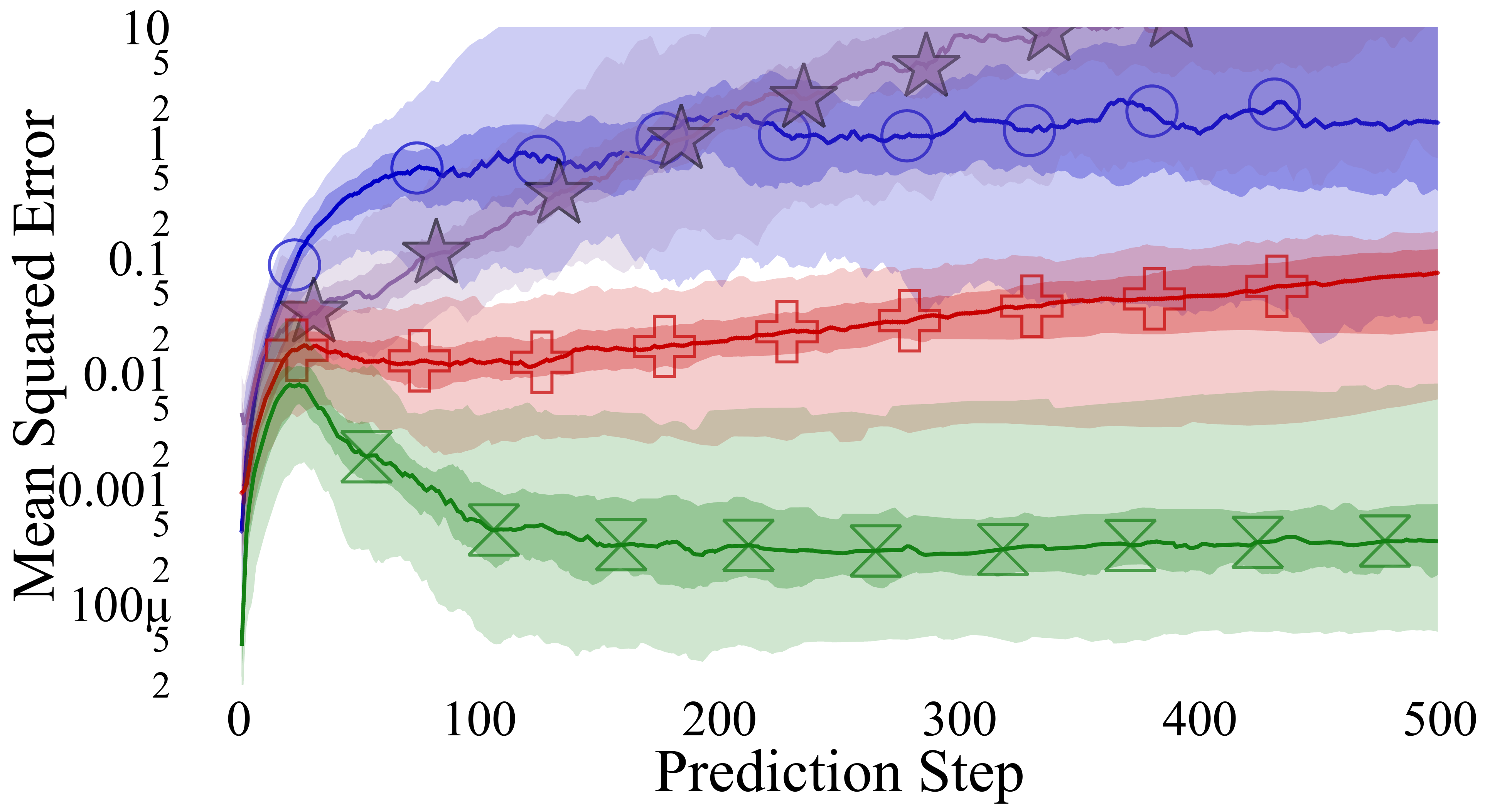}
        }
        \hfill
        \subfigure[Reacher. $d_s + d_a = 20$.]{
        \includegraphics[width=0.31\linewidth]{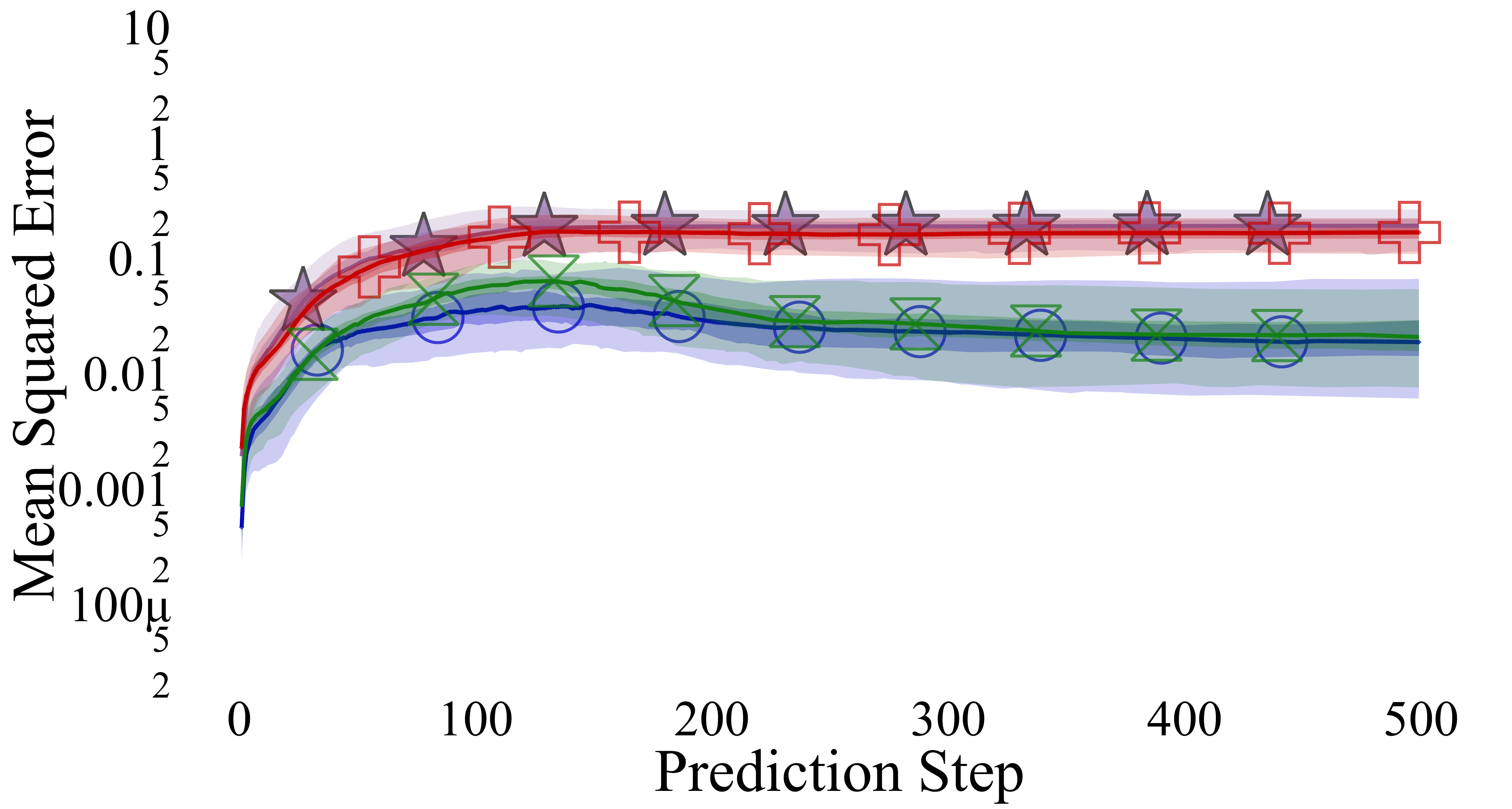}
        \label{fig:reacher-compare}
}
    \else    
     \begin{subfigure}[t]{0.32\linewidth}  
        \centering 
        \includegraphics[width=\linewidth]{figures/deltaVtrue/cartpole_react.pdf}
        \caption{Cartpole. $d_s + d_a = 5$.}  
        \label{fig:cp-mod-fb}
    \end{subfigure}
    ~
    \begin{subfigure}[t]{0.32\linewidth}  
        \centering 
        \includegraphics[width=\linewidth]{figures/deltaVtrue/crazyflie_react.pdf}
        \caption{Quadrotor. $d_s + d_a = 13$. }    
        \label{fig:reacher-compare}
    \end{subfigure}
    ~
    \begin{subfigure}[t]{0.32\linewidth}  
        \centering 
        \includegraphics[width=\linewidth]{figures/deltaVtrue/reacher_react.pdf}
        \caption{Reacher. $d_s + d_a = 20$. }  
        \label{fig:cp-mod-fb}
    \end{subfigure}
    \fi
    \vspace{0pt}
    \input{figure_latex/legend}
    \caption{
Showing how re-computing actions has a dramatically different effect on prediction MSE depending on the policy type (median, $65^\text{th}$, and $95^\text{th}$ percentiles).
We hypothesize that these diverging predictions could be worsened when coupled with feed-forward control policies.
The data used is the same as in \fig{fig:examples-compare} where the policy is computed based on the predicted state, rather than mimicking the original action.
Note, the predictions for the different environments are across different horizons.
}
    \label{fig:examples-compare-react}
\end{figure}

\subsubsection{System: Re-computing Actions}
\label{sec:recompute}
When planning into the future there are two potential actions sequences: 
a logged action sequence to compare the model accuracy of a learned model to measured data and 
a generated action sequence to evaluate the potential usefulness of a simulated trajectory.
The effect of re-computing the actions passed into the predictive model as $\mdpaction_t = \pi(\hat{\mdpstate}_t)$ instead of the original action sequence being provide by an oracle can be crucial to if a model will be useful under a certain controller.
The action on a predicted state will take the form of $\mdpaction = \pi(\hat{\mdpstate}) = \pi(\mdpstate + \epsilon_t)$, so the action returned varies in most the model accuracy and policy robustness to perturbation.
Depending on the problem formulation, long horizon prediction is often done with an action sequence passed into the model (representing the ground truth), but model accuracy can be dramatically different if the actions are re-computed from the predicted state as computed action sequences will also exhibit compounding error. 

The original results of the models on the simulated robotic tasks are shown in \fig{fig:examples-compare} and the results where the oracle no longer provides the original action sequence are shown in \fig{fig:examples-compare-react}.
In this case, all environments show approximately a $10\times$ increase in error when not given the action sequence from an oracle.
Crucially, this type of planning without a real action sequence is how most MPC algorithms compute the action with a learned model.
One may expect that reflexive policies recomputing actions would diverge faster because they compute the control off only the current state, while controllers with built in damping or slew limiting can predict more accurately (such as the integral or derivative terms in a PID controller), but this trend is not clear in our simulated results.

Generated action sequences are closely linked to using the models for control, but generally model training is only evaluated on accuracy with logged data. 
To date, there are no methods to evaluate the potential accuracy of randomly generated actions leaving future work to understand this relationship -- for example, by evaluating the bootstrapped uncertainty estimate of a probabilistic ensemble across a planned trajectory.

\begin{figure}
    \centering
    \ifjmlrutilsmaths
        \subfigure[ Predictions - clustered $s_0$. ]{
        \includegraphics[width=0.4\linewidth]{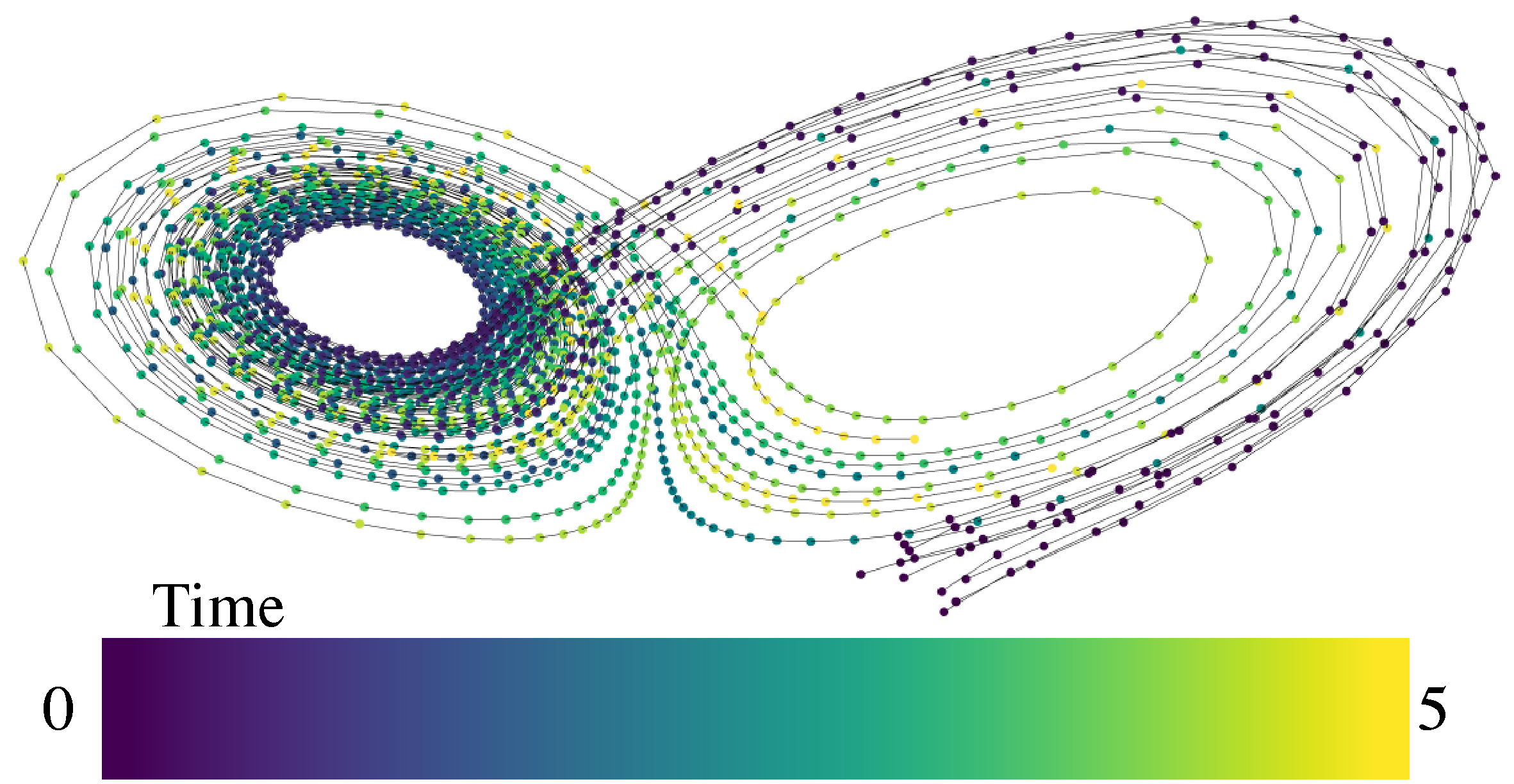}
        }
        \quad
        \subfigure[Errors - clustered $s_0$.]{
        \includegraphics[width=0.4\linewidth]{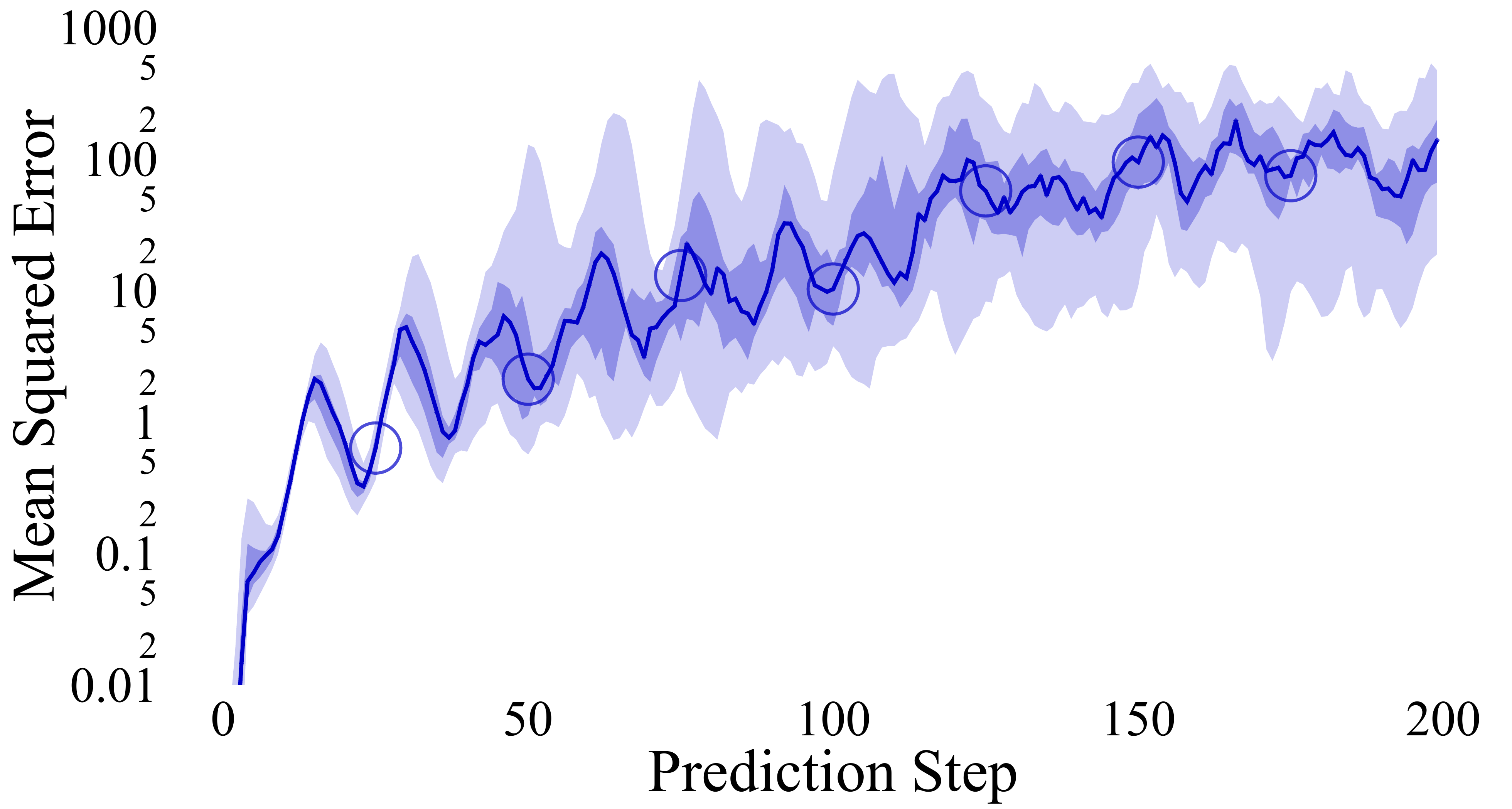}
        }
        \\
        \subfigure[Predictions - broad $s_0$.]{
        \includegraphics[width=0.4\linewidth]{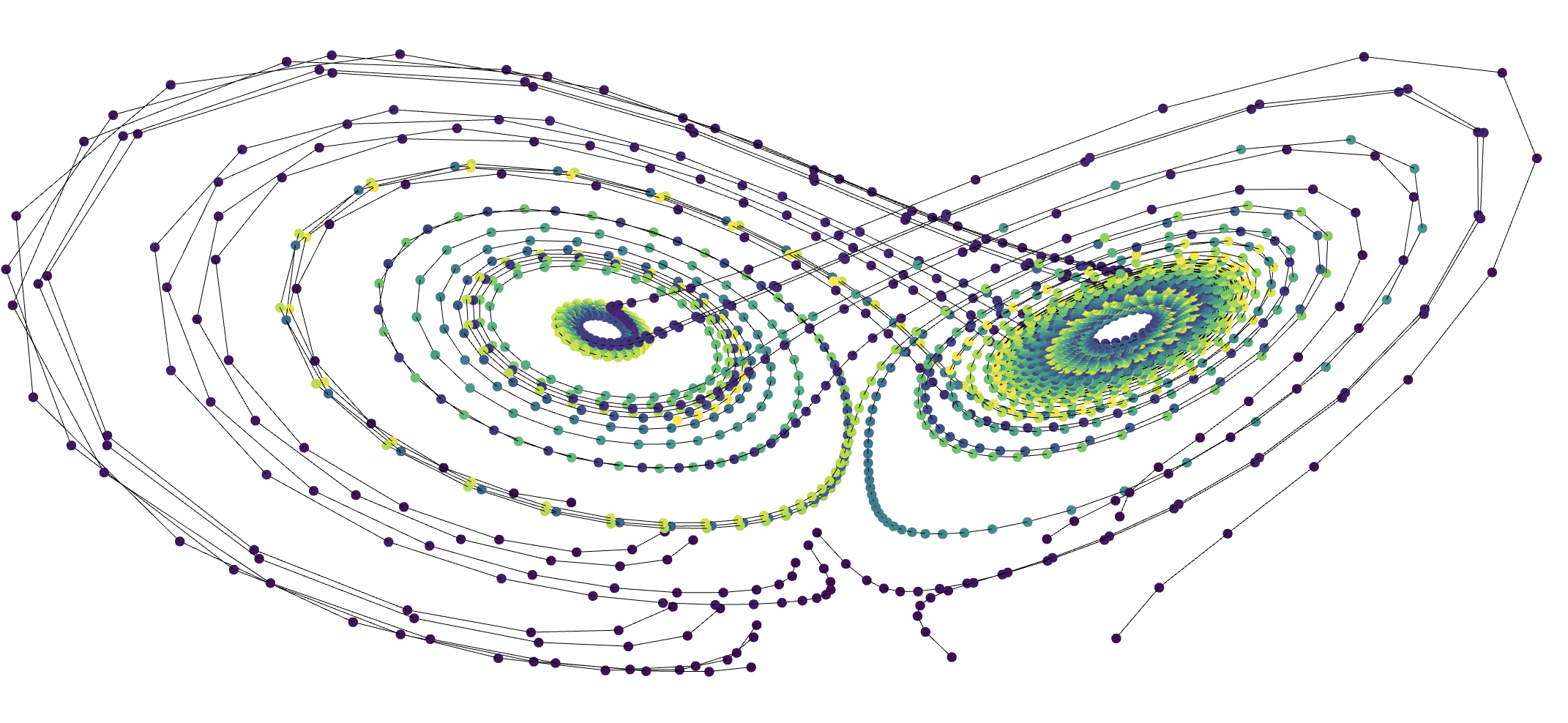}
        }
        \quad
        \subfigure[Errors - broad $s_0$.]{
        \includegraphics[width=0.4\linewidth]{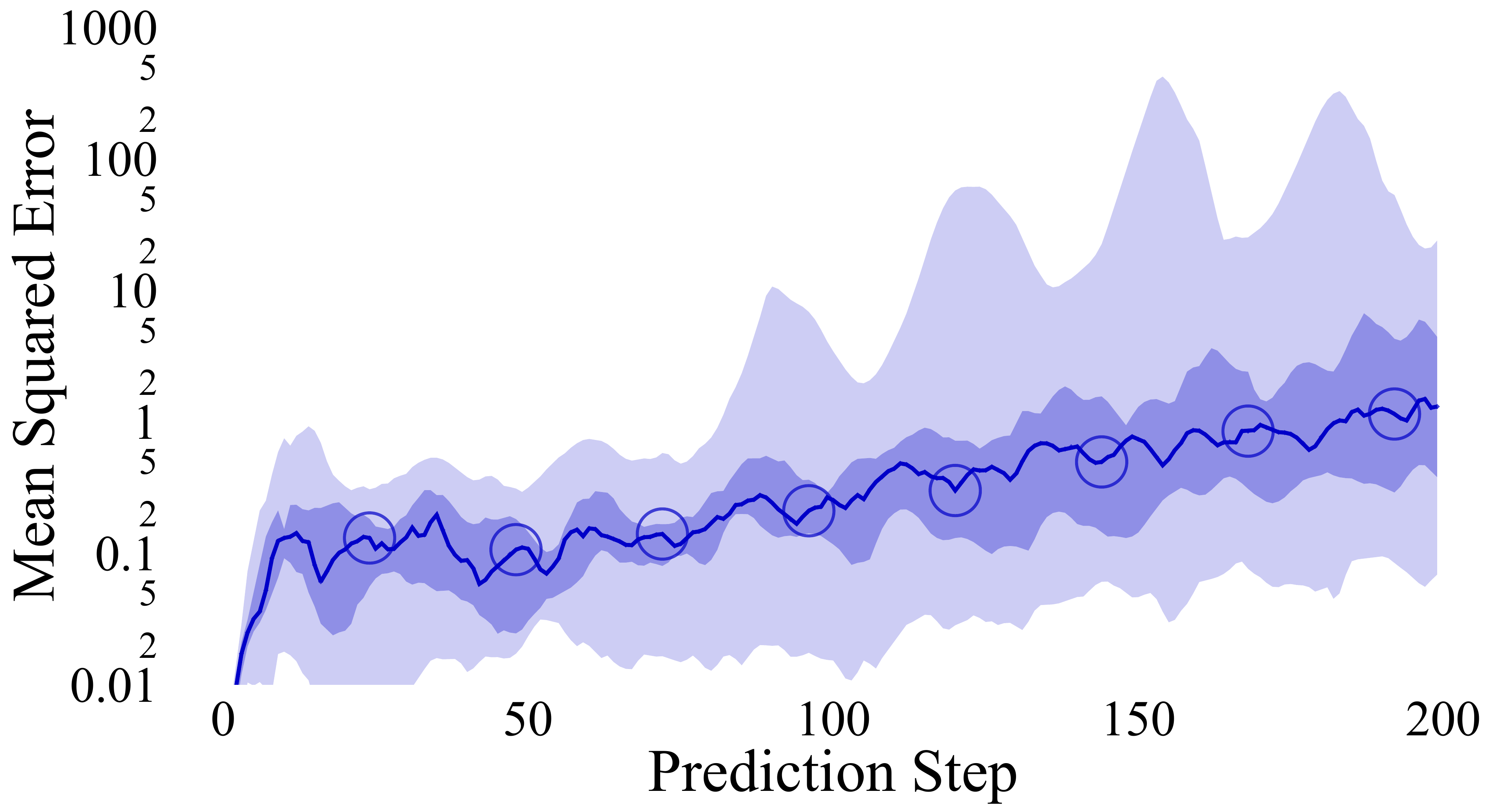}
        }
    \else
       \begin{subfigure}[t]{0.22\linewidth}
            \centering
            \includegraphics[width=\linewidth]{figures/lorenz/lorenz-bar.png}
            \caption{\centering Lorenz system, $x_0,y_0,z_0 \in [5,10] $. 
            }    
            \label{fig:lorenza}
        \end{subfigure}
        ~
        \begin{subfigure}[t]{0.25\linewidth}  
            \centering 
            \includegraphics[width=\linewidth]{figures/lorenz/lorenz_narrow.pdf}
            \caption{\centering Compounding prediction error, $x_0,y_0,z_0 \in [5,10] $. 
            }    
            \label{fig:lorenz2}
        \end{subfigure}
        ~
        \begin{subfigure}[t]{0.22\linewidth}
            \centering
            \includegraphics[width=\linewidth]{figures/lorenz/broad.png}
            \caption{\centering Lorenz system, $x_0,y_0,z_0 \in [-10,10]$.
            }    
            \label{fig:lorenza}
        \end{subfigure}
        ~
        \begin{subfigure}[t]{0.25\linewidth}  
            \centering 
            \includegraphics[width=\linewidth]{figures/lorenz/lorenz_broad.pdf}
            \caption{\centering Compounding prediction error, $x_0,y_0,z_0 \in [-10,10]$. 
            }    
            \label{fig:lorenz2}
        \end{subfigure}
    \fi
    \caption{Highlighting the dynamics of the of the Lorenz system (\textit{a,c}) as a challenging dynamics problem due to its chaotic dynamics that result in multi-modal behavior.
    The chaotic system also happens to be stable, which is reflected by its bounded prediction error per-step (\textit{b,d}: median, $65^\text{th}$, and $95^\text{th}$ percentiles) and a testing set of new trajectories.
    (\textit{a,b}) are trajectories sampled from a more restricted initial condition, where the initial $x,y,z$ coordinates fall in $[5,10]$.
    (\textit{c,d}) is a more diverse training and testing set, where the initial states $x,y,z$ are sampled from $[-10,10]$, though the dynamics model generalizes better to new previously unseen data. 
    }
    \label{fig:lorenz}
\end{figure}

\subsubsection{Example: Predicting Chaotic Dynamics}
\label{sec:lorenz}

A fundamental limit of prediction can be posed as how to predict chaotic systems.
A chaotic system is defined by the idea that a small perturbation in state can grow to an exponential difference over time. 
As a case study, we include prediction errors for the Lorenz system~\cite{lorenz1963deterministic}, shown in \eq{eq:lorenz}.
The canonical parameter $\eta$ is often $\rho$, but we have replaced it to avoid overloading our symbol for pole.
\begin{align}
    \dot{x} & = \sigma(y-x) \\
    \dot{y} & = x(\eta - z) - y \\
    \dot{z} & = xy-\beta z
    \label{eq:lorenz}
\end{align}

\begin{wrapfigure}{r}{0.4\textwidth}
    \vspace{-15pt}
    \centering
    \begin{center}
    \small{
    $\CIRCLE$ \text{True},
    $\diamond$ \text{Predicted Trajectory}
    }
    \end{center}

    \includegraphics[width=\linewidth]{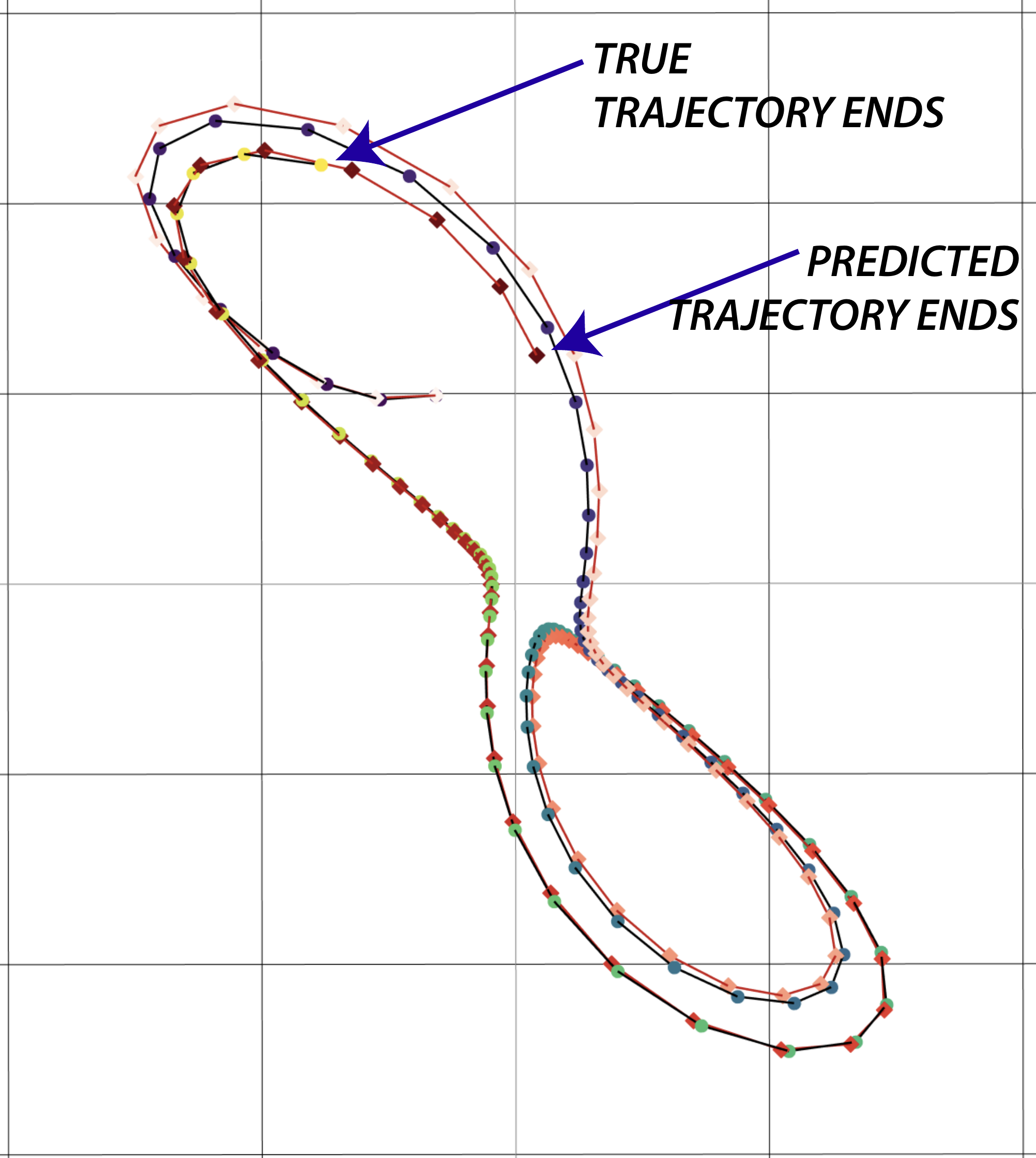}
    \caption{Example showing the prediction on the Lorenz system, where the predictions tend to advance in time and do not diverge rapidly.
    }
    \label{fig:lorenz-ex}
    \vspace{-30pt}
\end{wrapfigure}

In this work, the initial states for the Lorenz system are constrained to two different distributions: $x_0,y_0,z_0 \in [5,10]$ or  $x_0,y_0,z_0 \in [-10,10]$, resulting in a comparison between how training and test data distribution can affect dynamics model performance. 
Both training sets include 100 trajectories of length 500 that are evaluated on 100 previously unseen trajectories of length 200.
While in practice learning to identify the three governing parameters of the dynamics could result in more accurate predictions, learning models for systems by which the analytical equations are unknown poses a problem of great interest for the field.
As the simulated version of system has no noise and stable dynamics, the prediction error do not growth to infinity, but rather proportional to the separation of the two stable points, shown by the oscillations in \fig{fig:lorenz}.

\subsubsection{Example: Control Frequency \& Signal to Noise Ratio}
\label{sec:snr}
The signal to noise ratio (SNR)~\cite{stanley1988digital} is a metric for evaluating the relative strength of a signal that one wishes to measure to the noise that will be present in measurements, commonly deploying in digital signal processing.
A related topic emerges with any dynamical system, where changing the sampling frequency of a system with uniform observation noise can implicitly change the SNR of the transition labels for supervised learning -- a shorter sample time leads to higher impact of noise.

Consider a canonical control system, the double integrator, shown in \eq{eq:doublei}, that is the underlying dynamics of Newtownian systems:
\begin{equation}
    \dot{\vec{x}}(t) = \begin{bmatrix}
        0 & 1 \\
        0 & 0 \\ 
    \end{bmatrix} \vec{x}(t) + 
    \begin{bmatrix}
        0  \\
        1 \\ 
    \end{bmatrix} u(t), \quad
    \vec{o}(t) = \vec{x}(t) + \omega^m(t)
    \label{eq:doublei}
\end{equation}
With a constant input -- corresponding to a constant force -- solutions to this equation take the form of a quadratic function.
With a set measurement noise level the dynamics of a given system when sampled at different frequencies results in modeling problems of varying difficult, which we propose understanding as a signal to (measurement) noise ratio.
The true change in state, $s_{t+1}-s_t$ is corrupted by some noise from the current and previous measurements, $\omega^m_t$ and $\omega^m_{t+1}$, acting on the current and past observations $\vec{o}$, where the relative size of the true dynamics can be described as a signal-to-noise ratio (SNR):
\begin{equation}
  \text{SNR} \approx  \frac{\|s_{t}-s_{t-1}\|}{  \|s_{t}-s_{t-1}+\omega^m_t+\omega^m_{t-1}\|}
\end{equation}
Crucial to accurately modelling dynamics is for the sampling rate to be slow enough by which the noise is a minor contribution to the targets, $\text{SNR}>>1$.
An illustration of this example is shown in \fig{fig:snr}, where a constant noise interval illustrates the possible data labels with different sample rates.
All three simulators in this work do not have measurement error, though every real system's measurement error is determined by the quality of on-board or external state measurement.
In the real quadrotor system that follows in \sect{sec:realworld}, we evaluate model accuracy for two sampling rates.
Finally, in real systems noise distributions often take on asymmetric and complex distributions, which can be measured and understood as specific detriments to learned model accuracy.

\begin{figure}
    \centering
    \ifjmlrutilsmaths
        \subfigure[\SI{250}{\milli \second} step size.]{
        \includegraphics[width=0.31\linewidth]{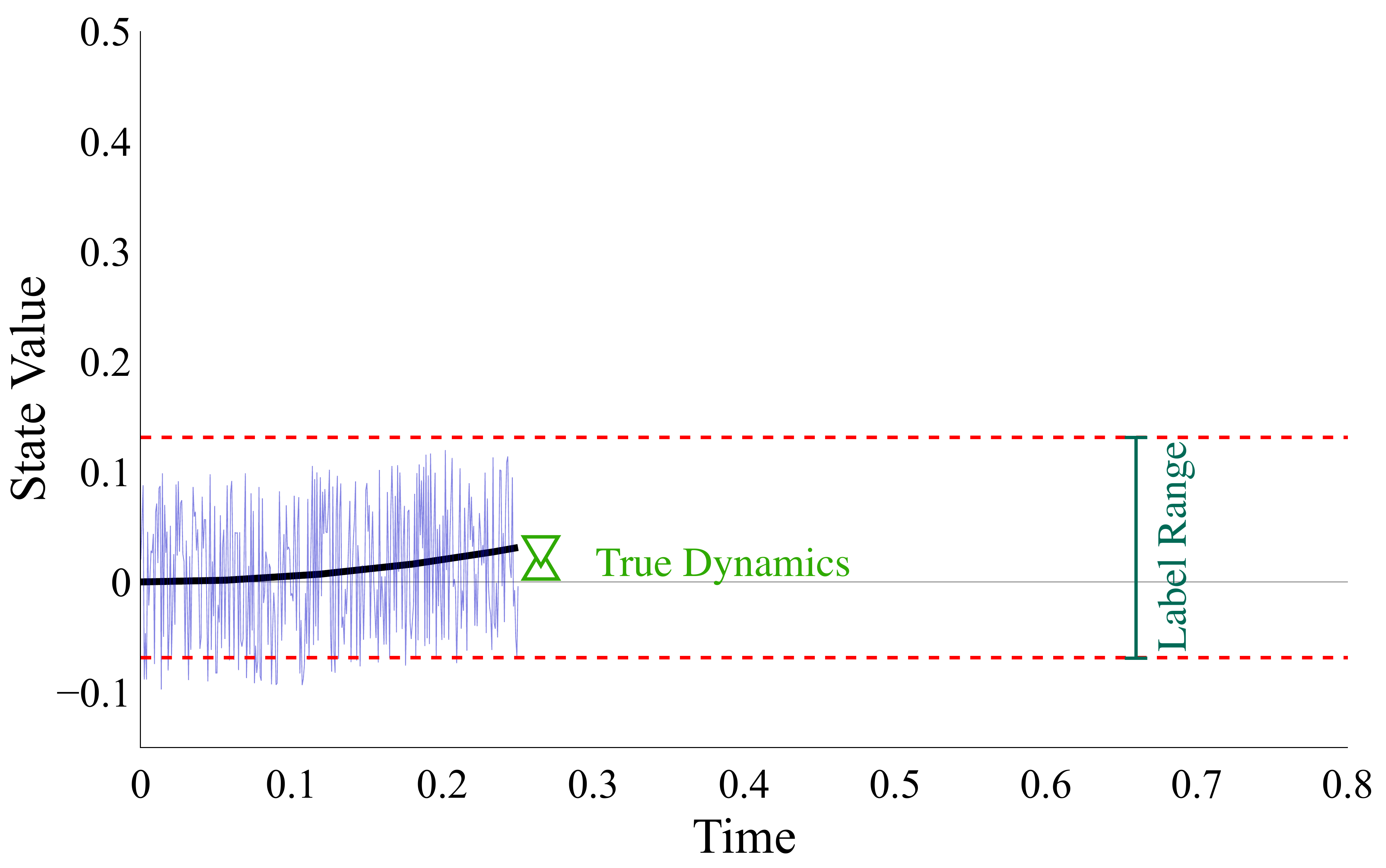}
        }
        \hfill
        \subfigure[\SI{500}{\milli \second} step size.]{
        \includegraphics[width=0.31\linewidth]{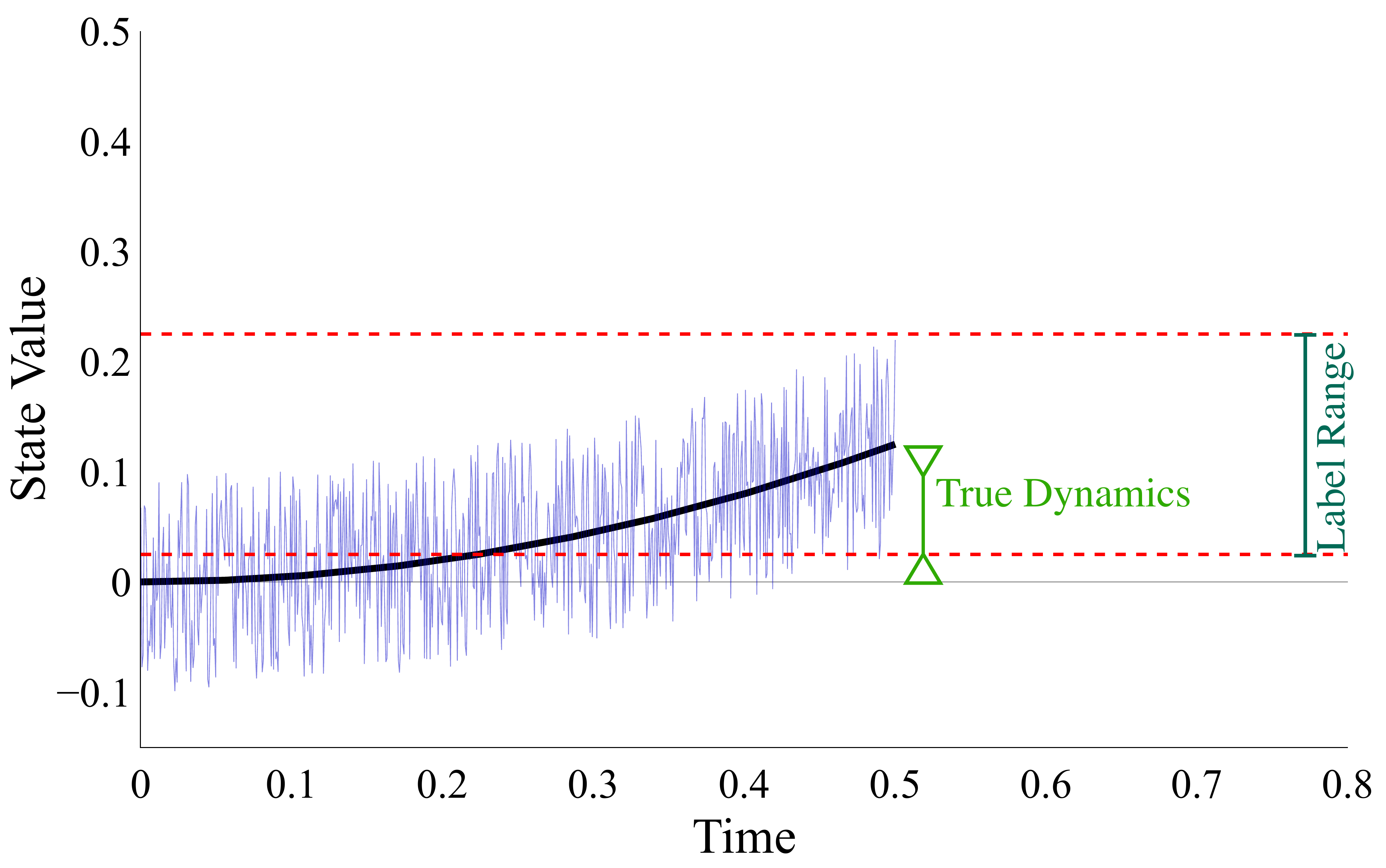}
        }
        \hfill
        \subfigure[\SI{750}{\milli \second} step size.]{
        \includegraphics[width=0.31\linewidth]{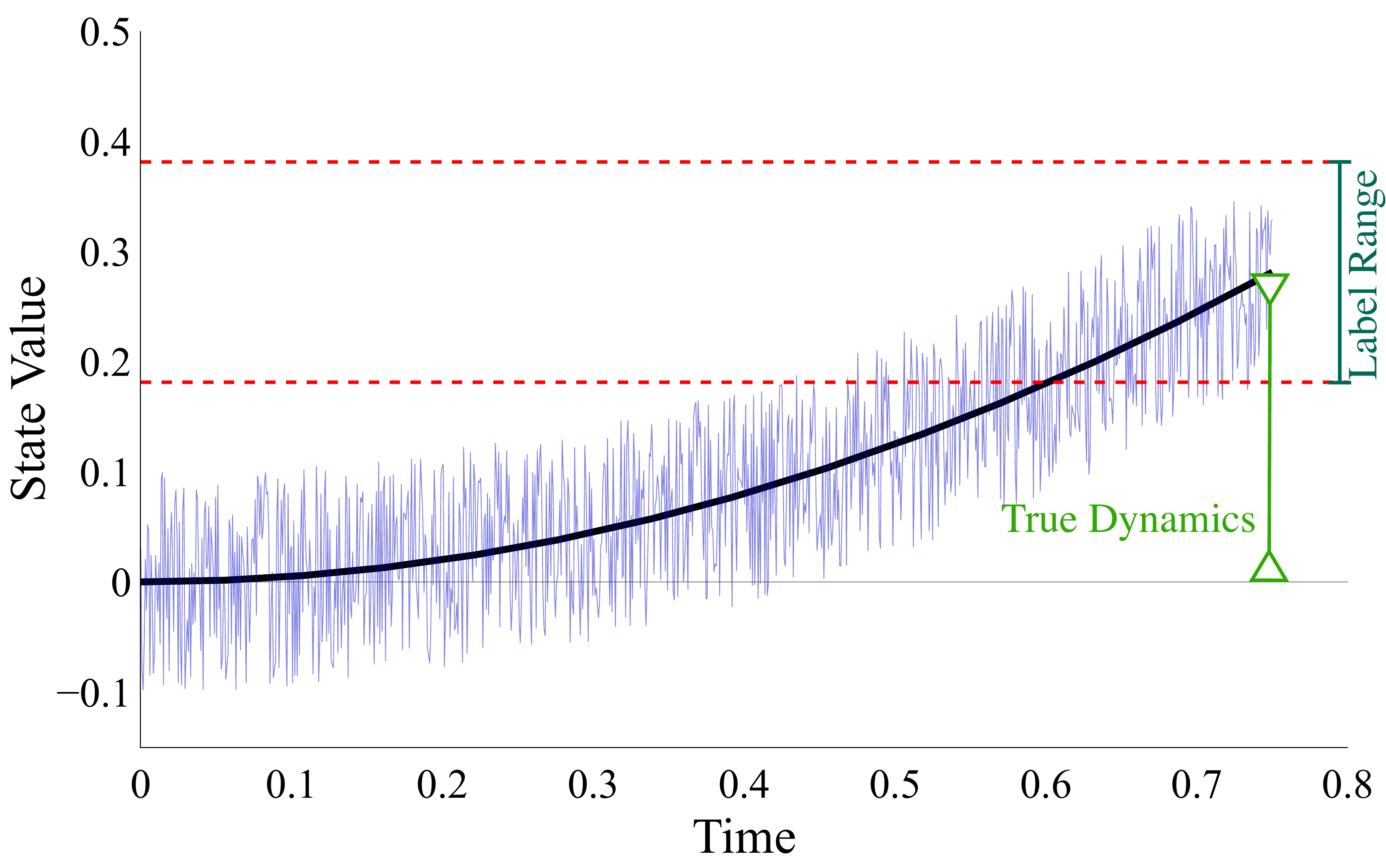}
        }
    \else
   \begin{subfigure}[t]{0.32\linewidth}
        \centering
        \includegraphics[width=\linewidth]{figures/snr/p25.pdf}
        \caption{\centering \SI{250}{\milli \second} step size.
        SNR $<<1$.
        }    
        \label{fig:snr250}
    \end{subfigure}
    ~
    \begin{subfigure}[t]{0.32\linewidth}  
        \centering
        \includegraphics[width=\linewidth]{figures/snr/p50.pdf}
        \caption{\centering \SI{500}{\milli \second} step size.
        SNR $\approx 1$.
        }    
        \label{fig:snr500}
    \end{subfigure}
    ~
    \begin{subfigure}[t]{0.32\linewidth}  
        \centering
        \includegraphics[width=\linewidth]{figures/snr/p75.pdf}
        \caption{\centering \SI{750}{\milli \second} step size.
        SNR $>>1$.
        }    
        \label{fig:snr500}
    \end{subfigure}
\fi
    \caption{
Showing how signal to noise ratio and step size are important for robotic tasks.   
The difference between the potential measurement regions is the magnitude of signal present in the labelled data. 
Crucially, a slower sampling frequency can increase the resolution of the labelled data, improving downstream prediction accuracy. 
}
    \label{fig:snr}
\end{figure}
\fi

\end{document}